\documentclass[conference, 10pt]{IEEEtran} 

\IEEEoverridecommandlockouts
\usepackage{amsmath,amssymb,amsfonts}
\usepackage{amsthm}
\usepackage{gensymb}
\usepackage{graphicx}
\usepackage{float}
\usepackage{textcomp}
\usepackage{multicol}
\usepackage{xcolor}
\usepackage{tcolorbox}
\usepackage{mathtools} % for 'dcases*' env.
\usepackage{listings}
\usepackage{hyperref}
\usepackage{url}
\usepackage{multicol,lipsum,environ}
\usepackage[top=1in,bottom=1in,left=0.75in,right=0.75in]{geometry}
\usepackage{booktabs}
\usepackage{pythonhighlight}
\usepackage{etoolbox}
\usepackage[font=small, labelfont=bf]{caption}

\usepackage{dirtytalk}

\usepackage{adjustbox} % for max totalheight

\usepackage{matlab-prettifier}
\usepackage{tikz}
\usetikzlibrary{positioning}
\usepackage{subcaption}
\usepackage{hyperref}
\usepackage[capitalize]{cleveref}
\usepackage[numbers]{natbib}
\def\BibTeX{{\rm B\kern-.05em{\sc i\kern-.025em b}\kern-.08em
    T\kern-.1667em\lower.7ex\hbox{E}\kern-.125emX}}

\crefname{figure}{Fig.}{Figs.}
\crefname{section}{Sec.}{Secs.}
\crefname{table}{Table.}{Tabs.}
\crefname{algorithm}{Alg.}{Algs.}

\usepackage{multibib}
\newcites{supp}{Supplementary References}

\usepackage{enumitem}   % for the description list
\usepackage{calc}
\usepackage[ruled,vlined]{algorithm2e}

\definecolor{darkgreen}{RGB}{0,100,0}
\definecolor{darkred}{RGB}{100,0,0}
\definecolor{darkblue}{RGB}{0,0,100}

\IEEEoverridecommandlockouts

\makeatletter
\let\@oldmaketitle\@maketitle
\renewcommand{\@maketitle}{\@oldmaketitle
\vspace{4pt}
\centering
\setcounter{figure}{0}
\includegraphics[width=\linewidth]{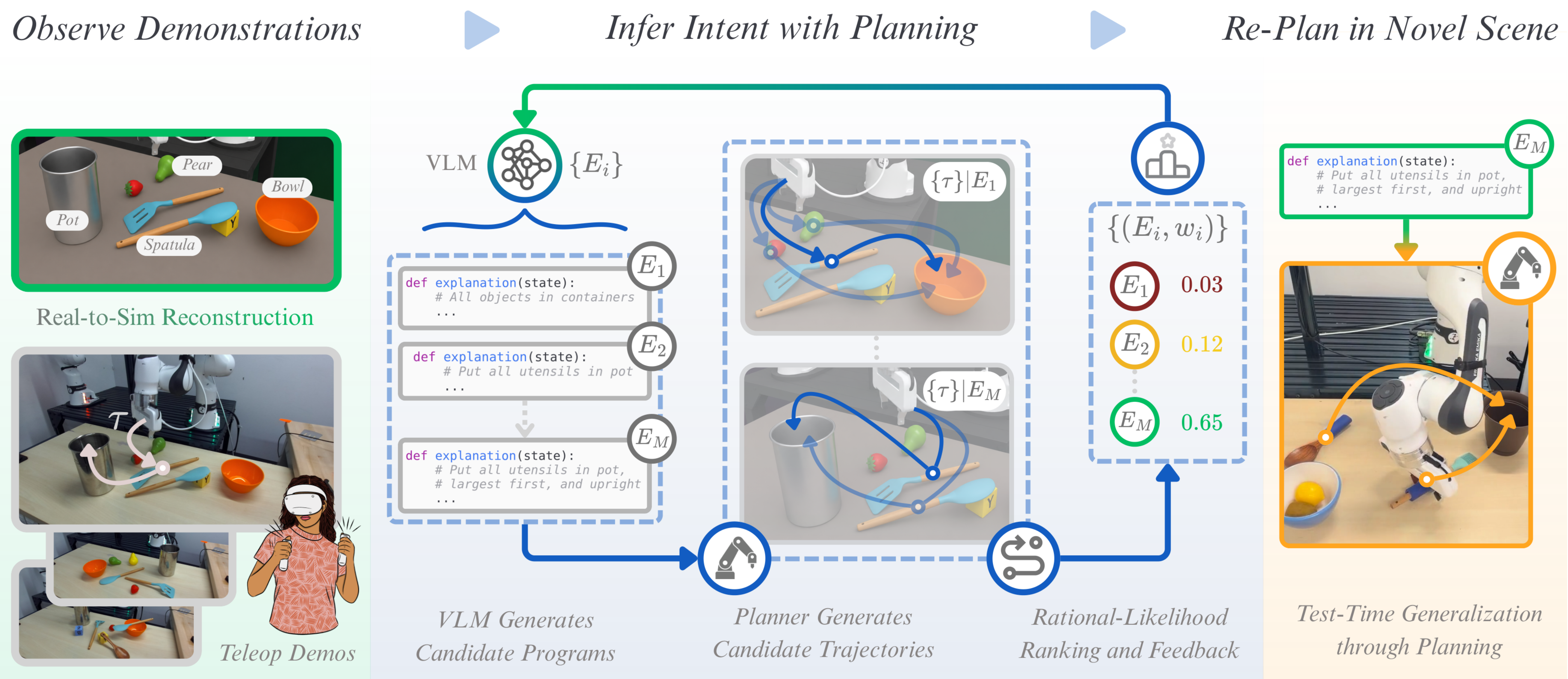}
\captionof{figure}{The RIR pipeline. (1) Given 1-3 demos, 3D real-to-sim reconstruction builds an object-oriented state representation of the demonstration and scene. (2) A VLM proposes intent hypotheses in the form of Python programs (the $E_i$ in the figure above are shown in language for compactness) that construct calls to a generalized task and motion planner. (3) These hypotheses are evaluated in terms of how rational the demonstrations would have been, had the explanation been the true demonstration intent. (4) A weighted particle inference loop uses the scored hypotheses to query the VLM for a set of improved hypotheses, ultimately producing a final best explanation. (5) After 3D reconstruction of a novel scene, the best explanation program is executed, with planner calls that produce a final robot trajectory which is executed on the real robot.
}
\label{fig:teaser}
}

\makeatother

\usepackage{etoolbox} % For patching commands
\usepackage[shortcuts]{extdash}

\usepackage{makecell}

\usepackage{amsmath,amssymb,amsfonts}
\usepackage{amsthm}

\tcbset {
  base/.style={
    arc=0mm, 
    bottomtitle=0.5mm,
    boxrule=0mm,
    colbacktitle=black!10!white, 
    coltitle=black, 
    fonttitle=\bfseries, 
    left=2.5mm,
    leftrule=1mm,
    right=3.5mm,
    title={#1},
    toptitle=0.75mm, 
  }
}

\definecolor{brandblue}{rgb}{0.07, 0.364, 0.761}
\newtcolorbox{mainbox}[1]{
  colframe=brandblue, 
  base={#1}
}

\newtcolorbox{subbox}[1]{
  colframe=black!30!white,
  base={#1}
}

\crefname{figure}{Fig.}{Figs.}
\crefname{section}{Sec.}{Secs.}
\crefname{table}{Tab.}{Tabs.}

\hypersetup{
    colorlinks=true,
    linkcolor=brandblue,
    filecolor=magenta,      
    urlcolor=brandblue,
    pdftitle={Rational Inverse Reasoning},
    pdfauthor={Ben Zandonati},
    pdfsubject={Robotics, sample-efficient imitation learning via planning}
    pdfpagemode=FullScreen,
    citecolor=brandblue
    }

\begin{document}

\title{\vspace{-0.5cm} Rational Inverse Reasoning: Few-Shot Imitation by Inferring Intent through Planning }

% \author{
%     \IEEEauthorblockN{Ben Zandonati$^{*1}$}
%     \IEEEauthorblockA{bzando@mit.edu}
%     \and
%     \IEEEauthorblockN{Tomas Lozano-P\'erez$^{1}$}
%     \IEEEauthorblockA{tlp@mit.edu}
%     \and
%     \IEEEauthorblockN{Leslie Pack Kaelbling$^{1}$}
%     \IEEEauthorblockA{lpk@mit.edu} 
%     \thanks{\small $^1$ MIT CSAIL. Correspondence to: Ben Zandonati (bzando@mit.edu)}
% }

\author{%
\IEEEauthorblockN{Ben Zandonati\textsuperscript{*},
Tom\'as Lozano-P\'erez,
Leslie Pack Kaelbling}
\IEEEauthorblockA{MIT CSAIL}
\thanks{\textsuperscript{*}Corresponding author: Ben Zandonati (bzando@mit.edu).}
}

\maketitle

\begin{abstract}

Humans can learn a new manipulation task from one or two demonstrations and then perform it in a new room, with new objects, under new constraints. Modern robot imitation learning, in contrast, typically needs hundreds to thousands of demonstrations and still degrades under modest shifts in layout, geometry, object set or task constraints. We argue this gap is not just about data, but also about the level of abstraction at which learning occurs; generalization requires inferring the latent intent underlying why a demonstrator behaved in a certain way, rather than reproducing how they moved. We present Rational Inverse Reasoning (RIR), which casts few-shot imitation as inference over latent explanation programs: compact, executable descriptions of intent that map an object-centric scene to a structured task-and-motion-planning (TAMP) specification of goals, subgoals and constraints. A vision-language model proposes candidate programs, and a hierarchical planner supplies a bounded-rational likelihood. By combining VLM program proposals, and planner-grounded feedback, RIR iteratively refines the candidate set to approximate a posterior over concise, executable programs. On a 2D reasoning benchmark and a real Franka FR3, RIR recovers transferable task structure from as little as one demonstration. Generalizing to substantially new layouts and object sets, RIR outperforms VLM-planning baselines that lack explicit rationality and planning-grounded inference, increasing downstream success rate by $34$ and $28$ percentage points in the one- and three-shot settings.

% \textcolor{red}{Generalizing to substantially new layouts and object sets, RIR outperforms object-centric few-shot imitation learning and VLM-planning baselines that lack explicit rationality and planning-grounded inference.}
% \ben{We can add then when I've finished doing my cool ridiculously statistically significant simulated experiments.}

\end{abstract}

\section{Introduction} \label{sec:introduction}

Imagine teaching a robot a new household chore the way you'd teach a person: demonstrate it once or twice, and the robot performs it without further training on new objects, in completely new homes. Humans learn this way constantly, generalizing from imperfect, under-specified and context-dependent demonstrations \cite{carpenter_2yrold_priors,metlzoff_intent_18mnth, carpenter_intentional_accidental}, yet this kind of learning remains beyond the bounds of current robot capabilities. The dominant paradigm for learning manipulation from demonstration, behavior cloning (BC) \cite{Torabi2018BehavioralCF, argall2009survey, osa2018algorithmic}, requires hundreds or thousands of demonstrations per task and can still degrade under minor variations in the problem setting \cite{ross10a,de2019causal,spencer2021feedback, jang2022bcz}. A new layout, a different object set, added clutter, or a changed geometric constraint can break a policy that was near-perfect in the training scene distribution. Few-shot imitation, where only 1-3 demonstrations are available, is exactly where this brittleness is most acute. 

In this paper, we argue that the central obstacle to few-shot imitation is not a lack of demonstrations, but a lack of abstraction. A demonstration shows \textit{what} a person did in a particular scene, but not directly \textit{why} they did it. The same task can give rise to many different action sequences depending on the objects, layout, and constraints of the environment. And conversely, the same action sequence might be executed in service of many different tasks. As a result, simply cloning an observed trajectory distribution is unlikely to generalize beyond the training environments. To learn from only a handful of demonstrations, a robot must instead infer the underlying intent: which objects were relevant, what intermediate objectives were being pursued, what ordering relationships mattered, and which constraints were essential. Achieving real generalization doesn't come from replaying past behavior, but from recovering this abstracted representation and using it to synthesize behavior for a new scene. This shifts the problem from learning a mapping from states to actions, toward inferring a structured and transferable representation of task intent. 

% Programmatic representations have recently emerged as an effective way to connect high-level task instructions with robot execution. We build on the same representational insight, but reverse the direction of inference: the robot is not given instructions, only demonstrations. Here, we represent task intent as an explanation program $E$: a compact, executable function that maps an object-centric scene representation to a structured task specification, including the goals, subgoals and continuous constraints that define successful execution. Under this representation, demonstrations are treated as evidence about the latent program that generated them. Few-shot imitation can then be framed as reasoning: inferring the program under which the demonstrator's behavior appears most rational. Once a likely program is recovered, the robot need not reproduce the original trajectory. Instead, it can apply the program to a new scene, re-ground the relevant objects and constraints, and use a planning system to generate an appropriate new behavior, far beyond the training distribution. \ben{potentially very unlike any of its training examples}

Programmatic representations have recently emerged as an effective way to connect high-level task instructions with robot execution \cite{Liang2022CodeAP, saycan2022arxiv, huang2022inner, driess2023palme, curtis2024trust, silver2024generalized}. We build on the same representational insight, but reverse the direction of inference: the robot is not given instructions, only demonstrations. Here, we represent task intent as an \textit{explanation program}: a compact, executable function that maps an object-centric scene representation to a structured task and motion planning (TAMP) \cite{Garrett2020IntegratedTA} specification; a composition of goals, ordered subgoals and continuous constraints that define successful execution. The program is not a low-level controller. Instead, for each problem, it instantiates a planning problem, which the TAMP system solves to produce a feasible motion. Few-shot imitation can then be framed as: inferring the program under which the demonstrator's behavior appears most rational. Once a likely program is recovered, the robot need not reproduce the original trajectory. Instead, the same program is applied to a new scene, re-grounding the relevant objects and constraints, and the planning system is used to generate an appropriate new behavior that may differ substantially from the demonstrations, while preserving their inferred intent. 
% \ben{potentially very unlike any of its training examples}

The challenge here is to recover the most likely explanation program. Explaining behavior by inverting a model of rational action is the focus of inverse reinforcement learning and inverse planning \cite{ziebart2008maximum, bakerActionUnderstandingInverse2009, zhi-xuanOnlineBayesianGoal2020, ZhiXuan2024InfiniteEF}; an observer assumes that an agent acts approximately rationally (optimally) with respect to some objective, and then infers the objective that best explains the observed behavior. We adopt this principle, but the robotics setting extends the problem in three essential ways. First, the space of programs is much larger than the simple symbolic goal labels or reward functions that prior approaches consider \cite{ziebart2008maximum, bakerActionUnderstandingInverse2009, ZhiXuan2024InfiniteEF, jacob2024modeling, Mirsky2021}, posing a hard search problem. Second, with such an expressive hypothesis class, many programs can describe the same demonstration, making this search degenerate without a prior. Third, the likelihood required by classical inverse planning is computationally intractable, as scoring candidates requires solving many long-horizon, continuous and geometrically constrained planning problems \cite{Garrett2020IntegratedTA}. 

In this paper, we introduce\textbf{ Rational Inverse Reasoning (RIR)}, a few-shot imitation learning framework with the rich hypothesis space of explanation programs. Given a broadly capable planning and control architecture, RIR infers explanation programs that \textbf{rationalize} the demonstrations by coupling VLM-generated programs with planner-grounded evaluation. Given 1-3 demonstrations, a vision-language model proposes candidate explanation programs. A hierarchical task-and-motion planning system then scores these programs. This procedure is made tractable by using a GPU-accelerated surrogate for the inverse-planning likelihood: evaluating whether the proposed hypothesis makes the demonstrated behavior \textit{rational}, \textit{consistent} and \textit{specific} enough to make the demonstration likely in comparison to plausible alternatives. The resulting planner-grounded feedback drives an iterative reasoning loop. The VLM is used to adaptively re-propose candidate program hypotheses. Repeating this score-conditioned proposal loop improves the finite-pool posterior over programs without exhaustive program search. Through this loop, RIR lifts few-shot imitation from trajectory matching to approximate Bayesian inference over explanation programs. Through planning, we can then directly generalize to substantially different scenes. 

% \ben{does scene sound too narrow?}

We evaluate RIR in (i) a 2D benchmark suite that isolates intent inference from noise and partial observability, and (ii) diverse, long-horizon real-robot manipulation tasks. In both domains, we show few-shot generalization to novel objects, layouts, and constraints. 

% \ben{Compared to strong object-centric imitation and VLM-planning baselines etc etc - something to add when we have baseline comparisons.}

In summary, our contributions are:
\begin{itemize}
    \item We formulate few-shot learning from demonstration as inference over executable explanation programs: scene-conditioned representations of task intent that compile to planning specifications and support interpretation, re-grounding, and re-planning.
    \item We introduce a GPU-accelerated planner-grounded surrogate likelihood for rational inverse planning in continuous manipulation, combining evidence from strategy rationality, execution feasibility, and constraint specificity to make intent inference tractable.
    \item We present an iterative VLM-planner inference loop in which a vision-language model proposes open-ended program hypotheses and the planner-based likelihood guides their iterative improvement, learning from only 1-3 demonstrations.
    \item We evaluate RIR end-to-end on a Franka FR3, demonstrating few-shot generalization across novel scenes, improving success rates over VLM-planning by $34\%$ in the one-shot case, and $28\%$ in the 3-shot case. 
\end{itemize}

% \ben{Needs some work to find relevant citations for all the stuff we've talked about in the introduction}

\section{Related Work}

Our work connects three main threads: (i) inverse planning and goal recognition, (ii) foundation models for robot planning and control and (iii) few-shot program induction.

\subsection{Inverse planning, IRL and goal recognition}

Inverse reinforcement learning (IRL) methods recover a reward function from demonstrations under which the observed behavior is approximately optimal \cite{ziebart2008maximum, ho2016generative, ng2000algorithms}. Bayesian inverse planning instead models observed behavior as the output of a (boundedly) rational planner, inverting this process to infer latent intent~\cite{bakerActionUnderstandingInverse2009,zhi-xuanOnlineBayesianGoal2020,ZhiXuan2024InfiniteEF,alanqary2021modeling,jacob2024modeling}. Similarly, plan recognition methods reduce goal inference to optimal planning over a candidate goal set \cite{Mirsky2021, RamirezGeffner2009, GoalRecognitionSurvey2021, Vered2017}. These Bayesian approaches share our use of \emph{model-based likelihoods} grounded in planning, but they typically consider only a finite set, or simple symbolic goal candidates. RIR instead infers open-ended \emph{python programs} that take in a full object-centric scene graph representation and compile to express subgoal structure, ordering conventions, and continuous constraints. In addition, we evaluate hypotheses by inverting a TAMP~\cite{Garrett2020IntegratedTA} pipeline, modeling bounded rationality at both the discrete (task) and geometric (trajectory) levels.

\subsection{Foundation models for planning and learning}

LLMs and VLMs are often used to propose code policies, skill sequences, or symbolic plans that are then grounded by affordances, value functions, or execution feedback~\cite{Liang2022CodeAP,saycan2022arxiv,huang2022inner,driess2023palme}. Other work embeds foundation models \emph{inside} task-and-motion planning, e.g., to infer open-world constraints, propose partial skeletons, or repair motion failures within a TAMP loop~\cite{curtis2024trust,kumar2024open,wang2024llm3, silver2024generalized}. In contrast to approaches that \emph{augment} the TAMP stack itself (e.g. learning predicates, operators or generating specific constraint parameters \cite{silverPredicateInventionBilevel2023, athalye2024pixels, curtis2024trust}), RIR uses a VLM as a semantic \emph{proposal prior} for explanation programs, with a fixed TAMP API and domain theory, and supplies an external likelihood to guide scoring and refinement of candidate explanations. 

\subsection{Few-Shot Program Synthesis.}

Program synthesis and programming-by-example methods infer programs from a small example set via DSL search guided by learned priors~\cite{ellisDreamCoderGrowingGeneralizable2023}. Robotics analogs synthesize interpretable (often control-flow-rich) programs from demonstrations~\cite{patton2024programming}. Recently, programming-by-example has been dominated by in-context learning with coding agents. Having unit tests, a verified compiler, and the ability to generate multiple candidate solutions \cite{li2022competition, chen2023codet, yang2024sweagent} has pushed these systems into mainstream use. Here, RIR targets continuous, long-horizon manipulation, using model-based planning as the feedback system: it induces programs whose denotation is a TAMP task specification, yielding a re-plannable, geometrically feasible representation of task intent rather than a closed-world controller or a high-level piece of code not grounded in physical execution.

\section{Problem Setting} \label{sec:problem_setting}

In this section, we outline the problem setting, defining our inference target, \emph{explanation programs}, and the robot's model-based perception and planning capacity, TAMP.

\subsection{Problem Definition}

\begin{figure}[t]
    \centering
    \includegraphics[width=\linewidth]{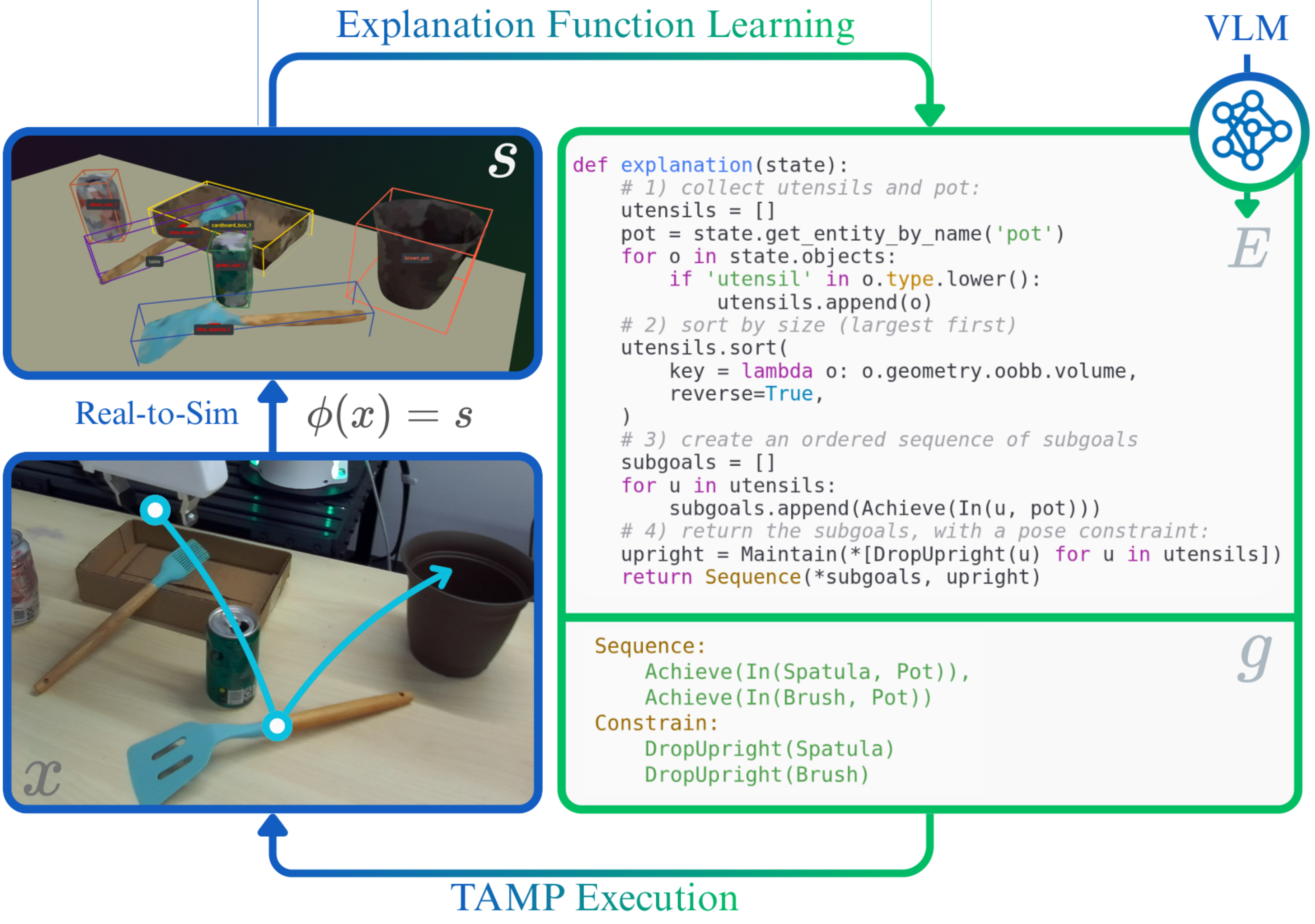}
    \caption{A candidate explanation program structure (\textit{details reduced for brevity}) and compilation of the following task: \emph{put the utensils in the pot upright, in size order}. For this particular scene, there are two utensils: the spatula and the brush. Notably, the learner \textit{never sees this language description}, only 3 demonstrations of the task, and must infer $E$.}
    \label{fig:problem-setting-ref}
\end{figure}

At training time, we are given a dataset of demonstrations $D = \{(\tau^{(k)}, x_0^{(k)})\}_{k=1}^K$ with small $K$ for sample efficient learning. Each demonstration $\tau^{(k)}=(x_0^{(k)},\dots,x_T^{(k)})$ consists of raw robot observations $x_t$: RGBD camera views and robot joint configurations. Although demonstrations may have distinct initial scenes (layouts, objects, clutter), we assume they all communicate a shared underlying intent that can be represented by a single explanation program $E^*$, in the space of possible explanation programs $\mathcal{E}$. At test time, the robot is presented with a novel initial scene observation $x_0'$. Then, using the inferred intent $\hat E$, the robot must produce an action sequence whose induced trajectory $\hat \tau$  satisfies user's true intent, $\hat \tau \models E^*(\phi(x'_0))$. This problem setting, from demonstrations to explanation programs to behavior, is highlighted in \cref{fig:problem-setting-ref}.

\subsection{Representations for Perception and Behavior}

\subsubsection{Object-Centric Scene Representation}

To encapsulate the perception, scene reconstruction and tracking pipelines, we define a reconstruction function $\phi : \mathcal{X} \rightarrow \mathcal{S}$ that produces an object-centric scene graph representation $s \in \mathcal{S}$. We write $s = (O, R)$, where $O = \{o_i\}_{i=1}^n$ are the objects/entities with attributes (e.g. type, semantic features, keypoints, 6D pose, mesh geometry), and $R$ is a set of ground instances of primitive spatial properties and relations on objects, $V$. Our overall approach can accommodate arbitrary relations in $V$ as long as they are testable by the perception pipeline; in our implementation, $V$ consists of the set of base geometric atoms \textit{in, on, holding}, as well as adaptive local pose and grasp constraints, such as \textit{upright}, flexibly parameterized by keypoint relations on the object, similar to \citet{huang2024copa}. An example scene reconstruction result is shown in \cref{fig:problem-setting-ref}. 

% \textbf{Task-Specifications $\mathcal{G}$:}~~To act as a shared contract/API between the VLM and the planning system, we leverage a small but expressive task specification language $\mathcal{G}$, allowing us to define goals, subgoals and constraints, as well as compositions thereof, using three constructors
% $$g ::= \texttt{Ach}(\Phi) \mid \texttt{Seq}(g_1,\dots,g_m) \mid \texttt{Maintain}(c, g'),$$
% defined as follows:
% \begin{itemize}
% \item $\texttt{Ach}(\Phi)$: Conjunction of ground predicates, all satisfied
%     \item $\texttt{Seq}(g_1,\dots,g_m)$: Rollout decomposes into consecutive segments satisfying specs $g_1, ..., g_m$ in order
%     \item $\texttt{Maintain}(c, g)$: satisfy $g$ while maintaining constraint $c$ over the segment.
% \end{itemize}
% For example, consider the task of placing the utensils in the pot upright, starting with the spatula, then the brush. This task specification $g$ is shown in \cref{fig:problem-setting-ref}. 

\subsubsection{Task-Specifications $\mathcal{G}$}

Given a new scene, we must ultimately produce robot behavior. To generate such behavior, the planning system takes the object-centric scene state $s$ as input, as well as a \textit{task specification} $g$. The task specification language $\mathcal{G}$ is a small but expressive subset of linear temporal logic (LTL) \cite{4567924}, allowing us to define goals, subgoals and constraints, as well as compositions thereof, using three constructors
$$g ::= \texttt{Ach}(\Phi) \mid \texttt{Seq}(g_1,\dots,g_m) \mid \texttt{Constrain}(c, g),$$
defined as follows:
\begin{itemize}
\item $\texttt{Ach}(\Phi)$: Conjunction of ground predicates, elements of $V$, all satisfied
    \item $\texttt{Seq}(g_1,\dots,g_m)$: Trajectory decomposes into consecutive segments satisfying specs $g_1, ..., g_m$ in order
    \item $\texttt{Constrain}(c, g)$: satisfy $g$ while maintaining constraint $c$ over the segment.
\end{itemize}
For example, consider the task of placing the utensils in the pot upright, starting with the spatula, then the brush. This task specification $g$ is shown in \cref{fig:problem-setting-ref}. 

\subsubsection{Explanation Programs $E : \mathcal{S} \rightarrow \mathcal{G}$}

Explanation programs are open-ended Python functions over the object-centric scene state which produce a grounded, executable task specification: $g^{(k)}=E(\phi(x_0^{(k)})) \in \mathcal{G}$. That is, they are constrained to output a valid task specification for the planner. Python is sufficiently expressive to filter and count objects, branch on scene properties, construct derived features, goals, and constraints, and invoke a VLM for additional semantic grounding (e.g. to identify \textit{fluffy} objects).

% Python is highly flexible, and can apply filters, counting, branching, extract and define derived features/goals/constraints, as well as write its own VLM calls for additional semantic grounding (e.g. find me the 'fluffy' objects).

For example, consider the explanation program in \cref{fig:problem-setting-ref}, describing the general task to \textit{put the utensils in the pot upright, in size order}. The explanation program constructs the sub-goal and constraint structure by filtering the object-centric state representation for entities which offer a semantic and geometric match to these descriptions. Consequently, we can apply the same explanation program to a wide variety of scenes while preserving intended behavior.

\subsection{Generalized TAMP as a Stochastic Forward Model}

To produce action sequences from high-level intent, we use a generalized
Task-and-Motion Planning (TAMP) algorithm
\cite{Garrett2020IntegratedTA}. The planner takes as input a task
specification $g \in \mathcal{G}$ and an object-centric scene
representation $s_0$, and produces a symbolic plan skeleton (the high-level \textit{strategy}) $\hat{\pi}$-an ordered sequence of high-level operations grounded in the objects in the scene-and an environment trajectory $\tau$ involving the robot and objects in
the scene. TAMP integrates discrete task reasoning with continuous
feasibility checking over grasps, placements, and collision-free
motions. Importantly, additional structure in $g$ does not require an
external interpreter: it directly constrains both the space of feasible
plan skeletons and the continuous refinement problems constructed by
the solver. Algorithm~\ref{alg:tamp} summarizes this procedure. Because skeleton search, continuous sampling, heuristic search, and
finite compute budgets may all introduce nondeterminism, we treat TAMP
as a conditional generative model:
\begin{equation}
    (\hat{\pi}, \tau) \sim \mathrm{TAMP}(g, s_0).
\end{equation}

\begin{algorithm}[t]
\caption{TAMP with Structured Specifications}
\label{alg:tamp}
\KwIn{Initial abstract state $s_0$, specification $g$, compute budget $B$}
\KwOut{Plan skeleton $\hat{\pi}$ and trajectory $\tau$, or failure}

\[
g \;\stackrel{(1)}{\longrightarrow}\; \hat{\pi}
\;\stackrel{(2)}{\longrightarrow}\; \tau
\;\stackrel{(3)}{\circlearrowleft}
\]

\While{$B$ is not exhausted}{
    $\hat{\pi} \gets \textsc{SkeletonSearch}(g,s_0)$\;
    \tcp{\footnotesize \textbf{(1)} Choose operators/discrete variables $\footnotesize{\hat\pi=(\small{\texttt{pick(cup)}},\texttt{place(cup, table)}, \dots)}$ consistent with the
    goals, subgoals, and constraints in $g$.}

    \If{$\hat{\pi}=\bot$}{
        \Return failure\;
    }

    $\tau \gets \textsc{Refine}(\hat{\pi},g,s_0)$\;
    \tcp{\footnotesize \textbf{(2)} Solve for grasps, placements, and collision-free motions
    satisfying constraints from both $\hat{\pi}$ and $g$.}

    \If{$\tau\neq\bot$}{
        \Return $(\hat{\pi},\tau)$\;
    }

    $\textsc{ReviseOrResample}(\hat{\pi})$\;
    \tcp{\footnotesize \textbf{(3)} On failure, revise the skeleton and/or resample continuous
    parameters.}
}

\Return failure\;

\BlankLine
\textbf{Stochastic interpretation:}\quad
$(\hat{\pi},\tau)\sim\mathrm{TAMP}(g,s_0)$.
\end{algorithm}

% \begin{algorithm}[t]
% \caption{TAMP with Structured Specifications}
% \label{alg:tamp}

% \SetKwFunction{SkeletonSearch}{SkeletonSearch}
% \SetKwFunction{Refine}{ContinuousRefinement}
% \SetKwFunction{Revise}{ReviseSearch}

% \KwIn{Initial abstract state $s_0$ and structured specification $g$}
% \KwOut{Plan skeleton $\hat{\pi}$ and feasible trajectory $\tau$, or failure}

% Initialize the symbolic search and continuous sampling state\;

% \While{the compute budget is not exhausted}{
%     $\hat{\pi} \gets \SkeletonSearch(g, s_0)$
%     \tcp*[r]{Search over operators and discrete assignments}

%     \If{$\hat{\pi} = \bot$}{
%         \Return failure
%     }

%     $\tau \gets \Refine(\hat{\pi}, g, s_0)$
%     \tcp*[r]{Solve for grasps, placements, and motion paths}

%     \If{$\tau \neq \bot$}{
%         \Return $(\hat{\pi}, \tau)$
%     }

%     \Revise($\hat{\pi}$)
%     \tcp*[r]{Revise the skeleton and/or resample parameters}
% }

% \Return failure\;
% \end{algorithm}

% \textbf{Scope:}~~We do not learn low-level motor skills from scratch within this representation. Instead, we infer explanation programs that produce reusable, planner-compatible task specifications, leveraging an existing TAMP system. Extensions that expand the planning system are discussed in \cref{sec:limitations}.

The TAMP algorithm is general-purpose and can in principle include non-prehensile operations such as pushing, pouring and throwing. In our implementation, we limit operation set to picking, moving and placing. RIR applies over whatever domain of operations have been implemented in the TAMP system, and does not learn any additional low-level motor skills. Instead, we infer explanation programs that produce reusable, planner-compatible task specifications, leveraging an existing TAMP system. Extensions that expand the planning system are discussed in \cref{sec:limitations}.

%%%%%%%%%%%%%%%%%%%%%%%%%%%%%%%%%%%%%%%%%%%%%%%%%%%%%%%%%%%%%%%

\section{Rational Inverse Reasoning} \label{sec:rir}

Rational Inverse Reasoning (RIR) casts few-shot learning as the inference of explanation programs $E$ from demonstration data $D$, by inverting a near-optimal embodied reasoning procedure for inferring and executing robot motions. In this section, we formalize this inference objective. 

Given a small dataset of demonstrations $D=\{(\tau^{(k)},x_0^{(k)})\}_{k=1}^K$ RIR aims to infer a distribution over explanation programs $E$, which has the form
\begin{equation}
     p(E\mid D) \propto p(E)\prod_{k=1}^K p(\tau^{(k)} \mid x_0^{(k)},E), 
     \label{eqn:central-posterior}
\end{equation}
assuming conditional independence of the demonstration trajectories $\tau$ given the explanation program $E$ (our representation of demonstrator intent). Three challenges arise in the inference process:
\begin{enumerate}
    \item Devising a tractable likelihood score $p(D \mid E)$ that measures how well a candidate $E$ explains $D$ in the presence of suboptimal execution and continuous state and action spaces. 
    \item Encoding a prior $p(E)$ that captures the human common-sense probability of various objectives in the world.
    \item Finding computationally efficient mechanisms for capturing (and maximizing) the posterior $p(E \mid D)$ when the space of explanation programs $\mathcal{E}$ is infinite and planning for evaluating $p(D \mid E)$ is computationally complex.
\end{enumerate}
We address these challenges by (1) extending the notion of bounded rationality to our hierarchical model of physical reasoning in a tractable manner; (2) using a large vision language model as a proposal distribution and taking advantage of its ability to write code as output; and (3) combining top-down and bottom-up proposals in an iterative refinement inference scheme. 

Sections \ref{sec:main-hierarchical-model}--\ref{sec:gpu-parallel-sec} develop a tractable approximation to the likelihood term in \cref{eqn:central-posterior}, and \cref{sec:iterative-reasoning} describes the adaptive program search used to approximate the resulting posterior over explanation programs.

\subsection{Boltzmann All The Way Down: The Hierarchical Generative Model of Behavior}
\label{sec:main-hierarchical-model}

For one demonstration, a program $E$ deterministically compiles the reconstructed initial scene $s_0= \phi(x_0)$ into a structured task specification $g = E(s_0)$. The planner first selects a symbolic plan skeleton $\hat \pi \in \Pi(g, s_0)$, where $\Pi(g, s_0)$ is the (typically infinite) set of feasible plans for $g$ and initial state $s_0$. Then, the planner samples continuous refinement parameters such as grasps, placements, configurations and motion plans formulating a generated trajectory $\tilde \tau$. Treating $E$ as a deterministic map between the initial state $s_0$ and the task specification $g$, the behavioral likelihood marginalizes out the unknown plan $\hat \pi$:  
\begin{equation}
    p(\tau\mid x_0,E)
    =
    \sum_{\hat\pi\in\Pi(g,s_0)}
    p_{\mathrm{traj}}(\tau\mid\hat\pi,g,s_0) \,
    p_{\mathrm{plan}}(\hat\pi\mid g,s_0).
    \label{eq:main-hierarchical-likelihood}
\end{equation}
Here, $p_{\mathrm{traj}}$ captures execution rationality: the likelihood of $\tau$ as a continuous refinement of a particular skeleton $\hat \pi$. The factor $p_{\mathrm{plan}}$ captures the strategy rationality: the probability of selecting the plan $\hat \pi$ for a specification $g$ and initial state $s_0$. 

\subsubsection{Execution Rationality}

Let \(q(\tau\mid\hat\pi, g, s_0)\) be the planner's normalized proposal distribution over complete refinements of a skeleton into trajectory $\tau$.  Let \(\mathbf{1}(\tau;\hat\pi,g)\in\{0,1\}\) indicate that a trajectory is geometrically feasible, adheres to \(\hat\pi\), and satisfies the goals and constraints in \(g\).  We choose to model the error in execution with a Boltzmann distribution on skeleton refinement,
\begin{align}
    p_{\mathrm{traj}}(\tau\mid\hat\pi,g,s_0)
    =
    \frac{
        q(\tau\mid\hat\pi, g,s_0)
        e^{[-\beta_{\mathrm{traj}}C_{\mathrm{traj}}(\tau)]}
        \mathbf{1}(\tau;\hat\pi,g)
    }{
        Z_{\mathrm{traj}}(\hat\pi,g,s_0)
    }, \nonumber \\
    \label{eq:main-execution-boltzmann}
\end{align}
\begin{align}
    Z_{\mathrm{traj}}(\hat\pi,g,s_0)
    =
    \mathbb E_{\tau \sim q(\cdot\mid\hat\pi, g,s_0)}
    \left[
        e^{[-\beta_{\mathrm{traj}}C_{\mathrm{traj}}(\tau)]}
        \mathbf{1}(\tau;\hat\pi,g)
    \right]. \nonumber \\
    \label{eq:main-execution-partition}
\end{align}
This cost $C_{\mathrm{traj}}$ may include path length, control effort or task-dependent risk (i.e. $C_{\mathrm{traj}}(\cdot ; \hat \pi, g)$ - for brevity we write $C_{\mathrm{traj}}$). The precise terms specify which kinds of execution suboptimality the model expects: the cost should be high for trajectories that are inefficient or otherwise unlikely. The inverse temperature \(\beta_{\mathrm{traj}}\) controls how strongly the trajectory distribution concentrates around low-cost refinements.  

The partition function $Z_{\mathrm{traj}}$ is also evidence about the higher level plan $\hat \pi$. It is large when $\hat \pi$ admits a broad collection of feasible, low-cost executions, and small when success requires rare or costly continuous choices. We summarize this evidence by the execution free energy,
\begin{equation}
    F_{\mathrm{traj}}(\hat\pi;g,s_0)
    =
    -\frac{1}{\beta_{\mathrm{traj}}}
    \log Z_{\mathrm{traj}}(\hat\pi,g,s_0).
    \label{eq:main-execution-free-energy}
\end{equation}
Intuitively, free energy is a plan's \textit{robust cost}: lower values mean that there are many near-optimal ways to execute it.

\subsubsection{Strategy Rationality} \label{sec:strat-rat}

We model the demonstrator's choice of plan skeleton $\hat \pi$ \textit{(strategy)} according to a second Boltzmann distribution,
\begin{equation}
    p_{\mathrm{plan}}(\hat\pi\mid g,s_0)
    =
    \frac{
        \exp[-\beta_{\mathrm{plan}}
             F_{\mathrm{traj}}(\hat\pi;g,s_0)]
    }{
        \displaystyle
        \sum_{\hat\pi'\in\Pi(g,s_0)}
        \exp[-\beta_{\mathrm{plan}}
             F_{\mathrm{traj}}(\hat\pi';g,s_0)]
    }.
    \label{eq:main-plan-boltzmann}
\end{equation}
The two temperatures have distinct roles: \(\beta_{\mathrm{traj}}\) controls trajectory variability within a plan, while \(\beta_{\mathrm{plan}}\) controls how optimally (given a goal) strategies are selected. To evaluate a skeleton, we use free energy,  rather than the cost of its cheapest refinement $C_{\mathrm{min}}$, modeling a human demonstrator's tendency to prefer strategies that are robustly executable, rather than those with one fortuitous refinement.

\begin{figure}[t]
    \centering
    \includegraphics[width=0.9\linewidth]{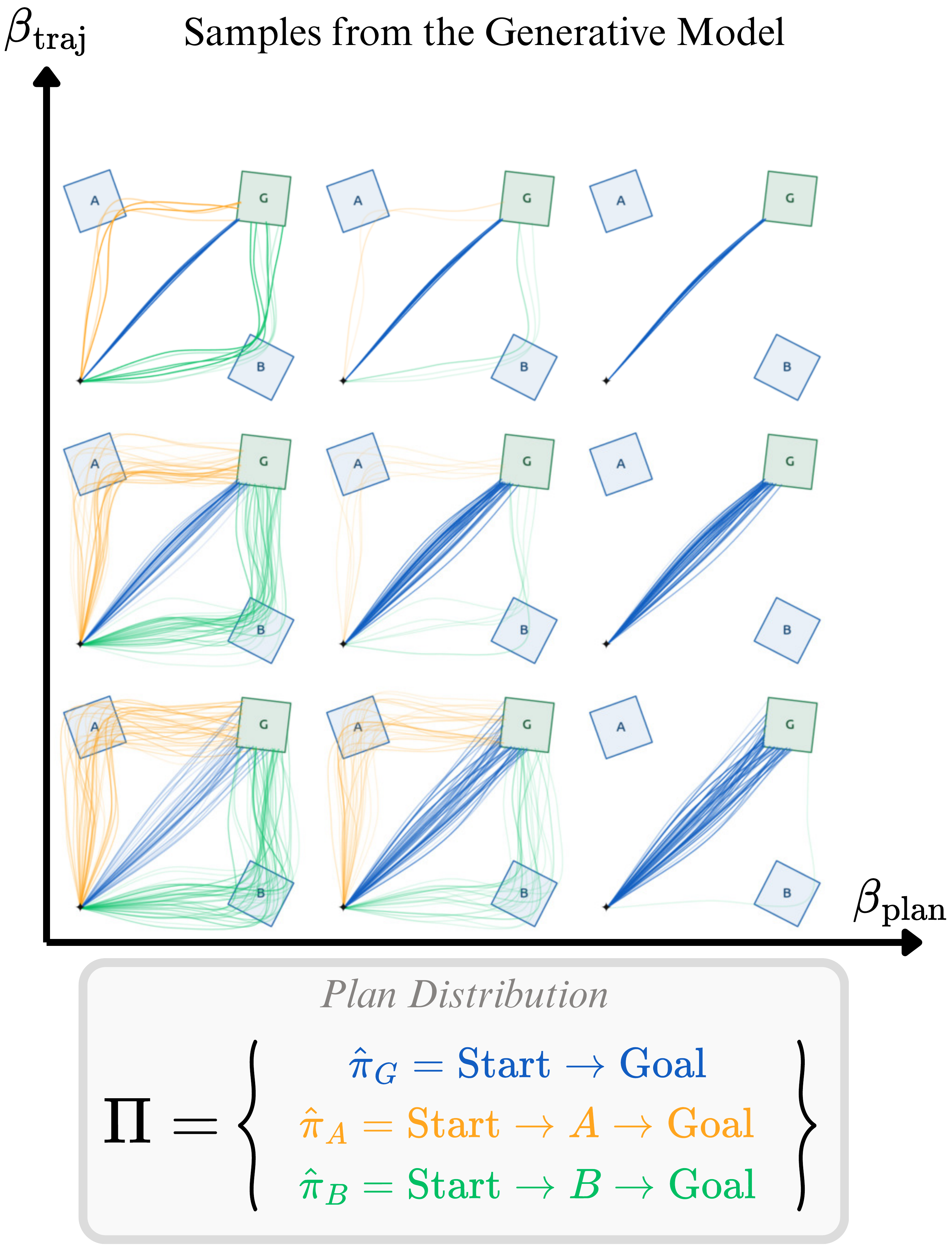}
    \caption{\textbf{Samples from the hierarchical Boltzmann model.} Increasing $\beta_{\mathrm{plan}}$ concentrates probability on lower-free-energy plan skeletons, while increasing $\beta_{\mathrm{traj}}$ concentrates each skeleton's refinements on lower-cost trajectories. Colors denote the three intermediate-region strategies.} 
    \label{fig:generative-model-samples}
\end{figure}

Unbiased samples from this hierarchical Boltzmann distribution of behavior are shown in \cref{fig:generative-model-samples}. In this simplified 2D environment, there are three possible plan skeletons: $|\Pi| = 3$, two of which passing through one intermediary region before moving towards the goal. From \cref{fig:generative-model-samples}, we can see the effect of these two inverse temperatures. As $\beta_{\mathrm{plan}}$ increases, we move from uniform plan selection across $\Pi$, to focusing primarily on the most efficient plan (with the lowest refinement free energy) $\hat \pi_G$. As $\beta_{\mathrm{traj}}$ increases, we see that the refinement distribution concentrates on the minimum-cost trajectories--here, the cost is defined in terms of path length.

\subsection{A Tractable Planner-Grounded Likelihood}
\label{sec:main-tractable-likelihood}

Our goal is to evaluate $p(\tau| x_0, E)$ (\cref{eq:main-hierarchical-likelihood}) for observed demonstration trajectories across sets of candidate explanation programs $E$. However, the model above is not directly computable: the planner cannot enumerate all skeletons or integrate over all continuous refinements. RIR replaces it with a finite surrogate that preserves three sources of evidence: (i) compatibility of the task and plan with the demonstration, (ii) rationality of the demonstrated (observed) plan relative to counterfactuals, and (iii) the cost and specificity of the observed execution.

\subsubsection{Symbolic Support and Demonstration Consistent Skeletons} \label{sec:symbolic-support}

For an observed trajectory, the indicator $\mathbf{1}(\tau;\hat\pi,g)$ tests whether the demonstration $\tau$ satisfies $g$, and can be segmented as $\hat \pi$. This defines the support for any plan-conditional likelihood contribution. 

Applying $\phi$, our reconstruction function, to each observation $x_t$ in the demonstration trajectory yields a symbolic trace $\phi(\tau) = (R_0, ..., R_{T_k})$; where $R_t$ is the set of true ground atoms from vocabulary $V$ in frame $t$. Based on this, we can evaluate $\mathbf{1}(\tau;\hat\pi,g)$ with two efficient verifiers: $\text{ValidDemo}(\tau,g)\in\{0,1\}$ which determines whether the trace satisfies the goals/ordering/constraints of $g$, and $\text{Consistent}(\tau,\hat\pi)\in\{0,1\}$ which determines whether the trace is a valid segmentation of skeleton $\hat\pi$. These can be computed efficiently based on the task planning model using dynamic programming, akin to LTL segmentation approaches \cite{10.1145/3641513.3650180, puranic2021learning}. 

\begin{figure}
    \centering
    \includegraphics[width=1.0\linewidth]{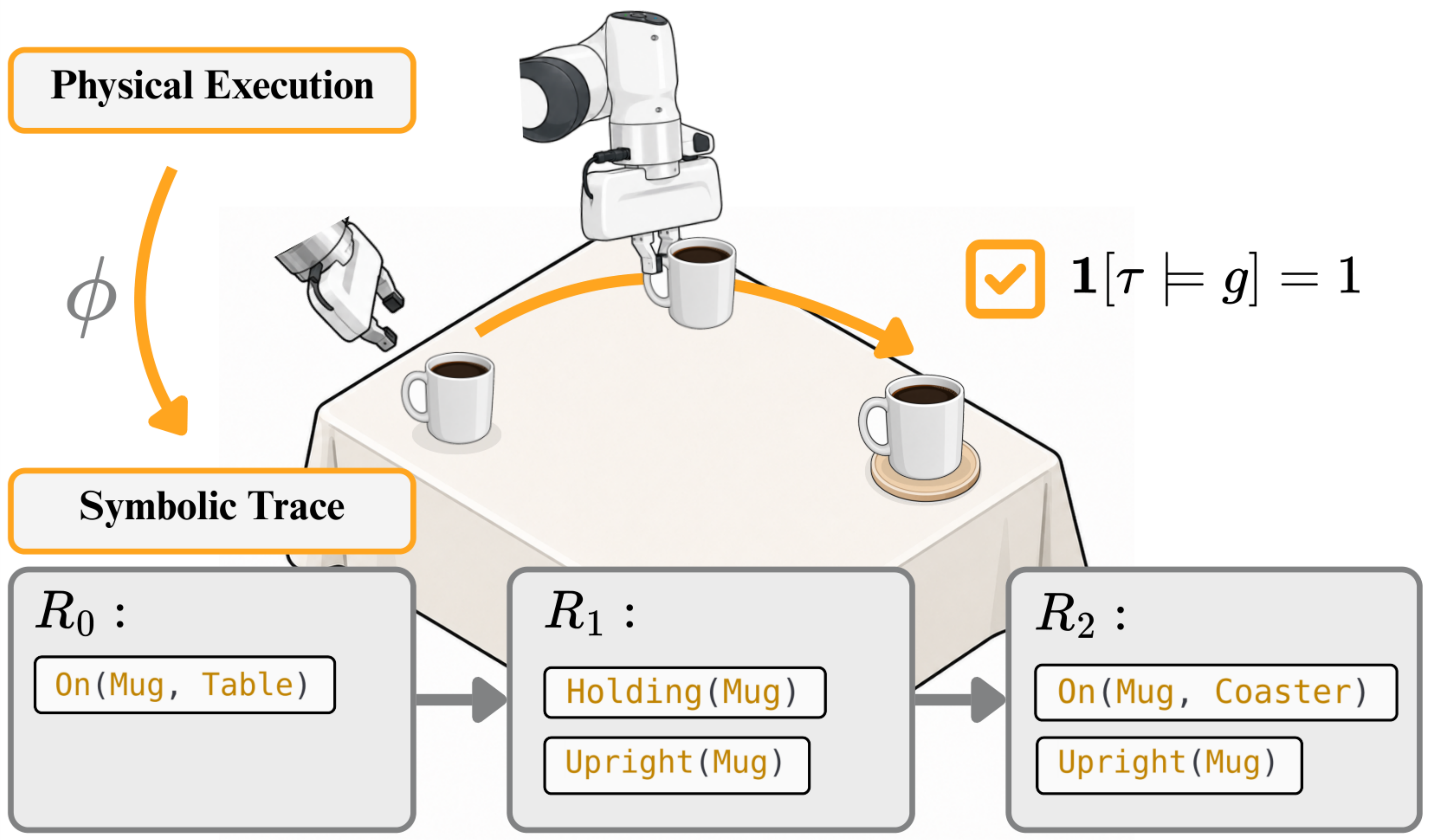}
    \caption{\textbf{From physical execution to symbolic evidence.} Reconstruction maps the demonstrated trajectory to a predicate trace, which is checked for satisfaction of the candidate task and consistency with a candidate plan skeleton.}
    \label{fig:sym-trace-viz}
\end{figure}

Often, this segmentation procedure produces multiple valid candidates. For example, consider the $\hat \pi_A$ trajectories in \cref{fig:generative-model-samples}. All these trajectories are consistent with $\hat \pi_G$, as they reach the goal region. The result is a set of \textit{bottom-up} plan proposals $\widehat\Pi_{\mathrm{bu}}(E)$ which have non-zero support in the generative model associated with the task specification $g$ and plan skeleton $\hat \pi$,
\begin{align}
p(\tau|\hat \pi, g) \propto e^{-\beta_{\mathrm{traj}}C(\tau)}\cdot \text{ValidDemo}(\tau,g)\cdot \text{Consistent}(\tau,\hat\pi). \nonumber \\
\label{eq:indicators-boltzmann-weighted}
\end{align}

For example, consider the demonstration shown in \cref{fig:sym-trace-viz}. Here, we can extract the symbolic trace from the demonstration, and verify it against the candidate task specification $g$, and plan $\hat \pi$. This is valid and consistent for $g = \texttt{Achieve(On(Mug, Coaster))}$, with $\hat \pi = \texttt{[PickPlace(Mug, Coaster)]}$ placing the mug on the coaster. 

\subsubsection{Demonstrated and Counterfactual Strategies} \label{sec:demo-counterfactuals}

For candidate program $E$, we construct a finite plan skeleton set
\begin{equation}
    \widehat\Pi(E)
    =
    \widehat\Pi_{\mathrm{bu}}(E)
    \cup
    \widehat\Pi_{\mathrm{td}}(E).
    \label{eq:main-finite-skeletons}
\end{equation}
Whilst the bottom-up set $\widehat{\Pi}_{\mathrm{bu}}$ is recovered from the demonstration, as noted in \cref{sec:symbolic-support}, the \textit{top-down} set $\widehat{\Pi}_{\mathrm{td}}$ is proposed by the planner to satisfy $g$. That is, given $g$, we sample a candidate set of high-level plan skeletons using classical task-planning search \cite{Garrett2020IntegratedTA}. This additional set contains alternative strategies that could have satisfied the same task specification but were not observed. These \textit{counterfactual strategies} are essential: A demonstrated plan is informative only relative to what else the demonstrator could rationally have done. Concretely, under the assumption of a rational agent, a detour is evidence for an intended subgoal only when a more direct plan would otherwise have been optimal.

% \ben{remind to investigate}

\subsubsection{Planner-Generated Refinement Evidence} \label{sec:plan-gen-ref}

For each \(\hat\pi\in\widehat\Pi(E)\), we draw \(N\) refinements \(\tau_j\sim q(\cdot\mid\hat\pi, g,s_0)\), and compute their costs \(c_j = C_{\mathrm{traj}}(\tilde \tau_j)\) and validity masks \(\mathbf{1}(\tilde \tau_j; \hat\pi,g_k)\). The refinement free energy is estimated by
\begin{equation}
    \widehat F=-\frac{1}{\beta_{\mathrm{traj}}}\log \left[
    \frac{1}{N}
    \sum_{j=1}^{N}
    \exp^{[-\beta_{\mathrm{traj}}c_j]}\,
    \mathbf{1}(\tilde \tau_j; \hat\pi,g) \right],
    \label{eq:main-mc-partition}
\end{equation}
a Monte-Carlo estimate relying on the proposal distribution $q$. We compute the log-sum-exp (LSE) in a numerically stable manner, treating invalid samples as having log-weight \(-\infty\). Using \cref{eq:main-plan-boltzmann} over the finite skeleton and refinement set, we also obtain our approximation of \(\widehat p_{\mathrm{plan}}(\hat\pi\mid g,s_0)\), the second component of \cref{eq:main-hierarchical-likelihood}. The use of free energy rather than the minimum sampled refinement cost $C_{min} = \inf_{j} C(\tilde \tau_j)$ also makes the estimator more robust to stochasticity introduced by resource-bounded planning.
\begin{figure}[t]
    \centering
    \includegraphics[width=1.0\linewidth]{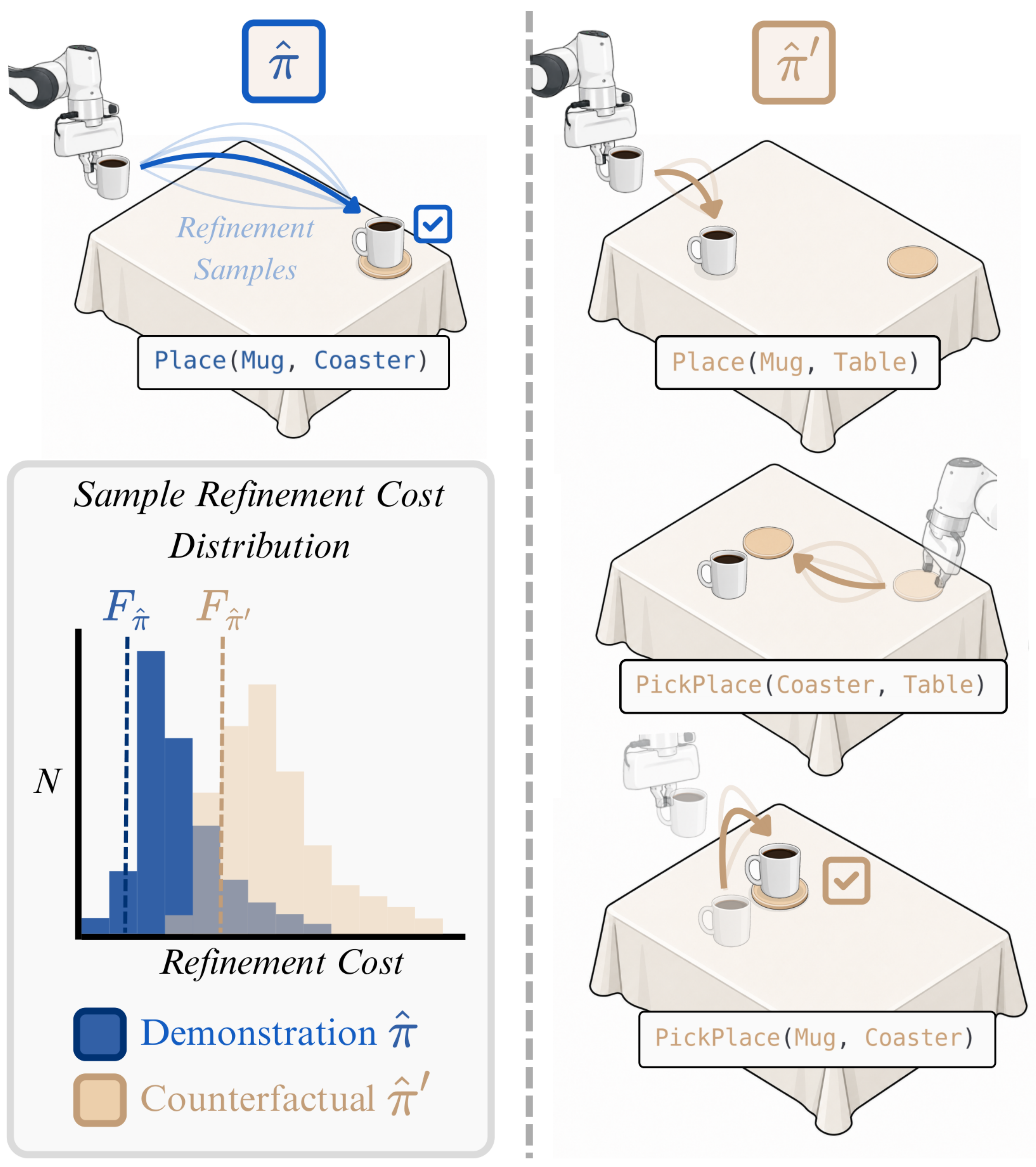}
    \caption{\textbf{Planner-grounded comparison of demonstrated and counterfactual strategies.} GPU-sampled refinement costs estimate each skeleton's free energy, comparing the observed direct placement with an alternative that first moves the coaster.}
    \label{fig:strat-rat-exp}
\end{figure}

A visualization of this procedure for computing $p_{\mathrm{plan}}$ via the free energy $\widehat F$ is shown in \cref{fig:strat-rat-exp}. Here the sub-task of \textit{placing the mug down on the coaster} admits two strategies: put the mug directly down on the coaster, $\hat \pi \in \Pi_{\mathrm{bu}}$, and a sampled counterfactual alternative $\hat \pi' \in \Pi_{\mathrm{td}}$ that pulled the coaster close first, then placed the mug down. In both cases, we construct a sample refinement set, and from there compute the free energy. Based on these free energies, the direct placement appears more rational, however, if we added additional spillage risk terms to the cost, it might be that dragging the coaster over is more rational. 

\subsection{Relative Specificity from Nested TAMP Proposals}

\begin{figure}[t]
    \centering
    \includegraphics[width=\linewidth]
        {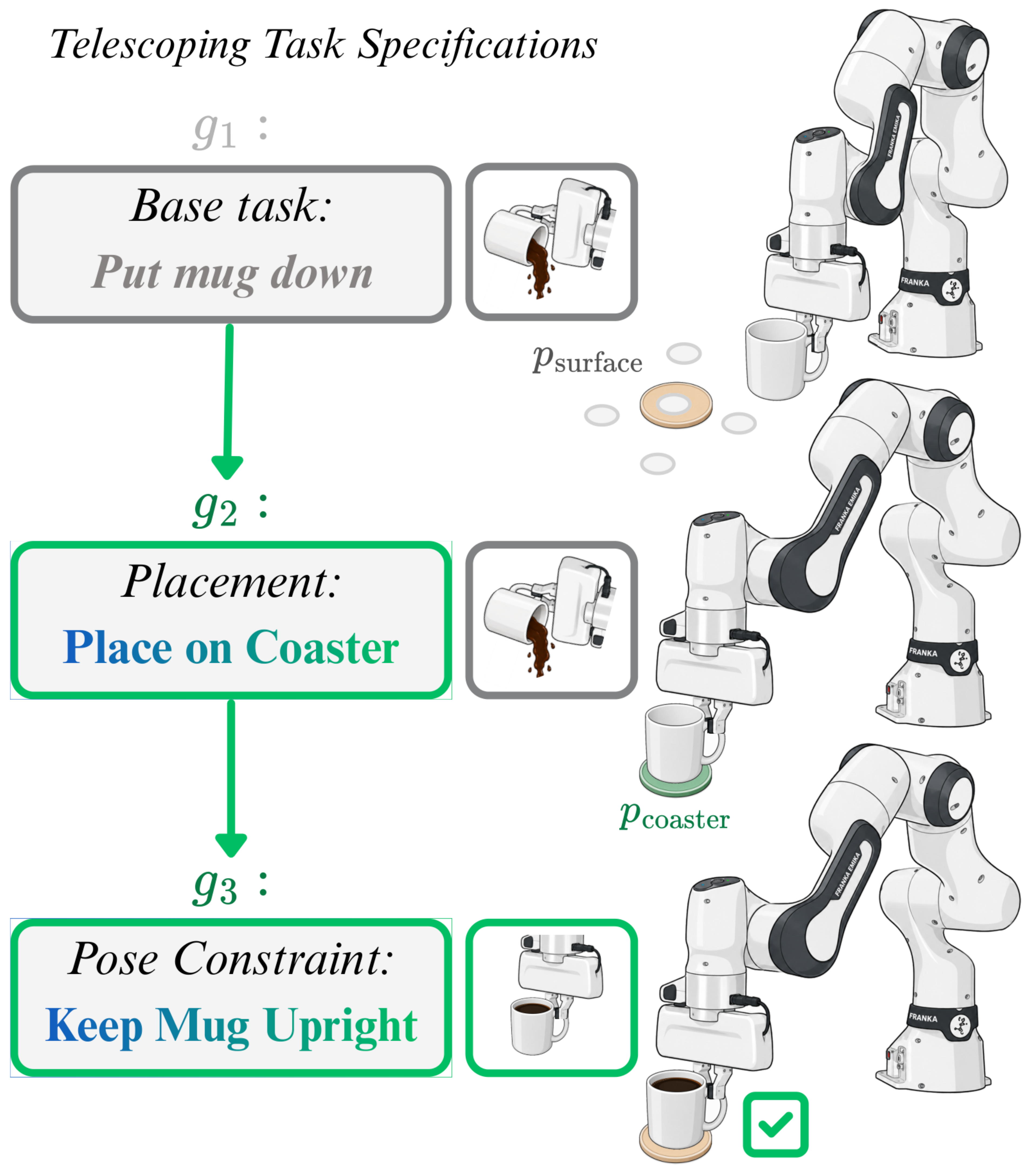}
    \caption{\textbf{Relative specificity from nested TAMP proposals.}
    Refinement banks for progressively stronger specifications estimate
    the fraction of executions that also satisfy the next constraint:
    placing the mug down, placing it on the coaster, and maintaining
    uprightness.}
    \label{fig:spec-const-2}
\end{figure}

A candidate task specification $g$ shouldn't just make the demonstration feasible, it should also narrow the set of plausible refinements. Consider \cref{fig:spec-const-2}: Under $g_2$, which requires only \textit{placing the mug on the coaster}, the fraction of low-cost refinements that also happen to keep the mug upright may be very small. Thus, the uprightness requirement in $g_3$ removes almost all of the execution mass available under $g_2$ and should provide strong evidence in favor of this more specific explanation if we observe it. In other words, a trajectory that keeps the mug upright would be very surprising under the weaker alternative hypothesis. This rarity is not reflected in the normalizer estimate for $g_3$ via \cref{eq:main-execution-partition}. Under a finite computational budget, we sample refinements using a planner \textit{already} conditioned on $g_3$, which deliberately proposes upright placements. The resulting refinement distribution therefore makes $g_3$ appear easy, even though it retains almost none of the execution mass available under $g_2$.

Ideally, we would compare all specifications under a shared weak proposal $q_0$. For a specification $g_\ell$, define its cost-weighted execution mass as
\begin{equation} 
Z_{\ell}^{(0)} = \mathbb{E}_{\widetilde{\tau}\sim q_0} \left[ \exp\!\left( -\beta_{\mathrm{traj}} C_{\mathrm{traj}}(\widetilde{\tau}) \right) \mathbf{1} \!\left( \widetilde{\tau}; \hat{\pi}, g_\ell \right) \right]. 
\label{eq:fixed_proposal_normalizer} 
\end{equation}
The ratio $Z_{\ell+1}^{(0)}/Z_{\ell}^{(0)}$ then measures the fraction of low-cost execution mass that remains after adding the requirements in $g_{\ell+1}$. Directly estimating the ratio between a weak and a highly specific task is difficult: with finitely many samples, their supports may barely overlap, giving us degenerate estimate. We therefore construct a nested sequence 
$
g_0 \preceq g_1 \preceq \cdots \preceq g_L = g, \label{eq:nested_task_specs} 
$
where each step weakens ordering relations in \texttt{Seq}, or drops \texttt{Constrain} or \texttt{Achieve} conditions. The desired endpoint ratio can then be decomposed into local comparisons:
\begin{equation} 
\frac{Z_L^{(0)}}{Z_0^{(0)}} = \prod_{\ell=0}^{L-1} \frac{Z_{\ell+1}^{(0)}}{Z_{\ell}^{(0)}}. \label{eq:specificity_telescoping} 
\end{equation}
In Fig.~\ref{fig:spec-const-2}, these bridges correspond to \emph{put the mug down}, \emph{place it on the coaster}, and \emph{keep it upright}. Adjacent tasks typically have enough overlap to measure the effect of each added requirement, even when the weakest task contains no samples satisfying the complete specification. In practice, we estimate each adjacent ratio using the refinement set generated under the weaker specification $g_\ell$:
\begin{equation}
    \widetilde r_{\ell,\ell+1}(\hat\pi)
    =
    \frac{
        \sum_{j=1}^{N}
        e^{-\beta_{\mathrm{traj}}
        C_{\mathrm{traj}}(\tilde\tau_{\ell j})}
        \mathbf 1(\tilde\tau_{\ell j};\hat\pi,g_{\ell+1})
    }{
        \sum_{j=1}^{N}
        e^{-\beta_{\mathrm{traj}}
        C_{\mathrm{traj}}(\tilde\tau_{\ell j})}
        \mathbf 1(\tilde\tau_{\ell j};\hat\pi,g_\ell)
    }.
    \label{eq:adjacent_specificity_estimate}
\end{equation}
We combine the answers across the bridge sequence into the specificity correction 
\begin{equation} 
\widetilde{\Delta}_{\mathrm{spec}}(\hat{\pi}) = -\sum_{\ell=0}^{L-1} \log \widetilde{r}_{\ell,\ell+1}(\hat{\pi}). 
\label{eq:specificity_correction} 
\end{equation}
An added specification element contributes strongly when it removes most otherwise plausible low-cost executions, and weakly when it adds little behavioral information. We use this correction to approximate the normalizer of the complete specification:
\begin{equation} 
-\log Z_L^{(0)} \approx -\log \widehat{Z}_0^{(0)} + \widetilde{\Delta}_{\mathrm{spec}}. 
\label{eq:corrected_specificity_normalizer} 
\end{equation}
The reference term is then omitted as it is shared by all hypotheses being compared. Because our TAMP proposals are specification-conditioned and their densities are unavailable, the adjacent ratios are an algorithm-dependent approximation rather than exact bridge-sampling estimates \cite{gelman}. 

% Further details, bias considerations, bridge construction, and fallback conditions are given in Appendix~\ref{app:specificity_derivation}.

\subsection{GPU-Parallel Refinement Evidence} \label{sec:gpu-parallel-sec}

As noted in \cref{sec:plan-gen-ref}, evaluating the surrogate likelihood requires drawing many continuous refinements $\tilde{\tau}_j$ for both demonstrated and counterfactual plan skeletons. This is more demanding than ordinary forward planning. A conventional TAMP system may terminate after finding a \textit{single} feasible refinement \cite{Garrett2020IntegratedTA}, whereas RIR requires a \textit{collection} of feasible refinements whose costs and validity characterize the distribution of executions available under each plan.

% With $K$ demonstrations, $M$ candidate explanation programs, multiple skeletons per grounded task specification, and $N$ refinements per skeleton, a naive implementation would require a larger number of long-horizon continuous planning problems in at inference round.

We make this computation tractable by exploiting the common computational structures inherent in continuous refinement. This allows us to instantiate many refinement problems simultaneously and batch components such as grasp generation, inverse-kinematics queries, trajectory optimization, collision checking, constraint evaluation, and cost computation on GPU hardware, following the general approach enabled by recent GPU-accelerated TAMP and motion-generation systems \cite{shen2024cutamp,sundaralingam2023curobo}. Rather than launching a separate planning process for every sample, the implementation evaluates large collections of candidate refinement variables and constraints through shared, GPU-resident operations. GPU batching converts the required Monte-Carlo evidence from a sequence of expensive planner calls into a smaller number of large parallel refinement batches.

The generated samples are stored in a \textit{refinement bank} $
\mathcal{B}_{k,g,\hat \pi, j}
=
\bigl(\tilde{\tau}_j,
C_{\mathrm{traj}}(\tilde{\tau}_j)
\bigr),
$ indexed by the refinement index $j \in 1, ..., N$ demonstration $k$, the plan skeleton $\hat{\pi}$, and the specification $g$ used to generate the refinement proposal. Candidate programs $E$ that compile to the same grounded planning problem $g$ therefore share the same bank, even when their source code or linguistic descriptions differ. The cache is preserved across hypotheses and inference rounds, \textit{amortizing continuous refinement} over the iterative VLM-planner search.

The algorithm in Alg.~\ref{alg:planner-grounded-likelihood} summarizes the computation of the planner-grounded surrogate likelihood for one explanation program and demonstration. \textsc{BottomUp} extracts the demonstration-consistent skeletons, while \textsc{TAMP-Plans} supplies counterfactual skeletons that could also satisfy $g$. For each skeleton and bridge specification, \textsc{RefineBank} generates the $N$ continuous refinements described in \cref{sec:plan-gen-ref}; \textsc{Collapse} reduces these banks over their refinement indices to obtain the execution normalizer, free energy, and relative-specificity estimate. The full skeleton set determines $\widehat p_{\mathrm{plan}}$, whereas the final log-sum-exp marginalizes only over skeletons consistent with the observed demonstration $\widehat \Pi_{\mathrm{bu}}$.

\begin{algorithm}[t]
\caption{Inverse Reasoning Likelihood (Planner-Grounded Surrogate)}
\label{alg:planner-grounded-likelihood}
\small

\KwIn{Demonstration $\tau$; scene $s$; explanation program $E$;
       TAMP; refinement budget $N$}
\KwOut{Surrogate log-likelihood $\widehat\ell(E)$}

$g \gets E(s)$\;

\If{$\neg\mathrm{ValidDemo}(\tau,g)$}{
    \Return{$-\infty$}\;
}

$\widehat\Pi_{\mathrm{bu}}
\gets
\textsc{BottomUp}(\tau,g)$\;

$\widehat\Pi_{\mathrm{td}}
\gets
\textsc{TAMP-Plans}(g,s)$\;

$\widehat\Pi
\gets
\widehat\Pi_{\mathrm{bu}}
\cup
\widehat\Pi_{\mathrm{td}}$\;

$(g_0,\ldots,g_L)
\gets
\textsc{BridgeSpecs}(g)$,
with $g_L=g$\;

\ForEach{$\hat\pi\in\widehat\Pi$}{
    \For{$\ell\gets0$ \KwTo $L$}{
        $\mathcal B_{\ell,\hat\pi}
        \gets
        \textsc{RefineBank}(s,\hat\pi,g_\ell,N)$\;
    }

    $(\widehat Z_{\hat\pi},
      \widehat F_{\hat\pi},
      \widetilde\Delta_{\mathrm{spec},\hat\pi})
    \gets
    \textsc{Collapse}
    (\{\mathcal B_{\ell,\hat\pi}\}_{\ell=0}^{L})$\;
}

$\widehat p_{\mathrm{plan}}(\hat\pi\mid g,s)
\gets
\operatorname{softmax}_{\hat\pi\in\widehat\Pi}
\left(
-\beta_{\mathrm{plan}}\widehat F_{\hat\pi}
\right)$\;

$\widehat\ell(E)
\gets
\operatorname{LSE}_{\hat\pi\in\widehat\Pi_{\mathrm{bu}}}
\left[
    \log\widehat p_{\mathrm{plan}}(\hat\pi\mid g,s)
    -
    \beta_{\mathrm{traj}}C_{\mathrm{traj}}(\tau)
    -
    \log\widehat Z_{\hat\pi}
\right]$\;

\Return{$\widehat\ell(E)$}\;
\end{algorithm}

\subsection{Iterative Particle-Based Reasoning} \label{sec:iterative-reasoning}

\subsubsection{VLM program proposal}

The surrogate in Alg.~\ref{alg:planner-grounded-likelihood} makes it possible to score a candidate explanation, but the hypothesis space $\mathcal{E}$ consists of open-ended Python programs and cannot be enumerated. Instead, RIR uses a VLM as an adaptive proposal mechanism and maintains a finite pool of weighted candidate programs $\mathcal{H} = \{(E_i, w_i)\}_{i=1}^M$ . The VLM contributes a strong semantic prior over plausible human objectives and useful program structure; the planner-grounded score determines which of those proposals are supported by the demonstrations. 

\subsubsection{Particle search over programs}

RIR then runs an iterative particle search. The VLM is used to propose the initial $M$ candidates. This first round ($t=1$) is primarily top-down; the VLM proposes broad explanations conditioned on context containing the demonstrations, scene reconstructions, task API and in-context examples $E \sim \text{VLM}(E | \text{context}_t)$. Each candidate is scored by the surrogate likelihood, and the weights $w_i$ normalized across the candidate set $\mathcal{H}$. In the following rounds we add the scored program set $\mathcal{H}_t$ to the VLM context, along with planner diagnostics (which constraints or subgoals failed, whether its plan strategy score is low, whether its specificity ratio is near 1). The VLM is then tasked with re-proposing a refined candidate set $\mathcal{H}_{t+1}$, with the aim of preserving supported structure, while repairing specific sources of surprise - erroneous programs, low-weight hypotheses and sub-goal or constraint failures. High-weight particles are retained across rounds, and the remaining slots are filled with revisions and diverse new proposals. After the final round, RIR reports the maximum-a-posteriori (MAP) program over the finite pool (or the top-K). The RIR algorithm is summarized in Alg.~\ref{alg:main-rir}. 

The VLM proposal is not a formal Markov kernel, so the loop does not perform principled rejuvenation or re-weighting; it is best understood as iterative finite-pool posterior scoring with adaptive VLM proposals (a particle-style posterior search over programs) similar to \citet{piriyakulkij2024doing}. Principled rejuvenation (e.g. MCMC moves on programs \cite{lewSMCP3SequentialMonte2023, schkufza2013stochastic}) is future work. 

By combining language-model priors with demonstration-based rational feedback, this iterative refinement mechanism allows the system to bootstrap from a generic prior to a structured and task-relevant explanation posterior, searching over an infinite hypothesis space of open-ended task programs. In a novel scene, the selected program is executed to produce a new grounded task specification, and planning synthesizes behavior for that scene.  The robot does not replay, warp, or geometrically imitate a training trajectory; it re-plans from the inferred intent.

% \begin{algorithm}[t]
% \caption{Rational Inverse Reasoning}
% \label{alg:main-rir}
% \small

% \KwIn{Demonstrations $D$; reconstruction $\phi$; TAMP; VLM;
%        pool size $M$; rounds $T$; refinement budget $N$}
% \KwOut{Weighted program pool and MAP explanation $\hat E$}

% $\{s_k\}_{k=1}^{K}\gets\textsc{Reconstruct}(D,\phi)$\;
% $\mathcal C\gets\textsc{VLM-Propose}(D,\{s_k\},M)$\;
% Initialize shared skeleton and refinement cache $\mathcal H$\;

% \For{$t\gets1$ \KwTo $T$}{
%     \ForEach{$E_i\in\mathcal C$}{
%         \eIf{$E_i$ fails to compile or produce valid specifications}{
%             $S_i\gets-\infty$; record diagnostics\;
%         }{
%             $S_i\gets\log p_{\mathrm{simp}}(E_i)$\;
%             \For{$k\gets1$ \KwTo $K$}{
%                 $g_{ik}\gets E_i(s_k)$\;
%                 $\widehat\Pi_{ik}\gets
%                 \textsc{BottomUp}(\tau^{(k)},g_{ik})
%                 \cup
%                 \textsc{TAMP-Plans}(g_{ik},s_k)$\;
%                 $(\widehat\ell_k,d_{ik})
%                 \gets
%                 \textsc{Likelihood}
%                 (\tau^{(k)},g_{ik},\widehat\Pi_{ik},N,\mathcal H)$\;
%                 $S_i\gets S_i+\widehat\ell_k$\;
%             }
%         }
%     }

%     $w_i\gets\operatorname{softmax}_i(S_i)$\;

%     \If{$t<T$}{
%         $\mathcal C\gets
%         \textsc{VLM-Refine}
%         (\mathcal C,\{w_i\},\{d_{ik}\},M)$\;
%         Compile, canonicalize, and deduplicate proposals\;
%     }
% }

% $\hat E\gets\arg\max_{E_i\in\mathcal C}S_i$\;
% \Return{$\mathcal C,\{w_i\},\hat E$}\;
% \end{algorithm}

\begin{algorithm}[t]
\caption{Rational Inverse Reasoning}
\label{alg:main-rir}
\small

\KwIn{Demonstrations $D=\{(\tau^{(k)},x_0^{(k)})\}_{k=1}^{K}$;
       reconstruction $\phi$; TAMP; VLM; pool size $M$;
       rounds $T$; refinement budget $N$}
\KwOut{Weighted program pool $\mathcal{H}$ and MAP explanation $\hat E$}

$s^{(k)}_0 \gets \phi(x_0^{(k)})$ for $k=1,\ldots,K$\;
$\mathcal H^{(1)}
\gets
\textsc{VLM-Propose}(D,\{s^{(k)}_0\}_{k=1}^{K},M)$\;

\For{$t\gets1$ \KwTo $T$}{
    \ForEach{$E_i\in\mathcal H^{(t)}$}{
        $S_i\gets\log p_{\mathrm{simp}}(E_i)$  \tcp{\footnotesize Optional MDL Prior} \; 

        \For{$k\gets1$ \KwTo $K$}{
            $S_i\gets S_i+
            \textsc{IR-Likelihood}
            (\tau^{(k)},s_0^{(k)},E_i,\mathrm{TAMP},N)$\;
        }
    }

    $w_i\gets\operatorname{softmax}_{E_i\in\mathcal H^{(t)}}(S_i)$\;

    \If{$t<T$}{
        $\mathcal H^{(t+1)}
        \gets
        \textsc{VLM-Refine}
        (\mathcal H^{(t)},\{w_i\},D,\{s_0^{(k)}\}_{k=1}^{K},M)$\;
    }
}

$\hat E
\gets
\arg\max_{E_i\in \mathcal{H}^{(T)}} S_i$\;

\Return{$\{(E_i,w_i)\}_{E_i\in\mathcal H^{(T)}},\hat E$}\;
\end{algorithm}

\section{Experiments}

We evaluate the efficacy of RIR in a 2D test environment, {\sc terc},
then in a full 3D robotic domain with a Franka FR3 robot manipulating multiple objects. In both cases, the system learns from a small number of human tele-operated demonstrations. The {\sc terc} domain allows us to carefully study the effects of design choices in RIR, without the noise and partial-observability of the real world. Based on these experiments, we configure an end-to-end real-robot system and show that RIR can generalize robustly from only a handful of demonstrations in physically grounded, complex real-world manipulation settings.

\subsection{Evaluation metrics}
We evaluate RIR at three levels (1) \textbf{Comprehension} (program equivalence) asks whether the inferred program matches the demonstrator's latent intent ($\hat E \equiv E^*$); it measures inference quality independent of any downstream planning. (2) \textbf{Behavioral success} asks whether the behavior the inferred program induces is correct; i.e, whether planning with $\hat E$ in a novel, held-out scene would, without execution noise and complex dynamics, satisfy the true intent $E^*$. (3) \textbf{Physical execution} success asks whether that behavior, when run on the physical robot (or rolled out within a sufficiently good simulation environment) with all real-world grasp, contact, controller and perception noise, achieves the task. These three are distinct and reported separately on purpose: a program can be correct yet planned behavior wrong (an imperfect re-grounding), and behavior can be correct in simulation yet fail physically (a slipped grasp). Each \textit{macro-task} (\textit{place the mug on the coaster}) is evaluated over a large pool of [algorithm seed $\times$ object set $\times$ object layout] combinations. 

\subsection{Tiny Embodied Reasoning Corpus}

We introduce a suite of challenging 2D object rearrangement tasks that require fine-grained geometric \textit{and} algorithmic reasoning, and nuanced inference through compositional task structure. 

\begin{figure}
    \centering
    \includegraphics[width=1.0\linewidth]{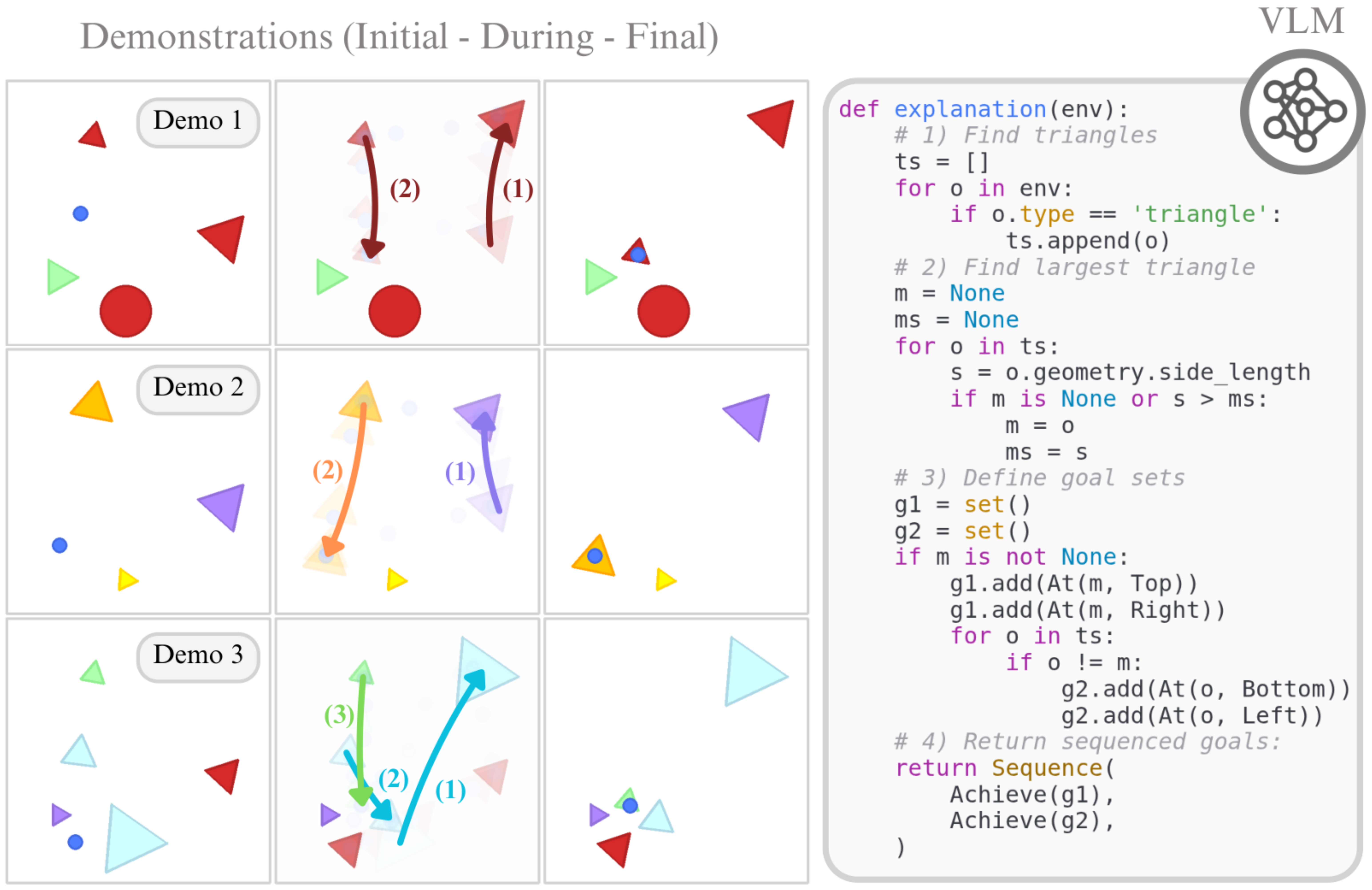}
    \caption{Example TERC task from the spatial+algorithmic subset. Here, the learner must infer, from 1-3 demonstrations, that it must \textit{place the largest triangle block at the top-right, then place all other triangle blocks at the bottom-left}. }
    \label{fig:terc-exmaple-1}
\end{figure}

\subsubsection{TERC Environment} 

We model a top-down tabletop manipulation environment in the Pymunk physics engine \cite{blomqvist2025pymunk}, similar to many common robotics benchmarks \cite{florence2022implicit, zeng2021transporter, shridhar2022cliport}. The agent can move anywhere and is able to slide objects around the environment while avoiding collisions. Reasoning tasks are expressed over object attributes, relations, and positions on the table. We evaluate our learned explanation programs $E$ in samples from this family. Notably, spatial predicates compose (\texttt{Top} $\wedge$ \texttt{Left} $\rightarrow$ \texttt{Top-Left}, resulting in three spatial predicates indicating the top, left and top-left regions of the environment), forcing the learner to choose among many competing hypotheses. An illustration of one of the TERC tasks is shown in \cref{fig:terc-exmaple-1}, requiring the agent to infer, from only 3 demonstrations, that the task is to \textit{place the largest triangle block at the top-right, then place all other triangle blocks at the bottom-left}. Here, even with this simple demonstration, we must disambiguate between many possible hypotheses, and produce a rich Python explanation program that compiles into the correct sorting behavior.

% \textbf{TERC Environment:}~~

\subsubsection{Task Distribution}

We define 35 unique tasks, corresponding to explanation programs, ranging from simple goal-reaching behaviors (\textit{Move all the red objects to the bottom-left corner}) to complex algorithmic reasoning tasks (\textit{If there is a triangle, move the circles to the top-right, otherwise move the boxes there}). We break down the $35$ macro-tasks into $25$ \textit{spatial} macro-tasks which mainly focus on physical reasoning (where objects moved, in what order etc.), and $10$ additional \textit{algorithmic} macro-tasks which add complex algorithmic reasoning on top of this (conditionals, complex filtering, counting, etc.).

\subsubsection{Approaches}
We compare the following approaches. 
\begin{itemize}
    \item Rational Inverse Reasoning (RIR): Our main approach.
    \item Reasoning without Rationality (for brevity: R w/o R): the VLM generates an explanation program using only self-refinement~\cite{madaan2023self, DBLP:conf/iclr/0002WSLCNCZ23}, and its own internal reasoning with no embodied feedback. However, it is prompted with the full scene reconstruction and prompted with additional bottom-up plan proposals.
    %\item Video-E: Inspired by \cite{Liang2022CodeAP}, the VLM generates an explanation program directly from a (set of) demonstration video(s) (as opposed to natural language), without rationalization/external planning feedback. 
    % \item NoPlan-NL: Based on the system outlined in \cref{...}, the VLM processes the demonstrations, and formulate a natural language description of the task. Then, at deployment, the VLM retrieves this natural language description and defines a goal specification, similar to \cite{silver2024generalized}. Here, the system doesn't have access to our programmatic task representation. 
    \item Human: We measured human performance (comprehension), by asking participants to formulate natural language explanations after viewing videos of the dataset demos. Further details are presented in Appendix~\ref{app:terc}. 
\end{itemize}

\subsubsection{Experimental setup}
For every experiment we sample fresh manipulation tasks from $\mathcal{D}_{\text{task}}$ in which the initial object poses, object attributes, and object counts are systematically varied. After conditioning on $K \in \{1, 3\}$ demonstrations, each method produces a (set of) learned explanation program(s) $E$ and a trajectory $\tau$ in the new environment(s). Each generated $E$ is tested in novel environments with  3 different object sets, each with 5 random pose initializations, using TAMP to solve for the robot trajectory. We also report top-$k \in \{1, 5, 10\}$ comprehension rate (top-$k$ is successful if the correct explanation is in the top-$k$ generated explanation programs).

\subsubsection{TERC Results}

\begin{figure}[t]
    \centering
    \includegraphics[width=1.0\linewidth]{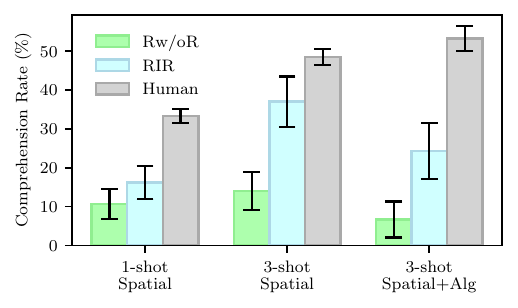}
    \caption{1- and 3- shot comprehension rate for the comparison approach (Rw/oR) and Humans, compared to RIR. Error bars denote the \% standard error (SE) for each method.}
    \label{fig:comprehension}
\end{figure}

\begin{figure}[t]
    \centering
    \includegraphics[width=1.0\linewidth]{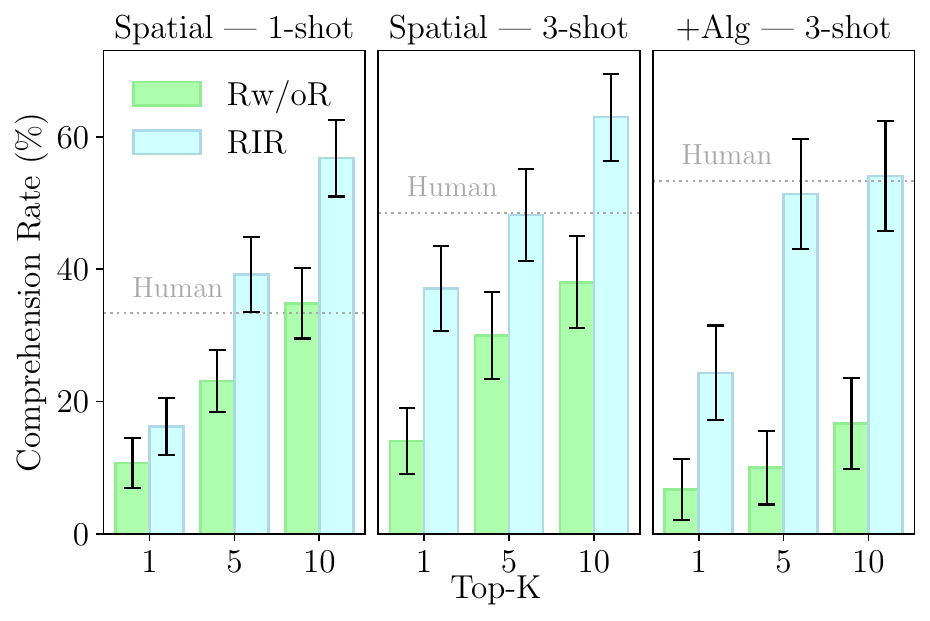}
    \caption{1- and 3- shot Top-K comprehension rates: The gray line indicates (Top-1) human performance. Error bars denote the \% standard error.}
    \label{fig:comprehension_topk}
\end{figure}

\begin{table}[t]
    
  \centering
  \setlength{\tabcolsep}{4pt}% tighten column spacing (optional)
  \caption{Top-1 \textit{behavioral} success rate (rollout in noiseless simulation therefore \textit{physical} success $\equiv$ \textit{behavioral} success) (\%) $\pm$ standard error (SE) for each method, difficulty, and \# of demonstrations. A trial is successful iff the behavior induced by the inferred explanation program satisfies the ground truth task $E^*$ on a held-out scene. }
  \label{tab:terc_succrate}
    \scriptsize

  \begin{tabular}{l c c c }
    \toprule
    \textbf{Method} & \textbf{Spatial (1-shot)} & \textbf{Spatial (3-shot)} & \textbf{Spatial+Alg (3-shot)}\\
    \midrule
    RIR   & \textbf{30.38} $\pm$ 2.50 & \textbf{58.87} $\pm$ 4.42 & \textbf{24.00} $\pm$ 4.27 \\
    R w/o R & 18.69 $\pm$ 2.23 & 22.44 $\pm$ 2.91 & 14.47 $\pm$ 4.04 \\
    $\Delta$ (RIR$-$Rw/oR) & $+11.7$ & $+36.4$ & $+9.5$ \\
    \bottomrule
  \end{tabular}
\end{table}

\paragraph{RIR improves comprehension of the latent explanation:}
In \cref{fig:comprehension} we see that RIR consistently outperforms RwoR, moving closer to human-level performance with more demonstrations. In contrast to the VLM-based approach which doesn't include rationalization steps during inference, the comprehension rate of RIR increases significantly from 1 to 3 demos. Notably, the human performance is far from perfect on these difficult tasks, and the 3-shot RIR performance is higher than 1-shot human performance.

\begin{figure}[b]
    \centering
    \includegraphics[width=\linewidth]{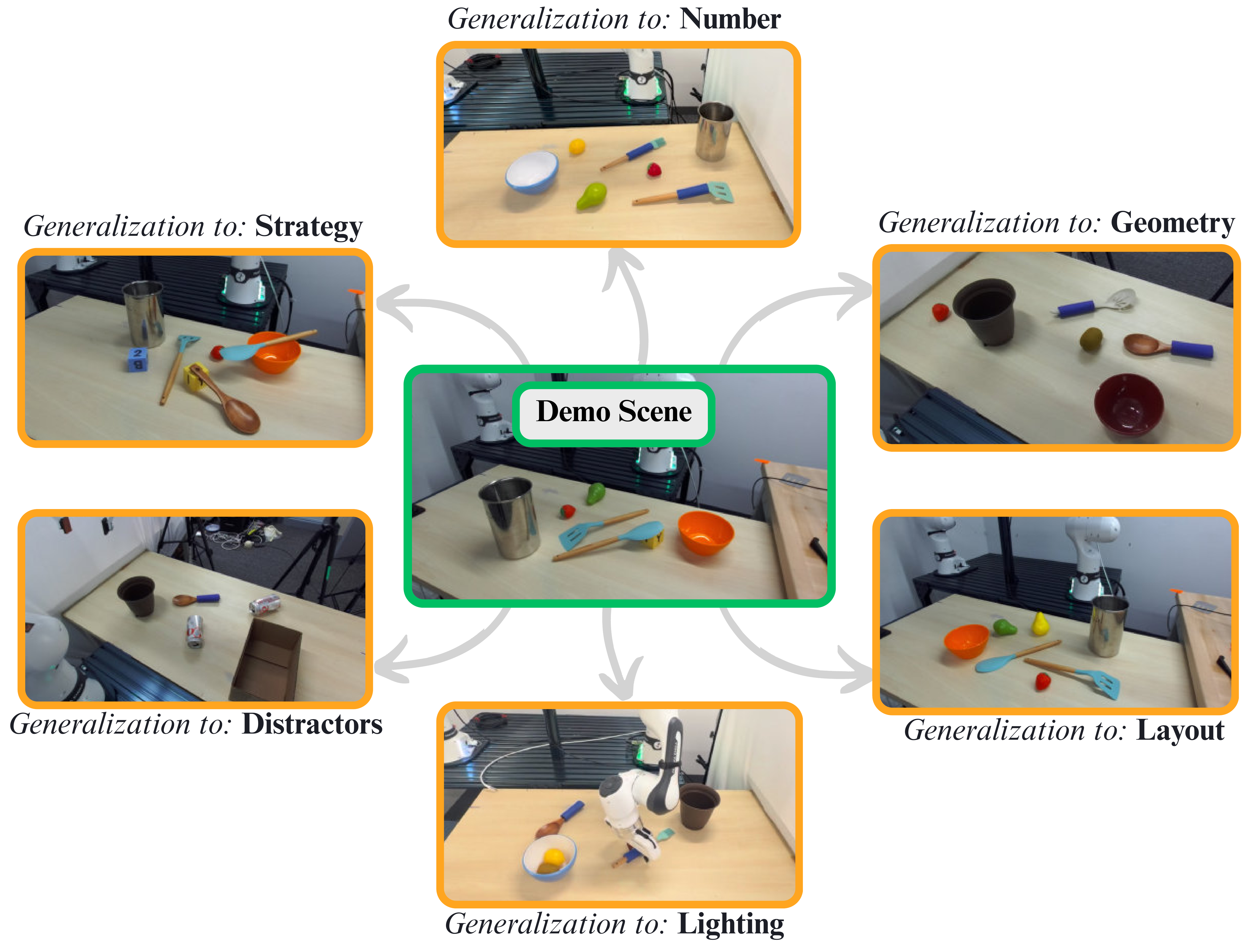}
    \caption{Generalization taxonomy of the RIR system. By leveraging the strength of the planning system, we're able to handle a huge variety of scenes once we've inferred the correct explanation program that describes the users intent, rationally explaining their behavior. }
    \label{fig:generalization-taxonomy}
\end{figure}

\begin{figure*}[t]
    \centering
    \begin{subfigure}[t]{0.49\linewidth}
        \centering
        \includegraphics[width=\linewidth]{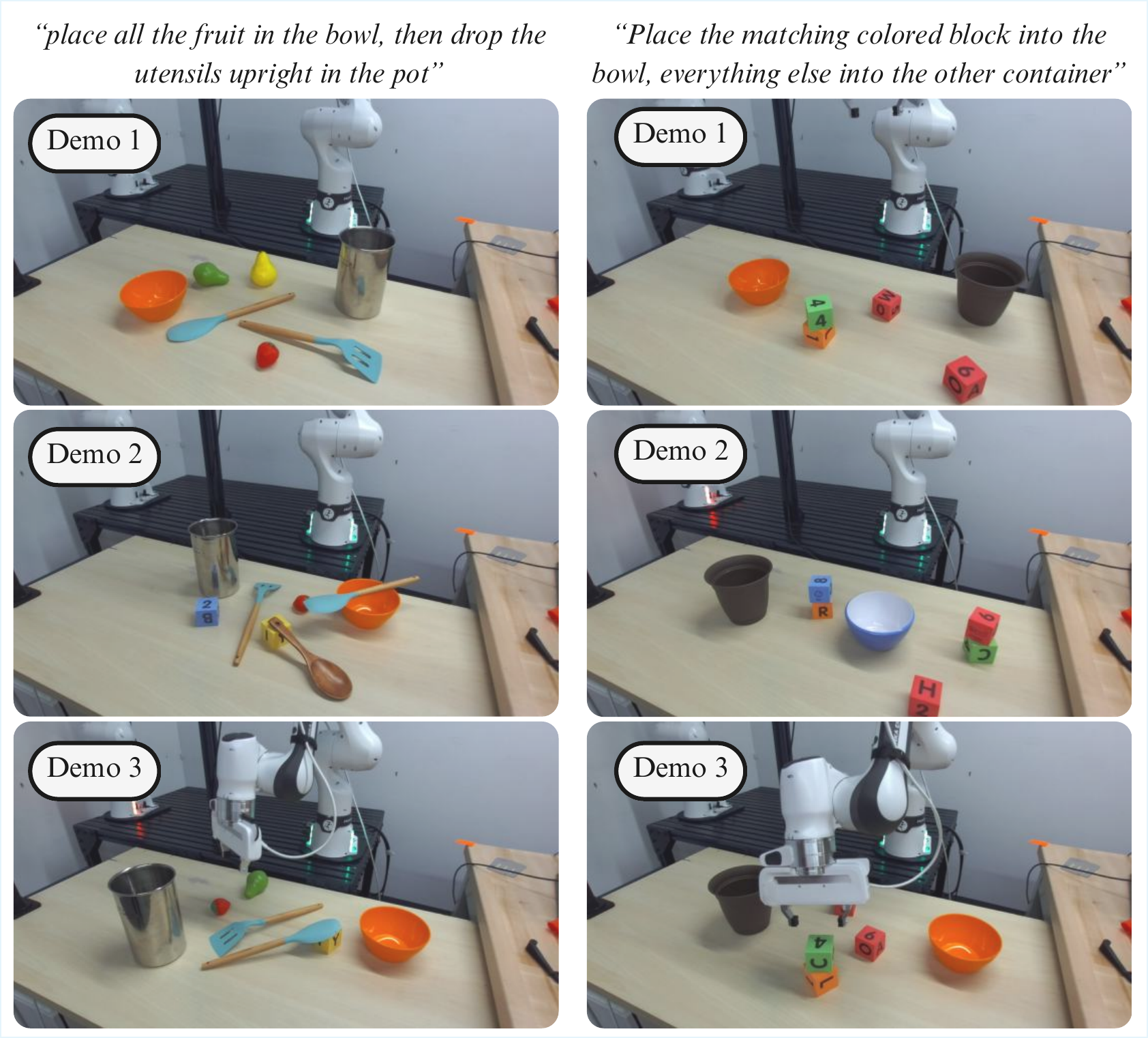}
        \caption{Diverse demonstrations from two long-horizon task families used for few-shot inference; language descriptions are shown only for readers.}
        \label{fig:real-robot-exp-a}
    \end{subfigure}
    \hfill
    \begin{subfigure}[t]{0.49\linewidth}
        \centering
        \includegraphics[width=\linewidth]{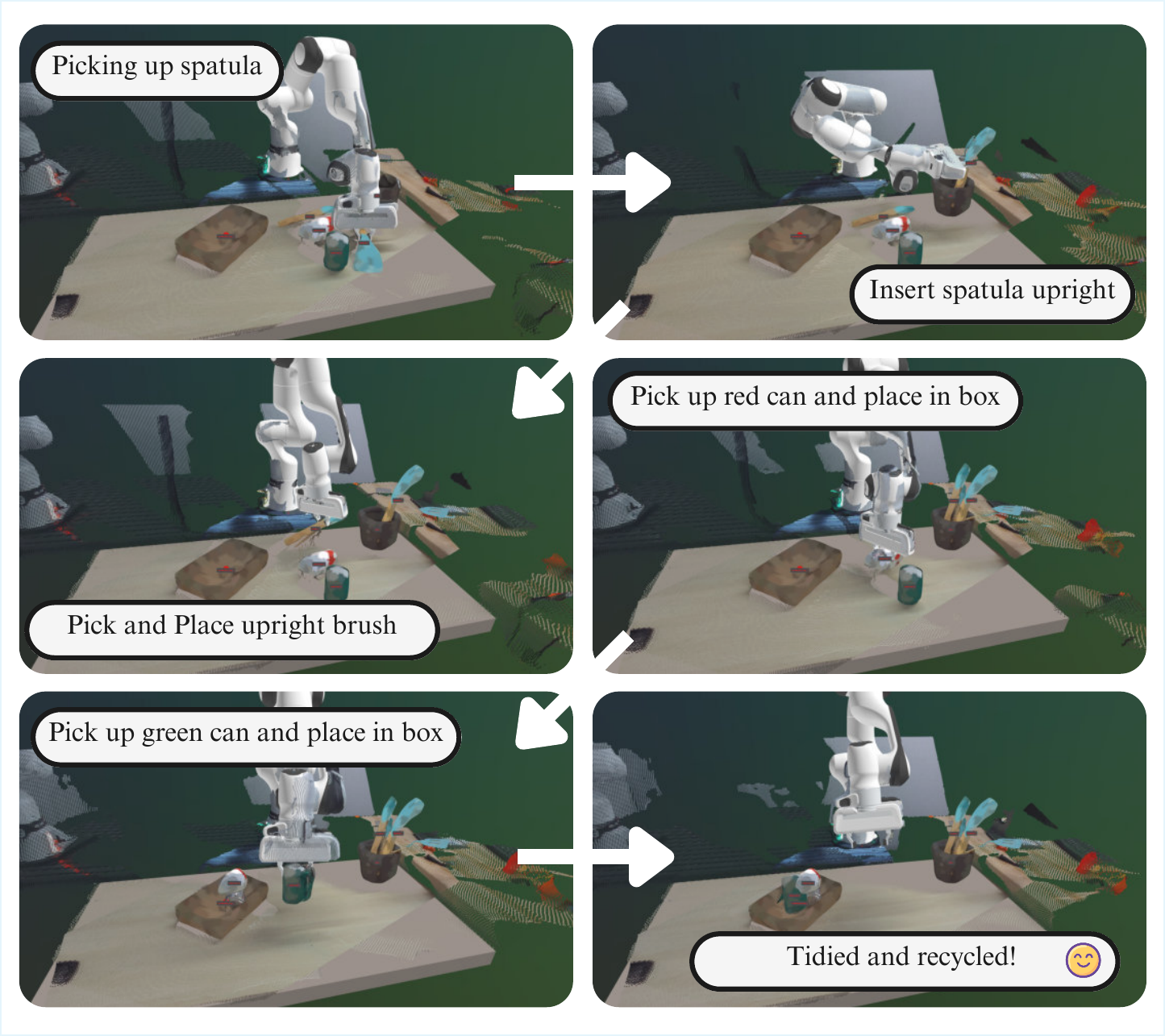}
        \caption{Reconstructed object-centric scene states and TAMP execution of the inferred program in a novel scene.}
        \label{fig:real-robot-exp-b}
    \end{subfigure}

    \caption{Real robot experiments}
    \label{fig:real-robot-exp}
\end{figure*}

\paragraph{RIR approximates the posterior over explanations:}
\Cref{fig:comprehension_topk} reports comprehension accuracy for the top-$k$ explanations ($k\!\in\!\{1,5,10\}$). Accuracy rises for all approaches as more samples are considered, yet the rate of improvement differs dramatically. With RIR, performance jumps to human parity at $k=5$ and exceeds it at $k=10$. The pattern is even clearer on the spatial+alg-3-shot subset: the closest VLM baseline gains little from additional samples; its prior rarely generates plausible explanations and therefore fails to approximate the posterior distribution. In contrast, RIR performs explicit inference over the candidate space, producing a much sharper posterior and reaching human-level comprehension with far fewer samples.

\paragraph{Inferring rational explanations improves generalization to novel task instances}
In \cref{tab:terc_succrate}, we present success rates when the inferred explanation functions are applied to unseen task instances. Across all evaluation subsets, RIR surpasses RwoR, with the largest margin in the spatial-3-Shot subset (22 \% $\rightarrow$ 59 \%). This improvement is expected, as higher comprehension drives higher success.

\subsection{Real-World Tabletop Manipulation}

In this section, we discuss the deployment of RIR inference, and subsequent TAMP execution, on a Franka FR3 Robotic arm across diverse tabletop scenes.

\subsubsection{Implementation}
In real environments, we do not have access to the privileged simulation state, and must contend with partial observability and noise. As a result, we develop a real-to-sim pipeline that reconstructs the required sequence of object-centric states $\{s_t\}_{t}^T$ defined in \cref{sec:problem_setting} from raw observations in the demonstrations (RGBD frames and robot joint states). Our system is similar to recent pipelines that extract 3D object centric reconstructions from demonstrations \cite{scenecomplete}, and we defer a full pipeline description to Appendix~\ref{app:real2sim}. Overall, the system relies on multi-view RGBD frames, annotated with modern VLM pointing, foundation model segmentation and object mesh reconstruction, registration and 6D pose tracking. Demonstrations of long-horizon multi-step manipulation tasks (each 2-4 minutes long) are collected via a VR teleoperation system, where a human operator controls the end-effector pose in real-time using the Meta Quest VR headset. The scene reconstruction is used by the TAMP system, both at inference-time, to obtain an explanation program $E$, and at execution-time, to execute $E$ in a novel scene.

\subsubsection{Task Distribution}
We evaluate our system on a set of 4 long-horizon macro-tasks that have a large set of potentially plausible explanations. They involve complex sorting and tidying tasks, as illustrated in~\cref{fig:real-robot-exp}, where the robot has the capacity to move, pick and place objects, and the VLM is able to generate programs to describe goals and constraints associated with these behaviors. Evaluating comprehension occurs across 3 trials with different demonstrations for each trial. Across trials, the system is presented with novel objects in unseen layouts, with additional random clutter/distractor objects. To highlight the breadth of scenes that the systems' generalization is evaluated over, we visualize a subset in \cref{fig:generalization-taxonomy}.

% \begin{table}[h]
%   \centering
%   \caption{\ben{I'm going to change this to a bar chart rather than a table, for now, imagine a nice bar chart please.} Real robot results across 4 long-horizon tasks (Franka FR3). @1/@3 = inferred from 1 vs 3 demonstrations. Comprehension (as defined above), is the human-adjudicated program equivalence. Behavioral success involves evaluating the generated behavior on many novel scenes, reconstructed using the sim-to-real pipeline. }
%   \label{tab:fr3-succrate}
%   \scriptsize
%   \setlength{\tabcolsep}{2.5pt}
%   \renewcommand{\arraystretch}{0.92}
%   \begin{tabular}{@{}l c c c@{}}
%     \toprule
%     \textbf{Approach} &
%     \makecell[c]{\textbf{Comprehension} (\%)\\1-shot / 3-shot} &
%     \makecell[c]{\textbf{Behavioral Succ.} (\%)\\1-shot / 3-shot} \\
%     \midrule
%     RIR &

%     \textbf{75} / \textbf{75} $\pm$ 12.5 &
%     \textbf{42$\pm$7} / \textbf{57$\pm$8} \\
%     $-$ [Rationality] (Rw/oR) &

%     0 / 0 &
%     8$\pm$2 / 10$\pm$2 \\
    
%     $-$ [Rat., Scene Recon] &

%     0 / 0 &
%     17$\pm$3 / 29$\pm$4 \\
%     $-$ [Rat., Scene Recon, Bottom-up] &

%     0 / 0 &
%     17$\pm$4 / 11$\pm$3 \\
%     \bottomrule
%   \end{tabular}
% \end{table}

\begin{figure}[t]
    \centering
    \includegraphics[width=\linewidth]{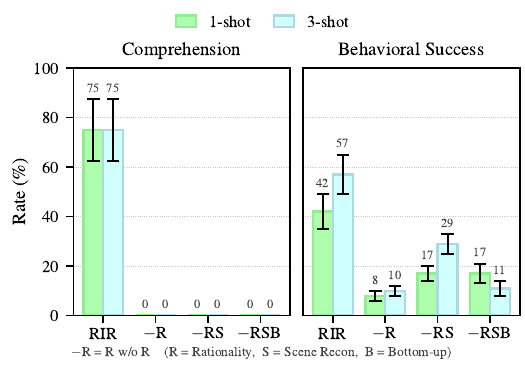}
    \caption{Real robot results across 4 long-horizon tasks (Franka FR3). 1-shot/3-shot = inferred from 1 vs 3 demonstrations. Comprehension (as defined above), is the human-adjudicated program equivalence. Behavioral success involves evaluating the generated behavior on many novel scenes, reconstructed using the sim-to-real pipeline. RIR is our base approach. Removing the rational inverse likelihood scoring yields the baseline approach, Reasonign w/o Rationality ($-$R). We also include results for removing the scene reconstruction ($-$RS), as well as the Bottom-up proposals $-$RSB from the VLM reasoning iterations. }
    \label{fig:real-robot-behave}
\end{figure}

\subsubsection{Real Robot Results}

Our experiments demonstrate feasibility on four long-horizon macro-tasks. \cref{fig:real-robot-behave} illustrates, for over test tasks, comprehension and behavioral success rates on novel scenes after 1 and 3 demonstrations. RIR was able to recover the true hidden intent $75\%$ of the time based on just a single demonstration; and in completely novel scenes it succeeded in achieving the demonstrator's complex objective $57\%$ of the time. Notably, we see that rational inference is a key driving factor for inferring generalizable programs from demonstration, even as compared to the ablated state-of-the-art VLM models provided with the full scene descriptions, processed demos and code examples (R w/o R).  

Looking at these results more closely: all approaches besides RIR recover 0\% comprehension. Their small non-zero behavioral success is coincidental satisfaction of $E^*$ by impoverished task specifications, not task understanding. This occurs due to the broad variation in eval environments we reconstruct from. Conversely, RIR's behavioral success is below 75\% due to planning failures associated with challenging initial scene configurations $x_0'$: while it's possible to infer the correct program from the 1 demonstration, executing it on a more cluttered scene is still a challenge. Additionally, we see that for RIR the comprehension stays the same even when more demonstrations are provided (1-shot and 3-shot comprehension is the same). For the same 3/4 tasks, RIR inferred the correct (equivalent) program (over 12 total trials). In this case, the comprehension rate's granularity fails to resolve the intra-program refinements that come with additional demonstration data (better filters, improved constraint allocation/ordering), an improvement which the behavioral success column illustrates ($42 \% \to 57 \%$). That is to say, there are cases where the stronger program equivalence conditions aren't met, yet additional demonstrations can allow RIR to refine the program to be closer to the true intent. Removing rational inference has a significant effect on the overall behavioral success (dropping 3-shot performance from $57\%$ to $10\%$). Interestingly, removing scene reconstruction from the VLM prompt actually improves performance over the Rw/oR baseline. We attribute this to the VLM overfitting its generated programs to the scene reconstruction (without the presence of RIR to prevent this), rather than using the reconstruction to guide proposals.

\begin{figure}
    \centering
    \includegraphics[width=\linewidth]{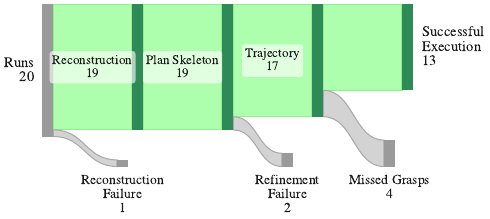}
    \caption{Sankey diagram for the physical execution of the specific learned macro task: \textit{sort the colored block into the matching colored bowl, then place the other blocks in a different container.}}
    \label{fig:sankey}
\end{figure}

Physical execution success is best understood qualitatively in terms of the conversion rates between an initial scene $x_0$ and program $E$, and the downstream long-horizon execution. For example, consider the failure sankey diagram \cref{fig:sankey}. Here, we see that the main barriers between behavioral success and physical execution success are related to missed or dropped grasps. This is a current limitation of the system; we execute the long-horizon plan open-loop. As this is not one of our main contributions, we leave closed-loop extensions as a separate line of future work.

% \ben{Make the final statement clear - better filters etc. Maybe also include the case study sankey diagram, as well as the program}

% In \cref{app:robot_exp}, we detail the failure modes associated with executing these trajectories on the real robot, evaluating the conversion rate between behavioral success and physical execution.

% \input{text/5_discussion}

\section{Limitations and Future Work} \label{sec:limitations}

In it's current instantiation, RIR is an imitation learning algorithm, learning offline from a demonstration set. Converting RIR to an online inference algorithm has the potential to improve human-robot interaction. For example, we currently only pick the MAP explanation program, throwing away the distribution we've constructed during inference. Instead, we can leverage this uncertainty over programs to make learning an interactive process. In this way, we can move from offline imitation learning to \textit{teaching rational robots} online. 

Additionally, RIR leverages a fixed TAMP planning system to provide the relevant structure for inference. This is often too rigid to apply to open-world robotics. Instead, a future system could leverage rationality principles as a means of guiding the on-demand synthesis of locally relevant policies and models, expanding our planning theory. 

The current system relies on external large-scale VLMs as the program proposal prior. This means we can't update our prior distribution with data. Recent work has shown that fine-tuning smaller VLMs improves performance for these more specific embodied code generation tasks \cite{fu2025capx}, and is therefore a promising direction of future work. 

% This form of learning is conducive to curricula, where an agents world knowledge and skills can be taught over time, in the same way you'd teach children.

\section{Conclusion}

We introduced Rational Inverse Reasoning, which treats demonstrations as evidence generated by a boundedly rational planner and recovers intent as a \emph{compact, compositional explanation program}. By coupling a foundation-model proposal prior with planner-grounded likelihood in TAMP, RIR produces explanations that are interpretable, executable, and transferable, supporting one- and few-shot generalization in continuous manipulation. We believe this lays groundwork for more generalizable and explainable imitation learning, moving closer to the human ability to learn robustly from just a few examples by reasoning about the \emph{why} behind observed behaviors.

% \newpage
\bibliographystyle{unsrtnat}
\bibliography{references}

\clearpage
\newpage

\appendix

\appendices

\section{Generalized Task and Motion Planning (TAMP)} \label{app:tamp}

As noted in the main paper, we fulfill the long-horizon embodied reasoning tasks $g$ with Task and Motion Planning (TAMP). In the following sections, we use a similar description and notation for many of the components of TAMP from \citet{shen2024cutamp}. A TAMP instance is defined as a tuple $\mathcal P = (\mathcal{W}, s_0, g)$ consisting of set of parameterized operators $\mathcal{W}$, the initial state $s_0$, and the grounded task specification $g$. States are represented symbolically using predicates $v \in V$; for example, the predicate \texttt{At(obj, location)} holds if object \texttt{obj} is at symbolic location \texttt{location}.

Each parameterised operator $\omega \in \mathcal{W}$ is specified by:
\begin{itemize}
    \item \textbf{Parameters:} A tuple $\theta_\omega = (\theta_{\text{disc}}, \theta_{\text{cont}})$ of discrete (objects, symbolic locations), and continuous (trajectories, configurations) parameters.
    \item \textbf{Constraints:} Constraints $\text{Con}(\theta_\text{cont})$ on continuous parameters which must be satisfied for action feasibility. For example, kinematic constraints connecting grasp poses and feasible inverse kinematics solutions.
    \item \textbf{Preconditions:} A logical formula describing conditions on the symbolic state $\text{Pre}(\theta_\text{disc})$ which must be valid for the action to be applicable.
    \item \textbf{Effects:} A logical formula describing the atoms $\text{Eff}(\theta_\text{disc})$ made true or false upon successful execution of the action.
    \item \textbf{Maintain:} A logical formula describing any conditions on the symbolic state $\text{Mnt}(\theta_\text{disc})$ which must \textit{remain} true during action execution, enabling runtime monitoring. For example, maintaining a hold on an object while moving it.
    \item \textbf{Cost:} An optional scalar cost $C(\omega)$ for optimization - usually set to $1$ for symbolic search.
\end{itemize}

For example, consider the \texttt{Place(q1, g, t, q2, p, obj, loc)} action for moving an object \texttt{obj} to symbolic location \texttt{loc} at placement \texttt{p} from configuration \texttt{q1} to configuration \texttt{q2} via trajectory \texttt{t} and grasping it with grasp \texttt{g}. This abstract action has:

\noindent\textbf{Preconditions:} \texttt{Holding(obj)}, \texttt{AtConf(q1)}, ...

\noindent\textbf{Effects:} \texttt{HandEmpty()}, \texttt{AtConf(q2)}, ...

\noindent\textbf{Constraints:} \texttt{Kin(q, g, p)}, \texttt{CollisionFree(q1, q2, t, g)}, ...

\noindent\textbf{Maintain:} \texttt{Holding(obj)}

Note that these operator invariants defined by the \textbf{Maintain} conditions differ from the task-level constraints associated with the explanation function compilation into the task specification $g$. Concretely, this is the difference between an operator invariant such as an IK configuration validity check, as opposed to a constraint which is task-relevant such as \textit{not placing objects down in a particular region}. 

A plan skeleton (high-level strategy) $\hat{\pi} = (\underline{\omega}_1, \dots, \underline{\omega}_n)$ is a sequence of grounded abstract operators, where each grounded operator $\underline{\omega}$ binds discrete parameters of an action template. For example, the grounding of the abstract operator \texttt{Place(obj, location, ...)} to the grounded operator \texttt{Place(box, shelf, ...)}.

This grounding procedure applied to the abstract actions over the space of predicates and symbolic entities in the world induces an abstract transition model. This abstract transition model $F: \mathcal{R}_{\phi} \times \underline{\mathcal{W}} \rightarrow \mathcal{R}_{\psi}$, where $\mathcal{R}_{\phi}$ is the symbolic state space induced by perception function  $\phi$, checks logical feasibility and updates states symbolically.

The planner searches iteratively. In the taxonomy of TAMP solvers, we consider a \textit{search-then-sample} procedure. The planner first generates candidate symbolic skeletons using abstract reasoning (via the abstract transition model $\mathcal{F}$), then \textit{refines} each candidate by sampling continuous parameters. If a refinement fails due to unsatisfied constraints, the symbolic planner proposes a new skeleton. This iterative search continues until a valid solution is found or a computational budget is exhausted.

The symbolic layer ensures logical and temporal soundness, while continuous refinement solves geometric constraints to produce an executable robot trajectory. This clearly defined bi-level TAMP formulation leverages symbolic reasoning to efficiently guide continuous refinement, enabling the generalization and execution of abstract task instructions across diverse environments.

\subsubsection{Plan Search}

Constructing plan skeletons $\hat\pi$ involves symbolic search through the transition model induced by the set of operators $\mathcal{W}$. In this case, we must consider an entire grounded task specification $g$ to solve for. This includes sub-goals (discrete constraints) and goals, which alters the search procedure. 

Specifically, we instantiate an A* search procedure with a staged Hamming heuristic to guide the A* search towards task-relevant solutions. Although more efficient search heuristics exist, the staged Hamming heuristic is very simple to compute, and as such, results in a sufficiently fast search, even if the number of expanded nodes in the tree is high. The idea behind the staged Hamming heuristic is that we treat each cumulative goal from the current state onward as way-points. The heuristic is the sum of Hamming distances (symmetric set difference sizes) between consecutive way-points:
\[
\begin{aligned}
h ={} & \left|\text{state} \;\oplus\; g_{\text{stage}}\right| \\
     & {}+ \left|g_{\text{stage}} \;\oplus\; g_{\text{stage+1}}\right| \\
     & {}+ \dotsb + \left|g_{\text{stage-1}} \;\oplus\; g_{\text{final}}\right| 
\end{aligned}
\]
We then construct a plan generator which allows us to sample plans from the tree by cost order. Instantiating one generator allows us to re-use computation across iterations. For a fixed compute budget, we sample up to a maximum number of plan skeletons based on their symbolic cost - this is large, as the costly step in this process is plan refinement, rather than symbolic search. Across runs, we perform de-duplication and prefix matching/blocking to ensure that plans which we've found to be impossible (cannot refine into a valid trajectory) are removed from the induced transition model. 

\subsubsection{Plan Refinement}

Refining a plan skeleton $\hat \pi = (\underline{\omega}_1, ..., \underline{\omega}_P)$ involves producing a set of parameters $\theta  = \{\theta_{\omega_1}, ..., \theta_{\omega_P}\}$ that satisfies the spatio-temporal constraint graph induced by $\hat \pi$. That is, each plan skeleton $\hat \pi$ corresponds to a constraint satisfaction problem (CSP). 

In addition to these abstract operators, we attach sampling sub-graphs to expedite the solution procedure. Specifically, we can design sampling mechanisms (motion planning, grasp sampling) on the constraint manifold, verifying constraints, and solving the local constraint subgraphs at the same time. 
\begin{figure}[h]
    \centering
    \includegraphics[width=1.0\linewidth]{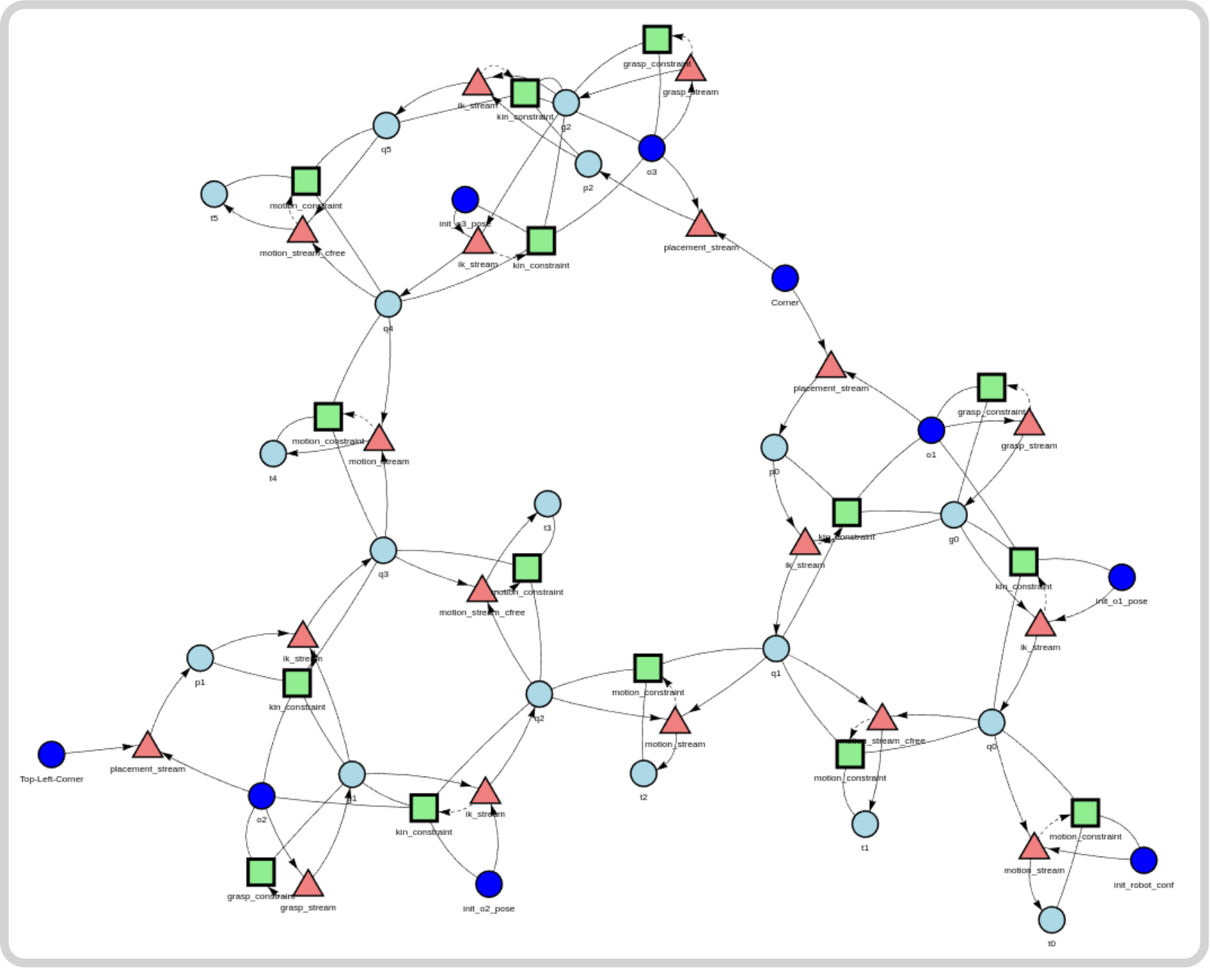}
    \caption{An example constraint graph for the task of moving two objects into the corners of the environment.}
    \label{fig:supp-const}
\end{figure}
For example, consider \cref{fig:supp-const}. This figure is a diagram of the semi-directed constraint graph involved in solving for the continuous parameters of the task of moving both objects in the scene to the corners. The blue circles represent the parameters $\theta$ that we have to solve for. The green squares represent the constraints that we have to satisfy for the solution to be valid. The red triangles here are the samplers that we use to produce candidates for each of the parameters. The dark-blue circles denote the known parameters - those which are imposed by the plan skeleton, or are already present given the initial state of the environment. 

Solving this kind of spatio-temporal constraint graph without heuristics or on-manifold sampling is not possible. From \cref{fig:supp-const} we can see that the graph is highly structured, with many local subgraphs with similar constraint structure. This allows us to devise an informed CSP solver.

\paragraph{Informed CSP Solver}

\begin{figure}[t]
    \centering
    \includegraphics[width=1.0\linewidth]{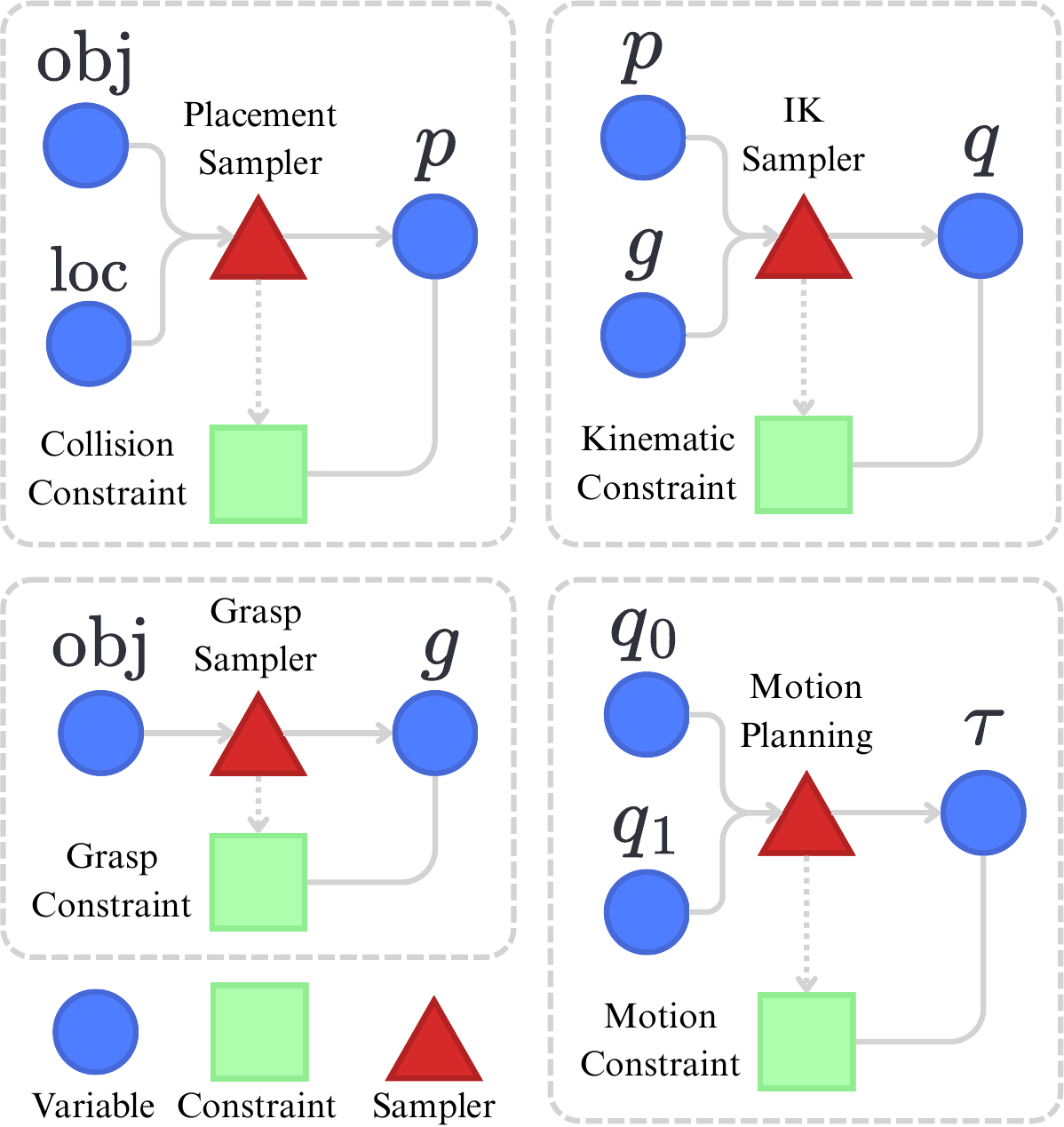}
    \caption{Local sampling procedures which we compose for CSP subgraph solving.}
    \label{fig:supp-subgraph_solve}
\end{figure}

The constraint graphs which result from embodied reasoning tasks are not without structure. This allows us to define a hierarchical sampling scheme which, by construction, satisfy local subgraphs in the CSP. Each of these verified sampling steps are shown in \cref{fig:supp-subgraph_solve}
In this way, we can prioritize: $\text{Pose}\rightarrow\text{Grasp}\rightarrow\text{IK}\rightarrow\text{Motion Planning}$ to solve each \texttt{Pick} and \texttt{Place} and any other similar skill subgraph sequentially. 

This affords us two modes of backtracking when we cannot solve for certain parameters (given the previous parameters). First, we can backtrack within the local sub-graph solve. In other words, if we cannot sample a valid trajectory, we can backtrack and re-sample a new placement for that sub-graph. Beyond this local backtracking, we can also perform a global backtrack by discarding the current solution and resetting the solver back to the original state. 

By fixing the number of each backtracks and samples we use during the plan refinement step, we fix the amount of computation to apply per-problem, when deciding whether a plan skeleton is downward-refinable - i.e., whether the plan-skeleton CSP can be solved (and therefore executed) in the environment. 

Notably, in \cref{app:stamp}, we detail our real-robot solver, which uses parallelized solving, and therefore doesn't require back-tracking given sufficient parallel solving particles. 

\begin{figure}[t]
    \centering
    \includegraphics[width=1.0\linewidth]{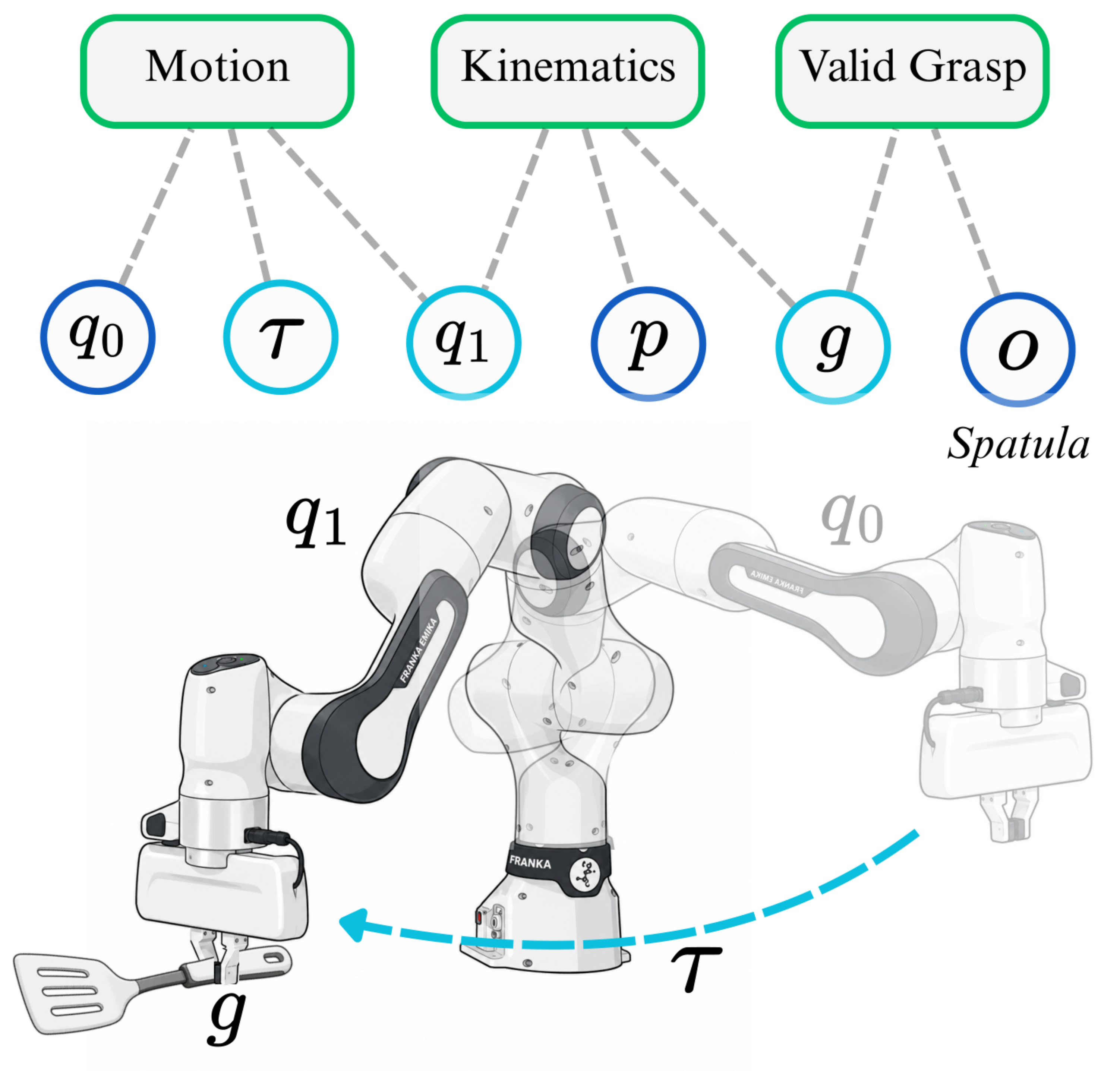}
    \caption{Example small local CSP for the problem of \texttt{Pick}-ing an object up. }
    \label{fig:csp-solver-pick}
\end{figure}

\begin{figure}[t]
    \centering
    \includegraphics[width=1.0\linewidth]{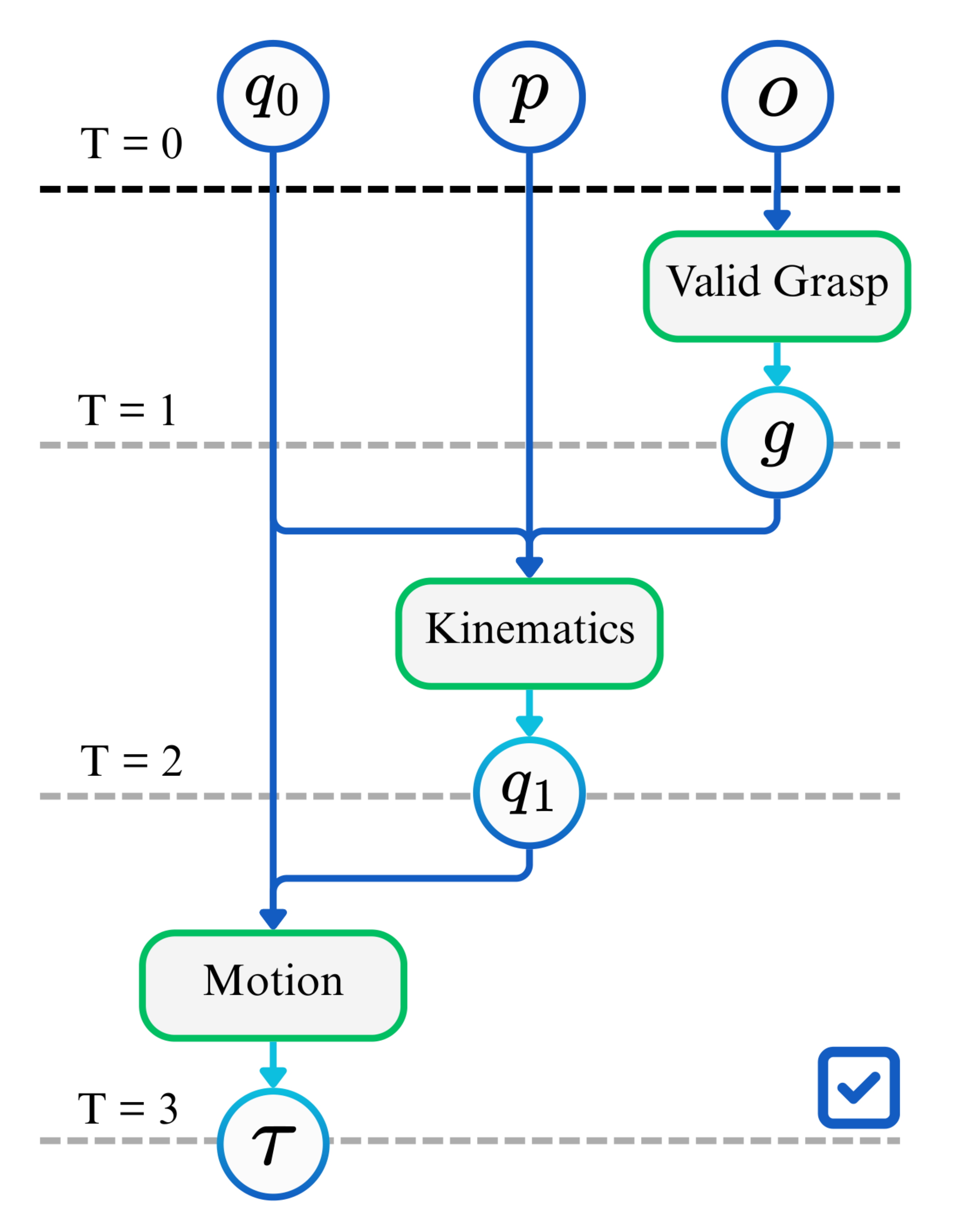}
    \caption{Example CSP sub-graph iterative solver based on the local on-manifold/dedicated samplers outlined in \cref{fig:supp-subgraph_solve}}
    \label{fig:csp-solving-pick}
\end{figure}

An example of the csp for a robot \texttt{Pick} operator is shown in \cref{fig:csp-solver-pick}. Here the unknown variables are $g, q_1, \tau$ (the grasp, final robot configuration, and the trajectory to get there), while the known variables are $o, p, q_0$ (the object, its current pose, and the initial robot configuration). The path to solving it therefore calls into use the grasp, IK and motion planning sub-graph solvers, and the constraint graph structure induces the solver flow illustrated in \cref{fig:csp-solving-pick}, where each constraint and variable is solved and sampled in turn.  

\section{Rational Inverse Reasoning (RIR)} \label{app:rir}

In this section, we provide additional details regarding the Rational Inverse Reasoning (RIR) algorithm. Domain specific instantiations and algorithmic details (TERC and Real-Robot) are described in their respective sections \cref{app:terc} and \cref{app:robot_exp}).

\subsection{Bottom-Up Demonstration Structure and Candidate Skeleton Sets}

A central algorithmic idea in RIR is to combine \emph{top-down} hypothesis proposal (via a VLM/LLM program prior) with \emph{bottom-up} constraints/proposals extracted from the demonstration. Bottom-up extraction serves two roles: it (i) identifies skeletons that \emph{could} have generated the demonstration under a broad class of specifications and (ii) provides a filtering mechanism to prune explanation programs whose grounded specs cannot admit any skeleton consistent with the demonstration.

\subsubsection{Symbolic trace and segmentation}
Given a demonstration trajectory $\tau=(x_0,a_0,\dots,x_T)$ we compute a symbolic trace
\begin{equation}
\phi(\tau) = (R_0,\dots,R_T), \qquad R_t = \phi(x_t),
\end{equation}
where each $R_t$ is a set of predicates true at time $t$. Each operator instance $\omega$ has symbolic \emph{preconditions} $\mathrm{Pre}_\omega$, \emph{effects} $\mathrm{Eff}_\omega$, and \emph{maintain} constraints $\mathrm{Mnt}_\omega$ that must remain true throughout the operator execution.

A \emph{segment} is a time interval $[i,k]$ associated with an operator $\omega$ such that:
(i) $R_i \models \mathrm{Pre}_\omega$,
(ii) for all $t \in \{i,\dots,k-1\}$ we have $R_t \models \mathrm{Mnt}_\omega$,
and (iii) $R_k$ is consistent with applying $\mathrm{Eff}_\omega$ (and any associated postconditions). This notion is domain-specific in its predicate content but domain-general in structure.

\subsubsection{Segmentation DAG and bottom-up skeleton proposals}
We build a directed acyclic graph (DAG) whose nodes correspond to time indices and whose edges correspond to feasible segments. A path from node $0$ to node $T$ corresponds to a discrete skeleton (operator sequence) that can symbolically explain the demonstration. This yields a finite set of \emph{bottom-up plan proposals}:
\begin{equation}
\hat{\Pi}_{\mathrm{obs}}(\tau) \subset \Pi(\cdot),
\end{equation}
often obtained by extracting the (all or k-best) paths under operator priors.
\begin{algorithm}[t]
\caption{Build segmentation DAG from a symbolic trace}
\label{alg:segdag}
\KwIn{Symbolic trace $(R_0,\dots,R_T)$, operator library $\mathcal{W}$}
\KwOut{Segmentation DAG $(\{0,\dots,T\},\mathcal{E})$}

Initialize the DAG with nodes $\{0,\dots,T\}$ and
$\mathcal{E} \gets \emptyset$\;

\For{$i \gets 0$ \KwTo $T-1$}{
    \For{$k \gets i+1$ \KwTo $T$}{
        \ForEach{operator instance $\omega \in \mathcal{W}$ compatible with the objects in $(R_i,\dots,R_k)$}{
            \If{
                $R_i \models \mathrm{Pre}_{\omega}$ \textbf{and}
                $(\forall t \in \{i,\dots,k-1\})\,
                R_t \models \mathrm{Mnt}_{\omega}$ \textbf{and}
                $R_k \models \mathrm{Post}_{\omega}$
            }{
                $\mathcal{E} \gets
                \mathcal{E} \cup \{(i,k,\omega)\}$\;
            }
        }
    }
}

\Return{$(\{0,\dots,T\},\mathcal{E})$}\;
\end{algorithm}

\begin{algorithm}[t]
\caption{Extract bottom-up skeleton proposals via $k$-best/all paths}
\label{alg:kbest}
\KwIn{Segmentation DAG $(\mathcal{V},\mathcal{E})$, path budget
$K_{\mathrm{obs}}$, edge cost $c_{\mathrm{edge}}$}
\KwOut{Bottom-up skeleton proposals $\hat{\Pi}_{\mathrm{obs}}(\tau)$}

Compute the $K_{\mathrm{obs}}$ best paths from $0$ to $T$ under
$c_{\mathrm{edge}}$, e.g., using a $k$-shortest-path algorithm for
DAGs\;

\ForEach{computed path $p=(0\rightarrow\cdots\rightarrow T)$}{
    Convert $p$ into a skeleton $\hat{\pi}$ by reading its operator
    labels\;
    Add $\hat{\pi}$ to $\hat{\Pi}_{\mathrm{obs}}(\tau)$\;
}

\Return{$\hat{\Pi}_{\mathrm{obs}}(\tau)$}\;
\end{algorithm}

\subsubsection{How bottom-up proposals enter the likelihood}
The bottom-up set constrains the skeleton marginal in two ways.

First, the execution model we use below includes an indicator $\mathbf{1}(\tau; \hat \pi, g)$, which is efficiently computable by verifying whether a valid path in the segmentation DAG corresponds to $g$. Thus, the skeleton marginal effectively restricts to skeletons (and therefore task specifications) consistent with the demonstration.

Second, bottom-up proposals provide a \emph{finite support approximation} to the skeleton space that is particularly well-aligned with the data, as described in \cref{sec:demo-counterfactuals}. In practice, for each $(E,k)$ we build a candidate set
\begin{equation}
\hat{\Pi}^{(k)}(E)
\;=\;
\hat{\Pi}_{\mathrm{bu}}(g^{(k)}, \tau^{(k)}) \;\cup\; \hat{\Pi}_{\mathrm{td}}(g^{(k)},x_0^{(k)}),
\label{eq:candidate_set_union}
\end{equation}
where $\hat{\Pi}_{\mathrm{td}}$ are additional \textit{top-down} skeletons produced by the planner under the grounded spec. The union ensures that we compare the demonstrated strategy to plausible alternatives under the same hypothesis, which is essential for skeleton-level Boltzmann rationality (Section~\ref{sec:strat-rat}). Notably, if the abstract/symbolic representation is sufficiently expressive - \textit{an efficient abstraction }- the total number of bottom-up plan proposals is reduced, directly boosting inference efficiency. 

\subsection{Execution Likelihood: From Boltzmann Rationality to Symbolic Indicators}
\label{app:rir-indicator_exec}
In this section, we shed additional light on the trajectory-level execution/refinement rationality $p_{\mathrm{traj}}(\tau \mid \hat{\pi}, g, s_0)$, the likelihood of a demonstrated execution given skeleton and spec, modeled by a Boltzmann reweighting of the refinement proposal distribution.

\subsection{Why a dense trajectory likelihood is intractable}
A classical inverse-planning model would apply Boltzmann rationality at the action level, e.g., $p(a_t\mid x_t,\hat{\pi}_t)\propto \exp(-\beta c_t(a_t))$, and accumulate likelihood over time. In high-dimensional continuous manipulation, this would require solving a nested planning/refinement problem for many prefixes and many hypotheses, making direct evaluation computationally prohibitive. In addition, much of this computation would be wasted, given that human demonstrators are rarely making new logical/task/constraint level decisions at every single time-step that require counterfactual reasoning. In smaller toy problems (e.g. discrete gridworld), we could solve for the value function for any task specification we're considering. However, doing this for real-world robotics problems which combine goals, subgoals and constraints during long-horizon execution is completely intractable. 

RIR therefore uses an \emph{efficient abstraction} assumption: the symbolic trace captures the semantically meaningful aspects of the demonstration relevant to the explanation, and details outside this trace, besides cost and relative specificity, are treated as nuisance variability. We define two fast verifiers:
\begin{equation*}
\mathrm{ValidDemo}(\tau,g) \in \{0,1\}, \qquad
\mathrm{Consistent}(\tau,\hat{\pi}) \in \{0,1\},
\end{equation*}
where $\mathrm{ValidDemo}$ checks whether the demonstrated trajectory satisfies the goals/constraints in $g$ at the symbolic level, and $\mathrm{Consistent}$ checks whether the demonstration can be segmented as an execution of skeleton $\hat{\pi}$ (equivalently, whether $\hat{\pi}$ is supported by the segmentation DAG). Using this in \cref{eq:indicators-boltzmann-weighted} corresponds to assuming the demonstrator is approximately uniform boltzmann-reweighted over satisfying trajectories under the constraints induced by $g$ and $\hat \pi$. From a value computation perspective, this indicator model implicitly assigns infeasibility as infinite cost. 

This is an important assumption. Demonstrations which come \textit{near} the goal but never reach it have zero probability mass under the goal they were trying to achieve. In other words, the current formulation of the system will never infer an explanation program which describes in an abstract sense a task that was never actually achieved in any of the demonstrations. Concretely, if the user hovers over the coaster, but never actually places the mug down and instead moves somewhere else, inference places no likelihood on an explanation which involves the mug being placed on the coaster. However, if the user accidentally dropped/placed the mug in the wrong location first, and then moved it to the coaster and \textit{did} place it down, this kind of failure \textit{is} captured in our bounded strategy rationality (non-zero probability mass). This is a fairly strong modeling assumption, but it allows us to significantly increase inference speed, allowing for symbolic verification and segmentation to act as a source of (noisy) truth.  

\subsection{Additional Mathematical Insights and Practical Notes}
\subsubsection{Why free energy is the right object for skeleton normalization}
The free energy in \eqref{eq:main-execution-free-energy} can be seen as the negative log of an empirical moment generating function of costs. Its use yields two benefits that are particularly important in planner-in-the-loop inference. First, it reduces sensitivity to heavy-tailed optimization noise: a skeleton with a single rare low-cost refinement but otherwise poor \textit{refinability} is not unduly favored. Second, it aligns the skeleton score with the same $\beta_{\mathrm{traj}}$ used in refinement distributions and normalizer ratios, making the hierarchy thermodynamically consistent: refinements are Boltzmann-weighted at inverse temperature $\beta_{\mathrm{traj}}$, and skeleton selection uses a Boltzmann model over the resulting free energies at inverse temperature $\beta_{\mathrm{plan}}$.

\subsubsection{Interpretation of evidence ratios as "rarity of constraints"}
In the nested case, \eqref{eq:adjacent_specificity_estimate} highlights the relative specificity approximation (under the weaker model) that a rollout also satisfies the stronger constraints. Taking logs, the execution evidence in \eqref{eq:specificity_correction} adds $-\log \Pr(\text{success under stronger} \mid \text{weaker})$, which is exactly a self-information (surprisal) term. This clarifies why subtle constraints that barely affect cost (e.g., uprightness) can nevertheless be strongly disambiguated: they can be \emph{rare} under the weaker model even when they are not costly when enforced.

\subsubsection{What is (and is not) being "Bayes factored"}
Although we refer to \eqref{eq:specificity_telescoping}  as Bayes-factor bridge style terms, the algorithm does not require specifying a full absolute probability measure on continuous trajectories. Instead, we exploit ratios of normalizers across closely related constrained models, which are well-defined even when the base measure is implicit. This is precisely the regime where classical techniques for estimating ratios of normalizing constants (subset ratios, bridge sampling, AIS) are most appropriate.

\subsubsection{Choice of canonical base $g_0$}
Evidence ratios depend on the reference. To preserve semantic meaning and comparability across hypotheses, the canonical base $g_0$ should preserve the task-defining goal semantics while removing optional continuous/ordering constraints. Bases that are too weak can destroy overlap and inflate estimator variance; bases that are too strong can hide the discriminative effect of stylistic constraints. Sample-size-guided bridging provides an automated safeguard when $g$ is far from $g_0$.

\subsubsection{Complexity and parallelism}
RIR’s computational bottleneck is refinement sampling for free energies and for reference caches. Both are parallel across skeletons, particles, and demos. The caching strategy is essential: by storing trajectories from all particles and weaker specs, RIR can evaluate many stronger hypotheses with only cheap symbolic checks, sharing all expensive refinement estimates, and enabling efficient planner-in-the-loop scoring over candidate sets across refinement iterations. 

\section{The Tiny Embodied Reasoning Corpus (TERC)} \label{app:terc}

In this section, we outline details relating to the dataset presented in this paper; the Tiny Embodied Reasoning Corpus (TERC). First, we discuss the simulation and parameters. We then discuss how we gather human teleoperation data. Next, we provide a taxonomy of the dataset, providing example rollouts/images from each required task, as well as example explanation functions that the system has to extract from teleop data. We then discuss the domain-specific details regarding implementing TAMP and RIR within the TERC environments and task family. Finally, we highlight our generalization taxonomy, illustrating how we evaluate each method by changing object and agent poses, object shape, color, size and numbers of objects.  

\subsection{TERC Simulation Environment Summary}

The TERC simulation environment is built on top of the Pymunk physics simulator \cite{blomqvist2025pymunk}, and the implementation used in \citet{florence2022implicit}. However, we significantly alter the agent and environment dynamics to allow for a physics-based pick-and-place simulation, rather than a contact-rich pushing simulation. To recap previous sections. The action space of the agent is $a \in \mathbb{R}^2 \times \{0, 1\}$, defining position waypoints as well as the gripper state. Similar to base implementations, a low-level PD controller is used to control the agent position towards the waypoints. 

\subsection{Dataset Breakdown}

The dataset used in this paper consists of 35 physical reasoning tasks, broken down into a 'spatial' subset (1-25) and a 'spatial+algorithmic' subset (26-35). Specifically, the 'spatial' subset focuses mainly on using the counterfactual reasoning afforded by the rationality computation to discern between fine-grained goals, with simple algorithmic reasoning components for filtering objects by (functions of) their properties. The 'spatial+algorithmic' subset extends this task archetype with more complicated algorithmic reasoning requirements. This includes abstract tasks which move beyond being able to be represented with first-order-logic, and explicitly require the programmatic representation. A full breakdown of the tasks in natural language are provided in \cref{tab:tasklist}. 
\begin{table}[h]
  \centering
  \footnotesize                    % shrink text a bit (optional)
  \setlength{\tabcolsep}{4pt}      % tighten column spacing (optional)
  \caption{TERC Task List.}
  \label{tab:tasklist}
  %---------------------------------------------------------------
  % Two-column layout:  idx-desc | idx-desc
  %---------------------------------------------------------------
  \begin{tabular}{c p{0.38\linewidth} c p{0.38\linewidth}}
    \toprule
    \textbf{Idx} & \textbf{Task Description} &
    \textbf{Idx} & \textbf{Task Description} \\
    \midrule
     1 & Move the red circle to the top-right-corner
    & 19 & Move the largest circle to the bottom-right corner \\

     2 & Put every circle in the middle
    & 20 & Move the smallest object to the centre \\

     3 & Place the green boxes at the bottom
    & 21 & Move the largest box to the bottom-left \\

     4 & Bring all the triangles to the left
    & 22 & Move the purple square to the top-left-corner \\

     5 & Put the circles in the corners
    & 23 & Move the largest blue or green circle to the left \\

     6 & Put every box on the left
    & 24 & Move the largest yellow or green box to the top-right \\

     7 & Place all the pink shapes in the middle
    & 25 & Move green objects left, and blue objects to the right \\

     8 & Place all green objects at the top
    & 26 & Put the boxes at the top-left, then put the circles at the bottom-right \\

     9 & Move all the triangles to the bottom-left
    & 27 & First move all the circles to the top, then put everything else on the bottom \\

    10 & Move all the red objects to the bottom-left corner
    & 28 & First move all the rectangles to the left, then move all the squares to the right \\

    11 & Move the green box to the bottom-right corner
    & 29 & Move the red objects to the right, then move the green objects to the left \\

    12 & Place the orange triangle in the bottom-right-corner
    & 30 & First move the largest triangle to the top-right, then move the other triangles to the bottom-left \\

    13 & Place the green objects on the left and the blue objects on the right
    & 31 & Put the objects for which there exists the most type of on the left \\

    14 & Move all the squares to the top-left
    & 32 & Put the odd-one-out by color at the top-right corner \\

    15 & Move the yellow triangle to the top-right
    & 33 & If there is a triangle, move the circles to the top-right; otherwise move the boxes there \\

    16 & Move all the purple objects to the top-left corner
    & 34 & If the pink triangle exists, move it to the top-left corner; otherwise move the pink circle there \\

    17 & Move every object except the yellow one to the bottom
    & 35 & Sort the three objects by size: largest top-left, medium centre, smallest bottom-right (in that order) \\

    18 & Move the orange box to the bottom-left corner
    &     & \\             % empty cell to keep grid square
    \bottomrule
  \end{tabular}
\end{table}
Across these tasks we observe varying degrees of difficulty associated with: discerning the specificity of a goal (top, top-corner, top-right-corner etc), using atomic parameters to construct filters (square vs rectangle, biggest etc), relating objects in the environment to conditional behavior (if x do y etc), and whether or not the order of specific goals is significant (A then B or A and B). All of these tasks require fine-grained physical reasoning combined with algorithmic reasoning components at the abstract level. 

\subsection{Examples of generated code blocks for various tasks}

\begin{figure}
    \centering
    \includegraphics[width=1.0\linewidth]{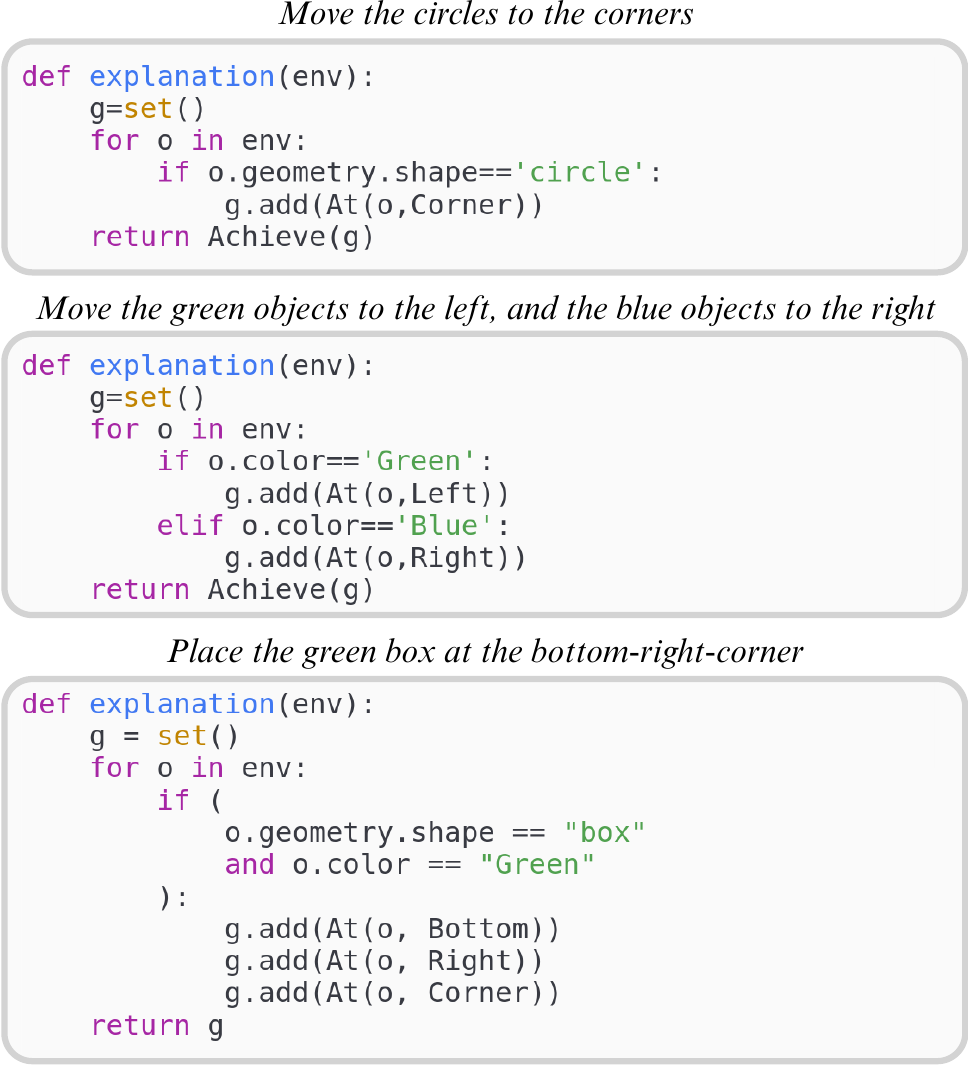}
    \caption{Tasks which require simple goal definition and composition. }
    \label{fig:supp_simple_goal}
\end{figure}

\begin{figure}
    \centering
    \includegraphics[width=1.0\linewidth]{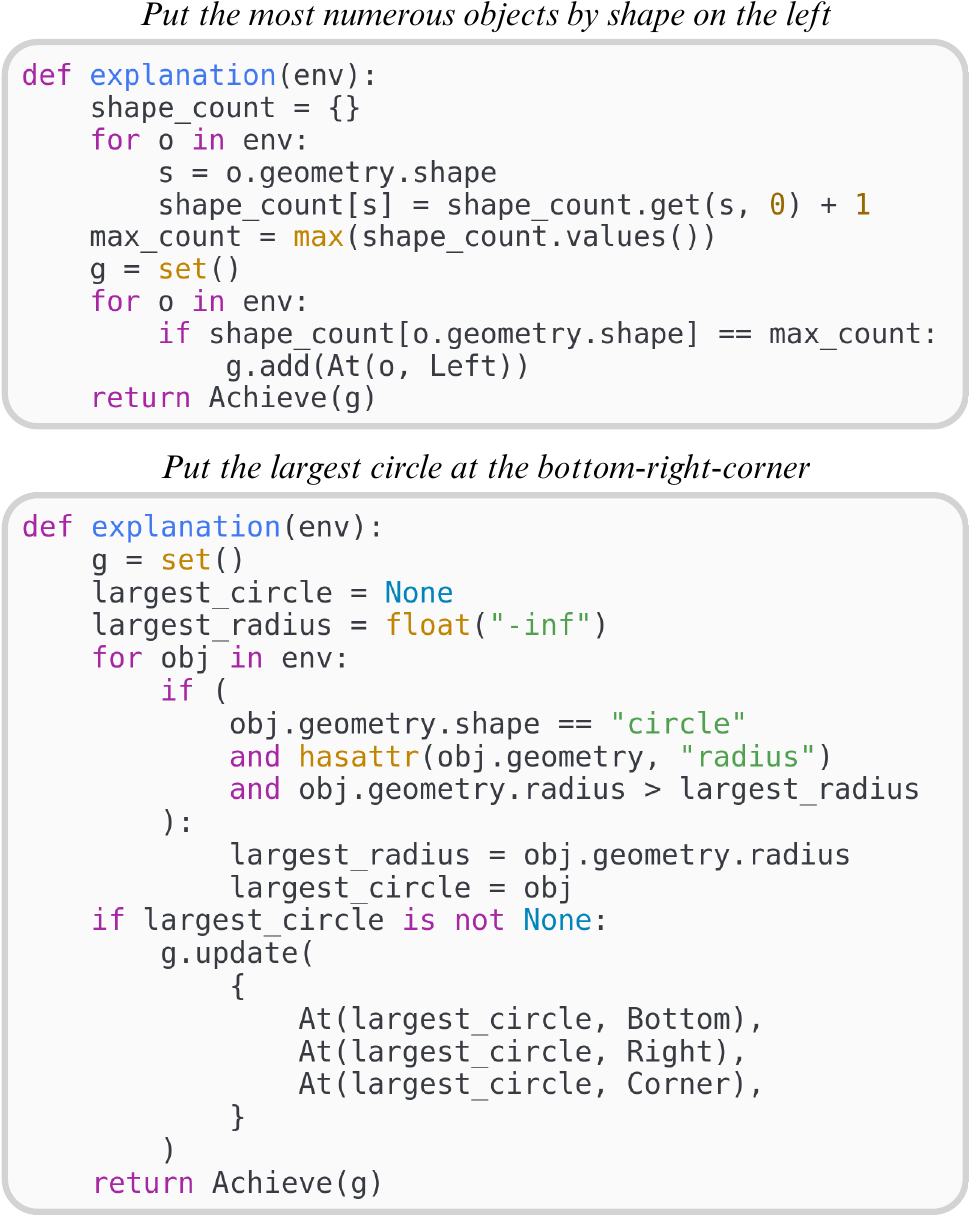}
    \caption{Tasks which require challenging goal definition and composition.}
    \label{fig:supp_hard_goal}
\end{figure}

\begin{figure}
    \centering
    \includegraphics[width=1.0\linewidth]{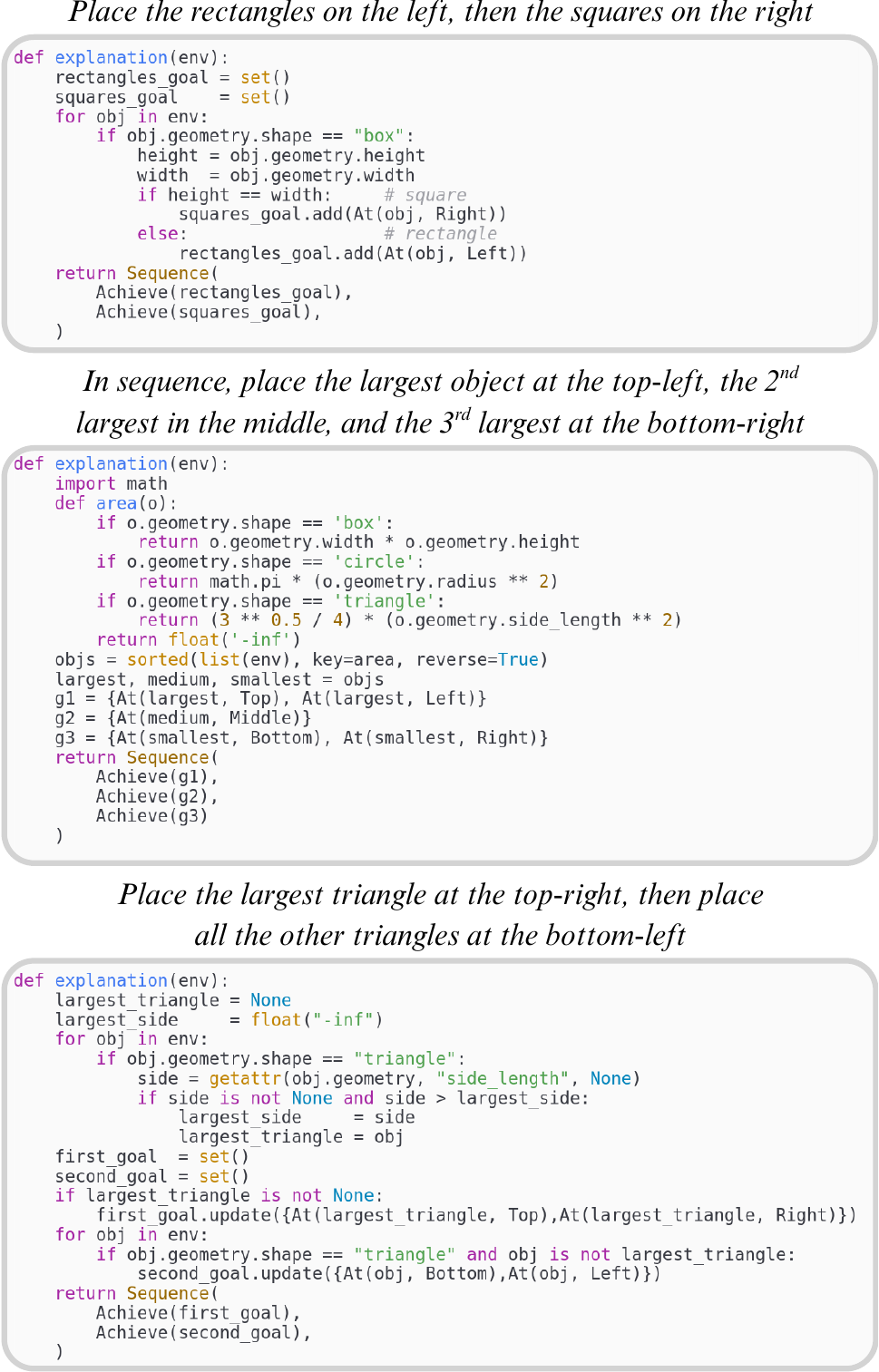}
    \caption{Tasks which require advanced goal definition, composition and temporal sequencing}
    \label{fig:supp_hard_goal_seq}
\end{figure}

To illustrate the range of hypotheses that the system must reason over, we outline some of the tasks in natural language, and the corresponding programmatic representation of $E$. For example, in \cref{fig:supp_simple_goal}, outline three examples of simple goal definition and composition. Specifically, these code snippets defined the following generalized tasks: $E_1=$\textit{'move the circles to the corners'}, $E_2=$\textit{'Move the green objects to the left, and the blue objects to the right'} and  $E_3=$\textit{'Place the green box at the bottom-right-corner'.} 
Other tasks require more challenging goal definition and composition. For example, \cref{fig:supp_hard_goal} shows two code snippets which describe the following tasks: $E_1=$\textit{'Put the most numerous objects by shape on the left'} and $E_2=$\textit{'Put the largest circle at the bottom-right-corner'}. Both of these cases require more complex filtering operations to be applied before task-spec construction; the first requires arithmetic and the max operation, whilst the second requires inequalities over numeric object properties. 
More complicated still are tasks that require the agent to perform complex physical and algorithmic reasoning, with the added difficulty of temporal and goal compositions. For example, consider \cref{fig:supp_hard_goal_seq}, which outlines the tasks: $E_1=$\textit{'Place the rectangles on the left, then the squares on the right'}, $E_2=$\textit{'In sequence, place the largest object at the top-left, the 2nd largest in the middle, and the 3rd largest at the bottom-right'} and $E_3=$\textit{'place the largest triangle at the top-right, then place all the other triangles at the bottom-left'}. These complex tasks require temporal and parallel compositions of the previous task archetypes, in addition to defining more advanced object/goal filtering mechanisms. 

\subsection{Comprehension and Success Evaluation Pipeline} \label{app:comprehension}

\begin{figure}
    \centering
    \includegraphics[width=1.0\linewidth]{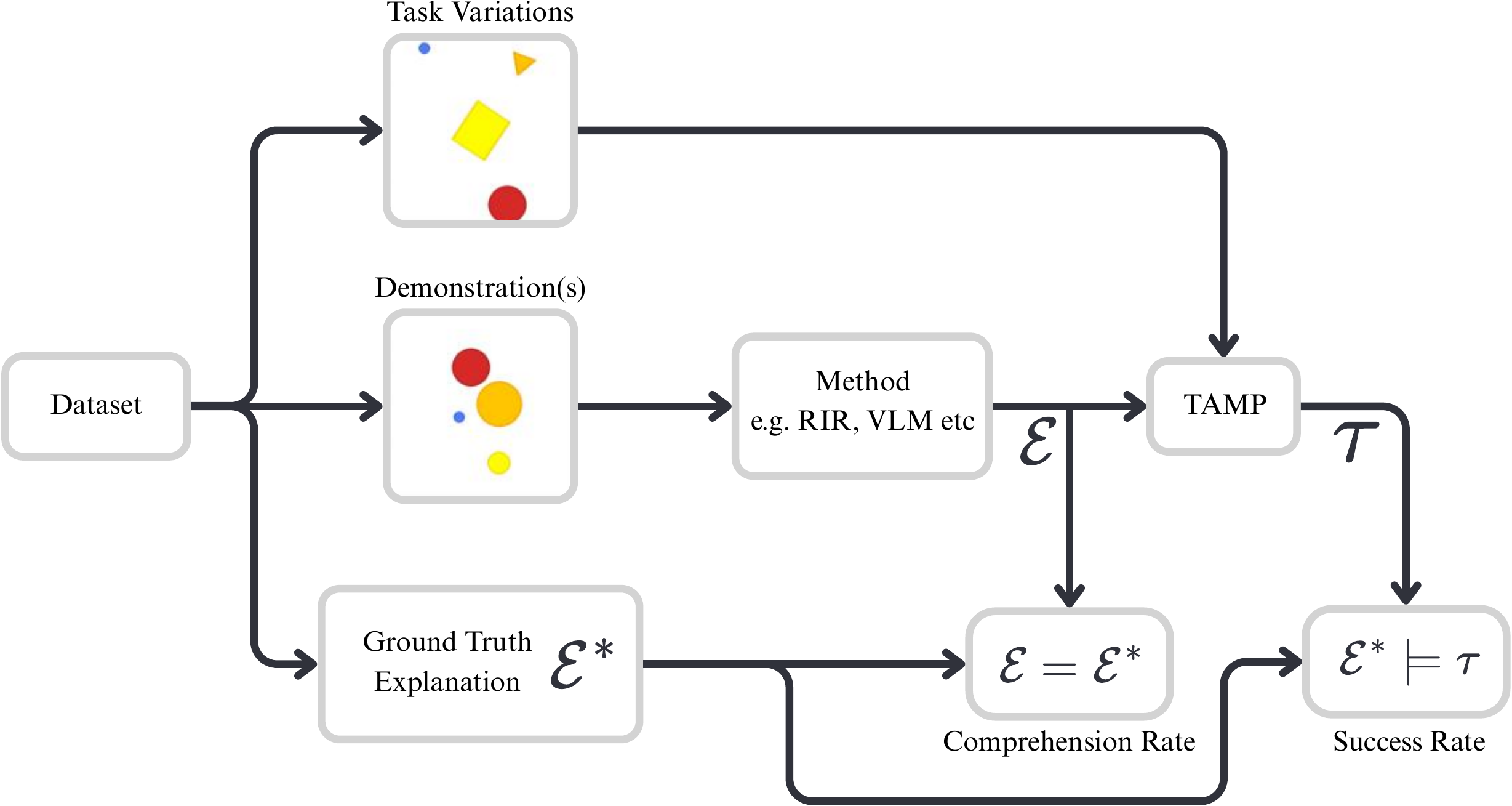}
    \caption{Pipeline diagram for the end-to-end evaluation pipeline.}
    \label{fig:supp_eval_pipeline}
\end{figure}

In this section, we outline the evaluation pipeline used for comprehension and success. Specifically, we leverage an LLM-based pipeline to evaluate the natural language or code-block explanations in comparison to ground-truth task descriptions (both natural language and code-blocks), and the TAMP solver with trajectory evaluation in the environment to detect success rate. The full pipeline for output generation and evaluation of these two metrics is shown in \cref{fig:supp_eval_pipeline}.

\subsubsection{LLM+Human Evaluation Pipeline} \label{app:evalpipe}

To extract comprehension rate, we leverage another LLM to compare the generated explanations in either natural language or in programmatic form to the ground-truth explanations (both natural language and example code snippets). The LLM is provided the same explanation function description (Listing~\ref{lst:function-definition-prompt}) and in-context examples (Listing~\ref{lst:in-context-prompt1} and Listing~\ref{lst:in-context-prompt2}) as the VLM hypothesis procedure. Specifically, we leverage the OpenAI GPT-4o model \cite{hurst2024gpt} for this evaluation pipeline. We apply the same evaluation procedure to both the generated explanation functions (from RIR or VLM-E), as well as the answers to the human survey. The LLM is encouraged to focus on evaluating overall functional equivalence, rather than exact wording/code formulation. We note that underlying semantics are anchored by an executable interface and can be audited with counterexample environments when eval uncertainty is high. The responses produced by the LLM are then verified by human evaluators, ensuring that the LLM is correct. 

\subsection{Human Visual Reasoning Survey} \label{app:human}

To contextualize VLM and RIR performance, we conducted a human survey measuring \emph{comprehension} on TERC. The intent is not to optimize human performance, but to establish that (i) the tasks are genuinely challenging and (ii) comprehension is imperfect even for humans when restricted to few demonstrations and when forced to articulate explicit explanations.

\paragraph{Survey structure}
Participants answered $70$ total questions, derived from the TERC task set and the same demonstrations provided to the VLM. Each question proceeded in two phases. First, the participant watched a single demonstration video and provided a natural language explanation of what they believed the task to be. Second, the participant watched two additional videos of the \emph{same} underlying task (3-shot total) and provided an updated explanation. The survey instructions emphasized precision and encouraged participants to state conditional and ordering structure explicitly when suspected.

\paragraph{Quality control and exclusions}
We collected responses from $24$ participants. One participant was removed due to survey misunderstanding, eliminating $70$ responses out of $1680$ total. The remaining responses were used to estimate human comprehension rates in the 1-shot and 3-shot settings.

\paragraph{Human comprehension evaluation}
Human responses are in natural language, while ground truth is a Python explanation function. To evaluate human comprehension consistently with VLM comprehension, we used the same LLM-based equivalence evaluator, prompting the evaluator with the ground-truth program and the human's natural language description and asking whether the description corresponds to the same task semantics. While imperfect, this evaluator is anchored by the formal DSL semantics and is used identically across methods, enabling consistent comparisons. We verify uncertain output with human annotators, ensuring that we correct LLM eval mistakes. 

\paragraph{Survey instruction excerpt}
The survey instructions are provided in Listing~\ref{lst:human-input-prompt}. In our released supplement, we include the full instruction block, including examples and clarifications about the meaning of regions such as \texttt{Corner} and about the strictness of \texttt{Sequence} ordering.

\subsection{TERC-specific likelihood instantiation and cost function}
\label{app:terc:likelihood}

The abstract likelihood used by RIR (Appendix~\ref{app:rir}) requires domain-specific choices for (i) a trajectory cost functional, (ii) compute budgets for enumerating and refining alternatives, and (iii) the domain specific strategy for bilevel TAMP planning. 

\paragraph{Trajectory cost}
In TERC we evaluate the trajectory distance cost:
\begin{equation}
C_{\text{traj}}(\tau)
\;=\;
\sum_{t=1}^{T} \|a_t - a_{t-1}\|_2^2 \;+\; \lambda \, |\hat\pi|,
\label{eq:terc_ctraj}
\end{equation}
where $a_t=(x_t,y_t,g_t)$ is the demonstrated action at time $t$, and $|\hat\pi|$ denotes the number of grasp events (equivalently, the number of pick/place mode switches) in the skeleton. The second term penalizes gratuitous grasping and empirically stabilizes inference by discouraging overly complex skeletons that match the demonstration only through excessive pick-place segments. In all TERC experiments we set $\lambda=80$.

\paragraph{Budgets and rationality hyperparameters.}
The plan-choice term in the likelihood requires enumerating alternatives and estimating their Boltzmann-normalized probabilities. In TERC, we implement this using a fixed compute budget and a Boltzmann rationality parameter over refinement free energies. We cap (i) the number of plan candidates refined per skeleton and (ii) the number of refinement iterations per planning call. For rationality estimation, we use a smaller budget than for full forward planning during evaluation, reflecting the fact that rationality estimation is performed many times per candidate explanation. We use $\beta_{\text{plan}} = 0.5$ for the Boltzmann plan-choice distribution. The full set of TERC hyperparameters is summarized in Table~\ref{tab:terc_hyperparams}.

\begin{table}[t]
\centering
\small
\setlength{\tabcolsep}{6pt}
\resizebox{\linewidth}{!}{%
\begin{tabular}{l c}
\toprule
\textbf{Parameter} & \textbf{Value (TERC)} \\
\midrule
Demo video rate & 10 Hz \\
Demo length & 80--120 frames \\
Action & $(x,y,\texttt{grip})$ \\
Trajectory penalty coefficient $\lambda$ (Eq.~\ref{eq:terc_ctraj}) & 80 \\
LCB confidence parameter $\delta$ (skeleton compute allocation) & 0.05 \\
Max plan candidates (rationality estimation) & 5 \\
Max solve iterations (rationality estimation) & 20 \\
Max plan candidates (full forward solve) & 10 \\
Max solve iterations (full forward solve) & 100 \\
Plan-choice inverse temperature $\beta_{\text{plan}}$ & 0.5 \\
Execution inverse temperature $\beta_{\text{traj}}$ & 1.0 \\
RIR iteration budget $T$ (default) & 3 \\
\bottomrule
\end{tabular}%
}
\caption{TERC-specific implementation hyperparameters. These values are fixed across all reported TERC results unless otherwise noted.}
\label{tab:terc_hyperparams}
\vspace{-2mm}
\end{table}

\subsection{Additional results: iteration budget and stopping criteria}
\label{app:terc:stopping}

RIR performs an iterative rationalization-generation loop. In TERC, the number of iterations $T$ is a central practical hyperparameter because each iteration requires multiple planner calls per program particle per demonstration, and allows the VLM to refine the hypothesis set. We therefore performed a targeted ablation to determine a robust default across task complexities.

\begin{figure}[h]
    \centering
    \includegraphics[width=1.0\linewidth]{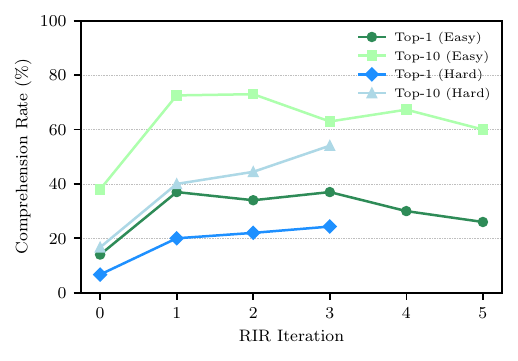}
    \caption{Average comprehension rate (Top-1/10) for varying RIR iterations and subset difficulty: spatial only (easy) vs spatial + algorithmic (hard).}
    \label{fig:supp_iter_rir}
\end{figure}

\paragraph{Fixed iteration choice ($T=3$).}
Empirically, $T=3$ iterations worked well across a range of program complexities. For simple spatial programs, a single round of rationalization often produced large gains because feasibility and goal restrictiveness were quickly corrected. For algorithmic tasks (Tasks 26--35), additional iterations were essential to refine program structure (e.g., introducing \texttt{Sequence}, correctly selecting maxima, or handling conditional branches), with $T=3$ providing a useful tradeoff between inference time and comprehension across all subsets. \cref{fig:supp_iter_rir} illustrates the trend in top-1 and top-10 comprehension as a function of iteration count.

\paragraph{Adaptive stopping attempts}
We also explored preliminary adaptive stopping heuristics based on stabilization of the candidate set across iterations. However, the per-trial variation in how many iterations were required to reach a stable estimate of the posterior was large, making such heuristics unreliable. A more principled direction is to stop when an estimate of the log marginal likelihood (or a related evidence lower bound) stabilizes; we leave this for future work.

\subsection{Reproducibility notes for TERC experiments}
\label{app:terc:repro}

\paragraph{VLM and baseline configuration.}
For TERC, our baseline explanation generator is a multimodal reasoning VLM equipped with structured prompting (Gemini-2.5-Pro in the original experiments). The first iteration of RIR uses the same model and prompt; improvements beyond the baseline are therefore attributable to the embodied rationalization loop rather than to differences in model class. The baseline ``reasoning without rationality'' variant (R w/o R) uses self-refinement only and does not incorporate embodied plan rationality signals, though it is provided the same scene reconstruction and optional bottom-up plan proposals for fairness.

\paragraph{Implementation and compute.}
All TERC experiments require repeated planning calls. To ensure tractability, we rely on caching of refinement samples and on fixed compute budgets summarized in Table~\ref{tab:terc_hyperparams}. We also provide top-$N$ metrics to expose posterior concentration and to better understand when the correct explanation appears among high-probability candidates even if the MAP is incorrect.

\section{Reconstructing Demonstrations with Real to Sim} \label{app:real2sim}

In this section we detail the Real2Sim reconstruction pipelines used to convert the raw scene observations $x$, consisting of multi-view (2) RGB-D inputs and robot joints, into an object-centric scene representation used for downstream reasoning and planning. The system transforms multi-view stereo RGB-D observations of tabletop manipulation scenes into complete simulation-ready 3D reconstructions, including per-entity watertight meshes, semantic annotations, spatial predicates, and temporal tracking of entity poses through robot demonstrations. We describe every component of the pipeline in full detail, including all algorithmic choices, hyperparameter settings, model configurations, and error handling strategies. We begin with an overview of the system architecture before providing detailed descriptions of each component in the order they are executed.

\subsection{System Overview} \label{app:real2sim-overview}

The Real2Sim pipeline is a modular, DAG-based system composed of seven core components, each implemented as a self-contained pipeline stage with typed inputs, outputs, caching, and profiling. The pipeline operates in two primary modes: (1)~\emph{initial frame reconstruction}, used at test time to produce a full 3D scene representation from a single multi-view stereo frame pair; and (2)~\emph{demonstration reconstruction and tracking}, used at training time to process an entire manipulation demonstration, reconstructing the initial scene and then tracking all entity poses through the demonstration using robot gripper state.
The component execution order follows a strict dependency DAG:

\begin{enumerate}
    \item \textbf{VLM Scene Parsing} (Section~\ref{app:real2sim-vlm}): Detects and labels all entities in multi-view images using a Vision-Language Model (VLM), producing canonical entity identifiers, bounding boxes, keypoints, semantic tags, support relations, and semantic part annotations.
    \item \textbf{SAM2 Segmentation} (Section~\ref{app:real2sim-sam2}): Segments each detected entity using the Segment Anything Model 2 (SAM2), prompted with VLM-provided bounding boxes and keypoints to produce per-entity binary masks in each camera view.
    \item \textbf{Neural Stereo Depth} (Section~\ref{app:real2sim-depth}): Enhances raw sensor depth using FoundationStereo, a learned stereo matching network, to produce dense, sub-millimeter depth maps from calibrated stereo image pairs.
    \item \textbf{Entity Generation} (Section~\ref{app:real2sim-entity-gen}): Combines segmentation masks and enhanced depth to backproject masked regions into 3D point clouds, fuses multi-view observations, extracts table geometry, and populates a world manager with semantically-annotated entity objects including 3D semantic part keypoints.
    \item \textbf{Mesh Generation} (Section~\ref{app:real2sim-meshing}): Generates per-entity watertight collision meshes using GPU-accelerated TSDF fusion followed by CGAL alpha wrapping, with optional Taubin smoothing for improved grasp sampling coverage.
    \item \textbf{State Extraction} (Section~\ref{app:real2sim-state}): Extracts the complete symbolic world state, including geometric predicates (on, above), semantic predicates (in, is\_container), and cross-validates geometric and VLM-derived relations.
    \item \textbf{Robot-Driven Tracking} (Section~\ref{app:real2sim-tracking}): Tracks entity poses through a demonstration using the robot gripper state, with mesh-based grasp detection, contact-based centering, a gripper state machine, and dynamic predicate maintenance.
\end{enumerate}

\subsection{VLM Scene Parsing}
\label{app:real2sim-vlm}

The VLM scene parsing component is the entry point of the pipeline and serves as the source of truth for entity identifiers throughout the system. 

\subsubsection{Model and API}

The system uses Google's Gemini family of VLMs. The default model is \texttt{gemini-3-flash-preview}, selected for its strong spatial reasoning capabilities and structured output support. The VLM is wrapped in a thin client (\texttt{GeminiVLM}) that supports both synchronous and asynchronous inference, structured JSON output with Pydantic schema validation, and batch processing with configurable concurrency.

\subsubsection{Two-Step Detection with Semantic Parts}

The VLM operates in a two-step mode that decouples entity detection from semantic part labeling:

\paragraph{Step 1: Entity Detection.} The VLM receives multi-view images (typically a front and back pair) and is prompted to label all individual objects on the table. The prompt includes several critical instructions for multi-view correspondence: Multiple objects of the same type must be labeled separately with unique suffixes (e.g., \texttt{wooden\_spoon\_1}, \texttt{wooden\_spoon\_2}). The table itself should be labeled with a keypoint on its surface but no bounding box.

For each detected entity, the VLM returns: a normalized label, a longer description, a confidence score in $[0, 1]$, a 2D keypoint $[y, x]$ normalized to $[0, 1000]$, and a bounding box $[\text{ymin}, \text{xmin}, \text{ymax}, \text{xmax}]$ also normalized to $[0, 1000]$.

\paragraph{Keypoint Placement Guidance.} The prompt includes critical guidance for keypoint placement that directly affects downstream segmentation quality. Keypoints must be placed on the actual visible surface of the object, not at the bounding box center (which may be empty space or another object). For objects with holes (mugs, bowls), the keypoint must be on solid material, not the opening. For stacked objects, the keypoint for the bottom object must be on an exposed portion not occluded by the top object.

\paragraph{Step 2: Semantic Part Labeling.} A follow-up prompt requests fine-grained semantic parts for entities that require them. This step uses a part taxonomy that defines known part types (e.g., \texttt{handle}, \texttt{tool\_tip}, \texttt{opening\_rim}, \texttt{spout}, \texttt{blade}) and includes guidelines for minimalism (most objects need zero or one part) and projection safety (parts must be on visible surfaces). For each part, the VLM returns: a part name, a 2D keypoint, and a confidence score. Additionally, the VLM provides support relation annotations: for each entity, it reports the \texttt{support\_surface} (what the entity rests on, e.g., ``table'' or ``bowl\_1'') and the \texttt{support\_relation} (exactly one of 'on' or 'in'), along with semantic tags such as 'container' or 'graspable.'. This is used later in the state extraction (\cref{app:real2sim-state}) to verify the geometric state extraction. 

Table~\ref{tab:vlm_params} summarizes all hyperparameters for the VLM component.

\begin{table}[h]
\centering
\small
\caption{VLM Scene Parsing hyperparameters.}
\label{tab:vlm_params}
\resizebox{\linewidth}{!}{%
\begin{tabular}{lll}
\hline
\textbf{Parameter} & \textbf{Default} & \textbf{Description} \\
\hline
\texttt{model} & \texttt{gemini-3-flash-preview} & VLM model identifier \\
\texttt{temperature} & 0.1 & Generation temperature \\
\texttt{max\_output\_tokens} & 16384 & Maximum response tokens \\
\texttt{thinking\_budget} & 0 & Thinking tokens (0 = disabled) \\
\texttt{two\_step} & True & Two-step part labeling \\
\texttt{max\_retries} & 3 & Retries on API failure \\
\texttt{articulation\_detection} & False & Enable articulation VLM \\
\hline
\end{tabular}%
}
\end{table}

\subsection{SAM2 Segmentation}
\label{app:real2sim-sam2}
The segmentation component uses the Segment Anything Model 2 (SAM2) to produce per-entity binary masks in each camera view, using the VLM-provided bounding boxes and keypoints as prompts. The SAM2 hyperparameters are outlined in \cref{tab:sam2_params}.
\begin{table}[h]
\centering
\small
\caption{SAM2 Segmentation hyperparameters.}
\label{tab:sam2_params}
\resizebox{\linewidth}{!}{%
\begin{tabular}{lll}
\hline
\textbf{Parameter} & \textbf{Default} & \textbf{Description} \\
\hline
\texttt{checkpoint} & \texttt{sam2\_hiera\_large} & SAM2 model checkpoint \\
\texttt{points\_per\_side} & 32 & Points for auto mask generation \\
\texttt{multimask\_output} & True & Output multiple masks per prompt \\
\texttt{use\_keypoint\_only\_for\_tables} & True & Keypoint-only for table entities \\
\texttt{keep\_server} & True & Keep SAM2 server alive \\
\hline
\end{tabular}%
}
\end{table}

\subsection{Neural Stereo Depth Enhancement}
\label{app:real2sim-depth}

The depth component replaces raw sensor depth with neural stereo depth to achieve sub-millimeter accuracy for precise 3D reconstruction. The system uses \textbf{FoundationStereo}, a learned stereo matching network, as the primary depth estimation model. FoundationStereo takes a calibrated stereo image pair (left RGB and right RGB with known baseline and intrinsics) and produces a dense disparity map, which is converted to metric depth using the camera baseline and focal length. The depth enhancement pipeline operates as follows:
\begin{enumerate}
    \item For each camera in the frame bundle, check whether a stereo pair is available. If no stereo pair exists, fall back to the raw sensor depth (from raw ZED-2i).
    \item Submit the stereo pair to the FoundationStereo server
    \item Critically, the system preserves the \emph{raw sensor confidence maps} from the ZED neural depth before enhancement. The raw confidence is preferred over the neural network's own confidence because the sensor confidence captures measurement-specific noise characteristics (e.g., reflective surfaces, edges) that are lost in the neural depth output. These confidence maps are used downstream for filtering during backprojection.
\end{enumerate}
The hyperparamters for this component are outlined in \cref{tab:depth_params}.
\begin{table}[h]
\centering
\small
\caption{Stereo Depth Enhancement hyperparameters.}
\label{tab:depth_params}
\resizebox{\linewidth}{!}{%
\begin{tabular}{lll}
\hline
\textbf{Parameter} & \textbf{Default} & \textbf{Description} \\
\hline
\texttt{model} & \texttt{foundation\_stereo} & Depth model identifier \\
\texttt{scale} & 1.0 & Depth scale factor \\
\texttt{hierarchical} & False & Hierarchical inference \\
\texttt{min\_depth} & 0.01~m & Minimum valid depth \\
\texttt{max\_depth} & 2.0~m & Maximum valid depth \\
\texttt{confidence\_threshold} & 0.3 & Confidence filter threshold \\
\texttt{keep\_server} & False & Keep server alive \\
\hline
\end{tabular}%
}
\end{table}

\subsection{Entity Generation}
\label{app:real2sim-entity-gen}

The entity generation component is the core stage that transforms 2D perception outputs into a 3D world representation. It performs four sequential steps: depth application, masked backprojection, multi-view fusion, and world building.
\subsubsection{Entity Point Reconstruction}

The first step applies the neural stereo depth to the frame bundle. For each camera view, the enhanced depth map and confidence map from the depth component replace the raw sensor values in the frame data structure. This produces an enhanced frame bundle that is used for all subsequent 3D operations. 

The enhanced and masked RGBD frames are back-projected into the 3D scene, where observations from all camera views for each entity are fused into a single entity reconstruction:
\begin{enumerate}
    \item All 3D points from all views are concatenated into a single array. Colors are similarly concatenated.
    \item The merged point cloud is voxel-downsampled using Open3D with a voxel size of $\texttt{voxel\_size} = 0.003$~m (3~mm). This both regularizes the point density and reduces memory requirements.
    \item If \texttt{outlier\_removal=True} (default), statistical outlier removal is applied with \texttt{nb\_neighbors=20} and \texttt{std\_ratio=2.0}. This removes isolated noisy points that may result from depth errors at object boundaries.
    \item Entities with fewer than \texttt{min\_points=100} points after fusion are discarded, as they lack sufficient geometric information for reliable reconstruction.
\end{enumerate}

\subsubsection{World Building}

\paragraph{Entity Creation} A rigid body is created from the fused point cloud, which computes: the entity centroid, an axis-aligned bounding box (AABB), a convex hull (if configured), and an oriented bounding box (OBB). When configured and/or detected (default: \texttt{False}), the OBB $z$-axis is constrained to align with the world $z$-axis, which produces better oriented bounding boxes for objects resting on flat surfaces - guided by the VLM on-table extraction.

\paragraph{Detection Info Attachment} VLM-derived semantic metadata is attached to each entity containing: per-camera bounding boxes and keypoints, confidence scores, the VLM label and description, support surface and relation annotations, semantic tags (e.g., 'container', 'graspable'), and per-view part detections. When the same entity is detected in multiple views, the system aggregates information: confidence is taken as the maximum across views, semantic tags are merged, and the first non-null support relation is preserved.

\paragraph{3D Semantic Part Keypoints} For each entity with VLM-detected semantic parts (e.g., handle, spout, tool-tip), the component backprojects the 2D part keypoints to 3D using the enhanced depth maps. The process:
\begin{enumerate}
    \item Groups part observations by name across all camera views.
    \item For each part and each camera view with a valid 2D keypoint, backprojects to 3D using the same pinhole model as entity backprojection. If the depth at the exact keypoint pixel is invalid, a search window of radius 5 pixels is used to find the nearest valid depth, taking the median of valid depths in the window.
    \item Averages the 3D keypoint positions from all valid views for robustness.
    \item Computes the part's relative pose as a pure translation offset from the entity centroid to the 3D keypoint position.
\end{enumerate}
The resulting semantic parts are stored on the entity and are used downstream for tracking and constraint application in planning (see \cref{app:robot_exp} (keypoints move rigidly with the entity) and for scene prompt generation.

Hyperparameters for this step are outlined in \cref{tab:entity_gen_params}.

\begin{table}[h]
\centering
\small
\caption{Entity Generation hyperparameters.}
\label{tab:entity_gen_params}
\resizebox{\columnwidth}{!}{%
\begin{tabular}{lll}
\hline
\textbf{Parameter} & \textbf{Default} & \textbf{Description} \\
\hline
\texttt{erosion\_pixels} & 5 & Mask erosion before backprojection \\
\texttt{confidence\_threshold} & 0.3 & Depth confidence filter threshold \\
\texttt{min\_points} & 100 & Min points per entity after fusion \\
\texttt{voxel\_size} & 0.003~m & Voxel size for downsampling \\
\texttt{outlier\_removal} & True & Statistical outlier removal \\
\texttt{compute\_hull} & True & Compute convex hull \\
\texttt{compute\_obb} & True & Compute oriented bounding box \\
\texttt{obb\_z\_up} & False & Constrain OBB $z$-axis to world $z$ \\
\texttt{enable\_table\_extraction} & True & Extract table geometry \\
\texttt{table\_xy\_margin} & $-0.1$~m & XY margin for table bounds \\
\hline
\end{tabular}%
}
\end{table}

\subsection{Mesh Generation}
\label{app:real2sim-meshing}

The mesh generation component produces per-entity watertight collision meshes suitable for simulation, grasp generation and motion planning. It follows a two-stage geometry strategy: (1)~observed geometry via masked TSDF fusion to capture surface details and concavities; (2)~robust watertight envelope via CGAL alpha wrapping to enforce manifoldness and suppress artifacts. 

For each entity, the component constructs a Truncated Signed Distance Function (TSDF) volume from multi-view depth observations and extracts a triangle mesh. After extraction, small disconnected components are removed. 

The TSDF mesh may contain holes, self-intersections, and non-manifold edges due to incomplete observations and depth noise. Alpha wrapping produces a watertight envelope around the observed geometry. The watertight mesh is decimated to a target triangle count of \texttt{collision\_target\_triangles} (default: $8000$) using quadric edge collapse. This reduces the computational cost of collision checking in downstream motion planning without significantly affecting collision accuracy. 

The mesh generation hyperparameters are outlined in \cref{tab:mesh_params}.

\begin{table}[h]
\centering
\small
\caption{Mesh Generation hyperparameters}
\label{tab:mesh_params}
\resizebox{\columnwidth}{!}{%

\begin{tabular}{lll}
\hline
\textbf{Parameter} & \textbf{Default} & \textbf{Description} \\
\hline
\texttt{voxel\_size\_m} & 0.002~m & TSDF voxel size \\
\texttt{weight\_threshold} & 0.5 & TSDF weight threshold \\
\texttt{depth\_max\_m} & 2.0~m & Max integration depth \\
\texttt{skip\_table} & True & Skip table meshing \\
\texttt{offset} & 0.001~m & Alpha wrap offset \\
\texttt{alpha} & 0.035 & Alpha wrap alpha \\
\texttt{collision\_target\_triangles} & 8000 & Target triangle count \\
\texttt{smooth\_mesh} & False & Enable Taubin smoothing \\
\texttt{smooth\_max\_edge} & 0.02~m & Max edge for subdivision \\
\texttt{smooth\_taubin\_iterations} & 3 & Smoothing iterations \\
\texttt{smooth\_taubin\_lamb} & 0.5 & Smoothing $\lambda$ \\
\texttt{smooth\_taubin\_nu} & 0.53 & Anti-shrink $\nu$ \\
\hline
\end{tabular}%
}
\end{table}

\subsection{State Extraction}
\label{app:real2sim-state}

The state extraction component produces the complete symbolic world state by combining geometric and semantic reasoning. It extracts per-entity states, spatial predicates, and cross-validates geometric and VLM-derived relations.

\subsubsection{Per-Entity State}

For each entity in the scene graph, the component extracts: the entity's $SE(3)$ pose (from the entity's centroid and/or OBB), detection confidence, point count, mesh availability, and semantic properties (container status, support surface, support relation, parts) derived from VLM annotations. Semantic information is consolidated across views via majority voting: the most frequently observed (surface, relation) pair is selected.

\subsubsection{Predicate Extraction}

The perception system extracts four types of predicates:

\paragraph{Exists Predicates} A unary \texttt{exists}$(e)$ predicate is created for every valid entity in the scene. 

\paragraph{On Predicates (Geometric)} The geometric \say{on} detector uses a multi-phase approach to determine support relationships:
\begin{enumerate}
    \item \textbf{Coarse Phase (AABB Band Test)}: For each pair of entities $(A, B)$, check whether $A$'s AABB bottom is within a contact band of $B$'s AABB top. The contact band is defined by \texttt{gap\_max} (default: $0.01$~m = 1~cm) for the maximum gap and \texttt{pen\_tol} (default: $0.005$~m = 5~mm) for allowed penetration. Both entities must also overlap in the $xy$ plane within \texttt{tol\_xy} (default: $0.005$~m).
    \item \textbf{Fine Phase (Convex Hull XY Intersection)}: If the coarse check passes, the convex hulls of both entities are projected onto the $xy$ plane and their intersection area is computed. If the intersection area divided by $A$'s hull area exceeds \texttt{min\_overlap\_ratio\_a} (default: $0.0$), the pair passes the fine check.
    \item \textbf{Contact Refinement (Raycast Sampling)}: When \texttt{contact\_refine=True} (default), the system samples \texttt{samples=256} random points within the $xy$ overlap region and casts rays downward to determine what fraction of $A$'s footprint is actually supported by $B$'s surface. A pair is accepted if the support ratio exceeds \texttt{min\_support\_ratio=0.1}.
\end{enumerate}

Optionally, when entity meshes are available, the \say{on} detector can use mesh geometry instead of point clouds (+convex hull) for more accurate contact computation.

\paragraph{In Predicates (Semantic).} Containment predicates are extracted from VLM annotations. The VLM reports a \texttt{support\_relation} of ``in'' when an entity is inside a container. These are extracted by parsing the per-view detections and consolidating across views. This is used in combination with a similar verification check to the \say{On} predicate, verifying with AABB intersections. 

\paragraph{IsContainer Predicates.} Container predicates are derived from VLM semantic tags. When a detection includes \say{container} in its \texttt{semantic\_tags} list, the corresponding entity receives an \texttt{is\_container} predicate.

\subsubsection{Cross-Validation}

Validating geometric predicates against VLM-derived support relations is critical. When there is a conflict e.g. the geometric detector says $A$ is \say{on} $B$, but the VLM says $A$ is \say{in} $B$, the \say{in} relation supersedes (when \texttt{vlm\_supersedes\_geometric=True}, which is the default). This is important because objects inside containers may geometrically appear to be \say{on} the container's bottom surface, but manipulating them requires different strategies (e.g., scooping from a bowl rather than picking from a surface). 

The hyperparameters for state extraction are shown in \cref{tab:state_params}

\begin{table}[h]
\centering
\small
\caption{State Extraction hyperparameters.}
\label{tab:state_params}
\resizebox{\columnwidth}{!}{%
\begin{tabular}{lll}
\hline
\textbf{Parameter} & \textbf{Default} & \textbf{Description} \\
\hline
\texttt{predicates} & [exists, on, in, is\_container] & Predicates to extract \\
\texttt{contact\_threshold} (gap\_max) & 0.01~m & On-contact gap tolerance \\
\texttt{above\_threshold} & 0.05~m & Above relation threshold \\
\texttt{validate\_geometric\_vs\_vlm} & True & Cross-validate predicates \\
\texttt{vlm\_supersedes\_geometric} & True & VLM ``in'' overrides geometric ``on'' \\
\texttt{extract\_containers} & True & Extract IsContainer from tags \\
On detector: \texttt{pen\_tol} & 0.005~m & Penetration tolerance \\
On detector: \texttt{samples} & 256 & Raycast sample count \\
On detector: \texttt{min\_support\_ratio} & 0.1 & Min support fraction \\
\hline
\end{tabular}%
}
\end{table}

\subsection{Robot-Driven Tracking}
\label{app:real2sim-tracking}

The tracking component processes entire manipulation demonstrations by tracking entity poses over time using the robot gripper state. This is a streaming component that processes frames sequentially, unlike the single-frame components described above. The system implements a \emph{robot-only tracking} paradigm based on two key assumptions:
\begin{enumerate}
    \item \textbf{Grasped objects move rigidly with the gripper.} When the gripper grasps an entity stably, the entity's pose is updated at each frame by composing the gripper's world-frame pose with the fixed relative transform between the gripper and entity: $T_{\text{world} \leftarrow \text{entity}}^{(t)} = T_{\text{world} \leftarrow \text{gripper}}^{(t)} \cdot T_{\text{gripper} \leftarrow \text{entity}}$, where the relative transform $T_{\text{gripper} \leftarrow \text{entity}}$ is computed at the moment of grasp and remains constant until release.
    \item \textbf{Ungrasped objects remain static.} Objects not currently held by the gripper stay at their last known pose. This is valid for tabletop manipulation where objects rest stably on surfaces and do not move unless manipulated. That is, we maintain a quasi-static assumption on the environment. 
\end{enumerate}

This paradigm eliminates the need for frame-by-frame visual object tracking (which is fragile for small objects under significant occlusion) and instead leverages the precise robot state available from joint encoders, with occasional updates to verify our quasi-static assumption. 

\subsection{Typical Timing}

Table~\ref{tab:timing} provides typical timing for each pipeline stage on representative tabletop scene (reconstructions during real-robot evaluation) with 4-10 entities, using a GPU-equipped workstation (RTX 5090), across 10 runs each (with and without server warm-up).

\begin{table}[h]
\centering
\small
\caption{Typical per-frame timing for pipeline stages. Times measured across multiple representative tabletop scenes with 4-10 entities using two stereo camera pairs. Server initialization times are one-time costs amortized across multiple runs.}
\label{tab:timing}
\resizebox{\columnwidth}{!}{%
\begin{tabular}{lr}
\hline
\textbf{Stage} & \textbf{Time (seconds)} \\
\hline
VLM Scene Parsing (Gemini API) & 5--20 \\
SAM2 Segmentation (per camera) & 2--5 \\
Neural Stereo Depth (per camera pair) & 2--5 \\
Entity Generation (backproject + fuse) & 0.5--2 \\
Mesh Generation (per entity, quality preset) & 0.5--2 \\
State Extraction & $<$1 \\
\hline
\textbf{Total (initial frame, uncached)} & \textbf{32.63} \textbf{(29-35)} \\
\textbf{Total (initial frame, uncached, w/o warmup)} &  \textbf{50.61} \textbf{(46--56)} \\
\textbf{Total (cached, state extraction only)} & \textbf{$<$10} \\
\hline
Tracking (per frame) & $<$0.01 \\
Tracking (full segment, $\sim$300 frames) & 2--5 \\
\hline
SAM2 server initialization (one-time) & 5--10 \\
FoundationStereo server initialization (one-time) & 5--15 \\
\hline
\end{tabular}%
}
\end{table}

\section{Real-Robot Task and Motion Planning Implementation} \label{app:stamp}

For our real robot experiments, we use a particle-based sequential TAMP solver that combines symbolic planning with GPU-accelerated continuous constraint satisfaction. Given a reconstructed 3D scene (produced by the Real2Sim pipeline described in Appendix~\ref{app:real2sim}) and a task specification, TAMP generates collision-free, kinematically feasible manipulation trajectories. We describe every component, algorithm, and hyperparameter in full detail.

Note that this implementation differs from the generalized TAMP implementation for the TERC experiments, as scaling to real robot experiments with more complicated tasks requires a more sophisticated and faster solver.  

\subsection{System Overview}
\label{app:stamp-overview}

The TAMP system operates as a three-layer pipeline:

\begin{enumerate}
    \item \textbf{Task Specification} (\cref{app:stamp-task-spec}): An explanation program queries the reconstructed scene and produces a grounded task specification $g$ (e.g., \say{place all red blocks in the bowl}), along with optional geometric constraints (e.g., \say{fork tines should point upward when dropped}).
    \item \textbf{Symbolic Planning} (Section~\ref{app:stamp-symbolic}): A PDDL-style planner enumerates candidate plan skeletons (sequences of high-level skills such as \textsc{Pick}, \textsc{Place}, and \textsc{Drop}) that achieve the goal from the initial symbolic state.
    \item \textbf{Continuous CCSP Solving} (Section~\ref{app:stamp-ccsp}): A particle-based sequential solver instantiates each skeleton by sampling and filtering continuous parameters (grasps, placements, IK solutions, trajectories) using importance-weighted particles with systematic resampling.
\end{enumerate}

The planner iterates over candidate skeletons (up to \texttt{max\_skeletons}$=10$, with a timeout of $60\,\text{s}$) and returns the first skeleton for which a feasible continuous solution is found at test time. 

\subsection{Task Specification}
\label{app:stamp-task-spec}

Explanation programs are Python functions that receive a reconstructed environment and return a task specification in terms of goals, subgoals and constraints. The environment exposes methods for querying objects (for example, \texttt{get\_pickable\_objects()}, \texttt{get\_containers()}, \texttt{filter\_by\_label()}), accessing metadata (VLM-derived labels, descriptions, semantic tags, keypoint parts), and constructing goals.

\subsubsection{Goal Types}

The system supports two goal structures:

\begin{itemize}
    \item \textbf{Achieve}$(\cdot)$: Achieve a set of predicates $\mathcal{G}$ simultaneously (single-stage planning).
    \item \textbf{Sequence}$(\cdot, ..., \cdot)$: Achieve subgoals in strict order (multi-stage planning). Each stage $k$ must be fully satisfied before the planner proceeds to stage $k+1$.
\end{itemize}

\subsubsection{Supported Predicates}

\begin{itemize}
    \item \texttt{On}$(o, s)$: Object $o$ is stably placed on surface $s$.
    \item \texttt{InContainer}$(o, c)$: Object $o$ has been dropped into container $c$.
    \item \texttt{Holding}$(o)$: The robot is currently grasping object $o$.
\end{itemize}

\subsubsection{External Constraints}

Task programs can register geometric constraints that are enforced during continuous solving. We manually provide (in-context and in the api) the following examples, however the system is free to write its own constraints based on the hierarchical object-centric state representation presented in \cref{app:real2sim}:

\paragraph{DropOrientationConstraint} Constrains the orientation of a dropped object so that a specified semantic part (e.g., fork tip, spoon bowl) is either \emph{upright} (part's world-frame $z$-coordinate above the object centroid) or \emph{downward} (below centroid). The constraint is evaluated by transforming the part's keypoint offset $\mathbf{o}_\text{part} \in \mathbb{R}^3$ (in object frame) by the candidate drop pose's rotation matrix $R$ and checking $\text{sign}((R \cdot \mathbf{o}_\text{part})_z - (R \cdot \mathbf{0})_z)$.

\paragraph{PlaceOrientationConstraint} Similar to the DropOrientationConstraint, but filtering placements instead.

% \paragraph{LayoutConstraint.} Constrains placements to match a learned spatial layout. Given a set of pairwise XY offsets $\{\Delta_{ij}\}$ between object keypoints (centroids and semantic parts) observed in a demonstration, each candidate placement is accepted only if the L2 deviation from the learned offset is below a threshold $\tau_\text{L2}$ (default $0.05\,\text{m}$). Formally, for a candidate placement of object $i$ at position $\mathbf{p}_i$ with already-placed objects $\{j\}$ at positions $\{\mathbf{p}_j\}$:
% \begin{equation}
%     \text{accept} \iff \forall\, (i,j) \in \mathcal{E}: \; \|\mathbf{p}_i^{xy} - \mathbf{p}_j^{xy} - \Delta_{ij}\|_2^2 \leq \tau_\text{L2}^2
% \end{equation}
% Poses exceeding the threshold are hard-rejected; those within the threshold incur a soft cost proportional to $10 \times \text{layout\_score}$.

\subsection{Planning Domain}
\label{app:stamp-symbolic}

The planning domains follows the structure outlined in the formal description of generalized Task-and-Motion Planning (\cref{app:tamp}). Here, we ouline the specific types, predicates, constraints, samplers and skills used for planning. 

\subsubsection{Types}

The domain defines the following types, split between the base domain and the extended domain:

\begin{table}[h]
\centering
\caption{Planning domain types.}
\label{tab:domain_types}
\begin{tabular}{lll}
\toprule
\textbf{Type} & \textbf{Source} & \textbf{Description} \\
\midrule
\texttt{object} & Base & Movable rigid objects (blocks, tools, etc.) \\
\texttt{conf} & Base & Robot joint configurations $\mathbf{q} \in \mathbb{R}^{\text{DOF}}$ \\
\texttt{grasp} & Base & Grasp transforms in object frame \\
\texttt{pose} & Base & Object placement/drop poses \\
\texttt{traj} & Base & Joint-space trajectories \\
\texttt{container} & Extended & Objects that can receive dropped items \\
% \texttt{articulated} & Extended & Objects with open/close mechanics \\
% \texttt{button} & Extended & Toggle switches \\
\bottomrule
\end{tabular}
\end{table}

\subsubsection{Skills (Operators)}

Table~\ref{tab:skills} lists all skills with their preconditions and effects:

\begin{table}[h]
\centering
\caption{Planning domain skills with preconditions and effects.}
\label{tab:skills}
\small
\resizebox{\columnwidth}{!}{%
\begin{tabular}{p{1.2cm}p{2.5cm}p{4.5cm}p{4.5cm}}
\toprule
\textbf{Skill} & \textbf{Parameters} & \textbf{Preconditions} & \textbf{Effects} \\
\midrule
\textsc{Pick} & $(o, g, q, q', t)$ &
  $\text{HandEmpty} \wedge \text{On}(o, \_) \wedge \text{Clear}(o)$ &
  $\text{Holding}(o) \wedge \neg\text{HandEmpty} \wedge \neg\text{On}(o, \_) \wedge \text{Clear}(\_\text{below})$ \\
\addlinespace
\textsc{Place} & $(o, g, q, q', p, t, s)$ &
  $\text{Holding}(o) \wedge \text{Clear}(s)$ &
  $\text{On}(o, s) \wedge \neg\text{Holding}(o) \wedge \text{HandEmpty} \wedge \text{Clear}(o) \wedge \neg\text{Clear}(s)$ \\
\addlinespace
\textsc{Drop} & $(o, g, q, q', p, t, c)$ &
  $\text{Holding}(o) \wedge \text{Accessible}(c)$ &
  $\text{InContainer}(o, c) \wedge \neg\text{Holding}(o) \wedge \text{HandEmpty}$ \\
% \addlinespace
% \textsc{Open} & $(a, q, t)$ &
%   $\text{Closed}(a) \wedge \text{HandEmpty}$ &
%   $\text{Open}(a) \wedge \neg\text{Closed}(a) \wedge \text{Accessible}(a)$ \\
% \addlinespace
% \textsc{Close} & $(a, q, t)$ &
%   $\text{Open}(a) \wedge \text{HandEmpty}$ &
%   $\text{Closed}(a) \wedge \neg\text{Open}(a) \wedge \neg\text{Accessible}(a)$ \\
% \addlinespace
% \textsc{PushButton} & $(b, q, t)$ &
%   $\neg\text{ButtonPressed}(b) \wedge \text{HandEmpty}$ &
%   $\text{ButtonPressed}(b)$ \\
\bottomrule
\end{tabular}%
}
\end{table}

\subsubsection{Continuous Parameter Generators}

Each stream maps known continuous parameters to unknown ones. The streams are processed in dependency order during CCSP solving:

\begin{table}[h]
\centering
\caption{Continuous parameter Samplers.}
\label{tab:streams}
\small
\resizebox{\columnwidth}{!}{%

\begin{tabular}{lp{3cm}p{2cm}p{5cm}}
\toprule
\textbf{Stream} & \textbf{Inputs} & \textbf{Outputs} & \textbf{Method} \\
\midrule
\texttt{sample-grasp} & object $o$ & grasp $g$ & Antipodal or cached grasp sampling \\
\texttt{sample-placement} & object $o$, surface $s$ & pose $p$ & Ray-based or AABB placement \\
\texttt{sample-drop-pose} & object $o$, container $c$ & pose $p$ & Drop pose sampling via opening extraction \\
\texttt{solve-ik} & object $o$, grasp $g$, pose $p$ & config $q$ & cuRobo GPU-accelerated IK \\
\texttt{plan-motion} & config $q_s$, config $q_g$ & traj $t$ & cuRobo trajectory optimization \\
\texttt{plan-motion-holding} & config $q_s$, config $q_g$, object $o$, grasp $g$ & traj $t$ & cuRobo with attached object \\
\bottomrule
\end{tabular}%
}
\end{table}

\subsubsection{Initial State Construction}

The initial symbolic state is constructed from the reconstruction pipeline output - taken from \cref{app:real2sim-state}:
\begin{itemize}
    \item $\text{HandEmpty}$: Always true initially.
    \item $\text{On}(o, s)$: Derived from reconstructed support relations. By default, all movable objects are on the table unless the reconstruction specifies otherwise (e.g., $\text{On}(\texttt{block\_1}, \texttt{block\_2})$ for stacked objects, or $\text{InContainer}(\texttt{spoon\_1}, \texttt{bowl\_1})$ for contained objects).
    \item $\text{Clear}(o)$: True for objects with nothing on top. Computed as the complement of $\{s : \exists\, o.\; \text{On}(o, s)\}$ restricted to movable objects.
    \item $\text{Accessible}(c)$: True for all non-articulated containers.
    % \item $\text{Closed}(a)$: True for all articulated objects initially (drawers, lids).
\end{itemize}

\subsection{GPU Parallelized Continuous CCSP Solver}
\label{app:stamp-ccsp}

The key implementation detail for the generalized TAMP planner used in the real system (and for inference) is the use of a GPU parallelized continuous constraint satisfaction problem solver. Specifically, we employ a particle-based solver that instantiates plan skeletons by sequentially sampling and filtering continuous parameters.

\subsubsection{Algorithm Overview}

Given a compiled plan skeleton $ \pi = (\text{skill}_1, \ldots, \text{skill}_M)$, the solver maintains $N$ particles, each representing a candidate continuous solution. The solver processes skills in order; within each skill, it processes subgraph-conditional samplers in dependency order, following the CCSP graph. At each stream, all $N$ particles are updated in parallel (batched on GPU):

\begin{algorithm}
\caption{Particle Sequential CCSP Solver}
\label{alg:particle_solver}
\KwIn{Compiled skills $(\text{skill}_1,\ldots,\text{skill}_M)$, world state $\mathcal{W}$, configuration $\theta$}
\KwOut{Solution bindings $\mathcal{B}^*$ or failure}

$\mathcal{P} \gets \textsc{InitializeParticles}(N,\mathcal{W},\mathbf{q}_0)$\;

\For{$m \gets 1$ \KwTo $M$}{
    $\mathit{subgraph} \gets
    \textsc{ExtractSubgraphOrder}(\text{skill}_m)$\;
    \tcp{Arrange streams in dependency order}

    \ForEach{stream $\sigma \in \mathit{subgraph}$}{
        $\textsc{SampleAndWeight}(\mathcal{P},\sigma)$\;
        \tcp{Batch sampling and feasibility filtering}

        $\mathrm{ESS} \gets
        \displaystyle\frac{1}{\sum_i w_i^2}$\;
        \tcp{Effective sample size}

        \If{$\mathrm{ESS} < N \rho_{\mathrm{ESS}}$}{
            $\mathcal{P} \gets
            \textsc{SystematicResample}(\mathcal{P},N)$\;
            \tcp{$\rho_{\mathrm{ESS}}=0.5$}
        }
    }

    $\textsc{ApplySkillEffects}(\mathcal{P},\text{skill}_m)$\;
    \tcp{Update the world state}
}

\If{$\left|\{p \in \mathcal{P} : p.\text{feasible}\}\right| > 0$}{
    $p^* \gets \displaystyle\arg\max_{p \in \mathcal{P}} w_p$\;
    \tcp{Select the best particle}

    \Return{$\textsc{StitchMotion}(p^*)$}\;
    \tcp{Plan trajectories}
}
\Else{
    \Return{\textsc{Failure}}\;
}
\end{algorithm}

\subsubsection{Subgraph-Conditional Sampler Dependency Resolution}

Within each skill, subgraph-conditional samplers are processed in a topological order determined by fixed-point iteration over parameter dependencies:
\begin{enumerate}
    \item Initialize the \emph{known} set with parameters that have data (constants from previous skills or the initial state).
    \item Find streams whose input parameters are all in the known set.
    \item Add their outputs to the known set and record the stream in the execution order.
    \item Repeat until all streams are processed or no progress can be made.
\end{enumerate}
\subsubsection{Particle State}

Each particle $p_i$ maintains:
\begin{itemize}
    \item \texttt{robot\_config}: Current joint configuration $\mathbf{q} \in \mathbb{R}^7$.
    \item \texttt{object\_poses}: Dictionary mapping object names to current $\text{SE}(3)$ poses.
    \item \texttt{held\_object}: Name of currently grasped object (or \texttt{None}).
    \item \texttt{grasp}: Current grasp transform (if holding).
    \item \texttt{variables}: Dictionary of solved continuous parameters (grasps, poses, configs).
    \item \texttt{constants}: Dictionary of known constants (initial poses, fixed configs).
    \item \texttt{waypoints}: List of $(q_\text{start}, q_\text{end})$ pairs for motion stitching.
    \item \texttt{is\_feasible}: Boolean flag; infeasible particles are excluded from sampling.
    \item \texttt{log\_weight}: Log importance weight (updated additively).
\end{itemize}
\subsubsection{Effective Sample Size and Resampling}

Particle weights are maintained in log-space and normalized via log-sum-exp. The effective sample size (ESS) is:
\begin{equation}
    \text{ESS} = \frac{1}{\sum_{i=1}^N \bar{w}_i^2}, \quad \bar{w}_i = \frac{\exp(\log w_i)}{\sum_{j=1}^N \exp(\log w_j)}
\end{equation}
When $\text{ESS} < N \cdot \rho_\text{ESS}$ (with $\rho_\text{ESS} = 0.5$), systematic resampling is triggered. Systematic resampling draws $N$ particles from the current set with probabilities proportional to their normalized weights, using a single uniform random offset $u \sim \text{Uniform}(0, 1/N)$ to select indices. After resampling, all particle weights are reset to uniform ($\log w_i = 0$). Resampling is checked after each parameter rather than only after each skill.

\subsection{Continuous Parameter Samplers}
\label{sec:stamp_samplers}

Each sub-graph solver in the CCSP is backed by a batched sampler that generates candidate values for all $N$ particles simultaneously.

\subsubsection{Grasp Sampling}

The primary grasp generation strategy is \emph{antipodal sampling} on object meshes, generated from the real2sim pipeline detailed in \cref{app:real2sim}. 

\subsubsection{Placement Sampling}
\label{sec:placement_sampling}

Placement sampling generates candidate SE(3) poses for placing objects on support surfaces. The primary engine is GPU-accelerated ray-based placement.

\paragraph{Ray-Based Placement.} The placement sampler uses NVIDIA Warp for GPU-accelerated mesh ray-casting. The procedure:
\begin{enumerate}
    \item \textbf{Stable Pose Computation.} For each object, compute the set of physically stable resting orientations via convex hull analysis. Up to $P = 16$ stable poses are retained, each with a probability weight. For each stable pose, footprint probe points (4 corners $+$ center on the bottom face) and ceiling probe points are pre-computed and stored on GPU.
    \item \textbf{Scene Mesh Construction.} All entity meshes (movable objects and statics including the table) are loaded onto GPU. Entity transforms are updated dynamically via to reflect current particle-tracked object poses.
    \item \textbf{Grid Sampling.} A grid of $N_\text{grid} \times N_\text{grid}$ (default $32 \times 32$) candidate XY positions is generated over the support surface extent, centered on the support object's current position (which may have moved in multi-skill plans).
    \item \textbf{Downward Ray-Casting.} For each grid point and each stable pose, downward rays are cast from above through the footprint probe points. A placement is valid if at least $f_\text{support}$ fraction (default $0.5$) of the footprint points hit the target support surface. 
\end{enumerate}

\subsubsection{Drop/Insertion Pose Sampling}

Drop pose sampling generates candidate poses above/inside container openings where the robot should insert and release held objects. The procedure operates in the container's local frame, where we use the mesh to formulate a canonical 2D opening region, and sample poses and heights that are valid for the containment of the object.

\subsubsection{Inverse Kinematics}
\label{sec:ik_solving}

IK solving uses NVIDIA cuRobo \cite{sundaralingam2023curobo} for GPU-accelerated collision-aware inverse kinematics.

\paragraph{Batch Solving.} Target gripper poses are computed from the object pose and grasp transform as $T_\text{grip} = T_\text{obj} \circ T_\text{grasp}$. All $N$ particle targets are batched into a single cuRobo IK call. To maintain CUDA graph efficiency, batches smaller than $N$ are padded to $N$ by repeating the last valid pose; padding results are discarded after solving.

\paragraph{Collision Management.} Before IK solving, the collision world is updated to reflect the current particle state:
\begin{itemize}
    \item For \textsc{Pick} skills, the target object's collision body is \emph{disabled} in cuRobo to allow the gripper to reach into the object. It is re-enabled after IK solving.
    \item For \textsc{Place}/\textsc{Drop} skills, all object poses are updated in cuRobo's collision tensors to reflect their current positions (objects may have moved in earlier skills).
    \item After \textsc{Place}/\textsc{Drop} skills, the collision world is synced with the particle state.
\end{itemize}

\subsection{Motion Planning and Trajectory Stitching}
\label{sec:stamp_motion}

Motion planning, the most expensive step, is deferred to the end, and only applied to \say{valid} particles found during CCSP solving.

\subsubsection{Waypoint Recording}

During CCSP solving with motion stitching enabled, motion planning samplers (\texttt{plan-motion}, \texttt{plan-motion-holding}) record $(q_\text{start}, q_\text{end}, \text{held\_obj}, \text{grasp})$ waypoints instead of actually planning trajectories. This dramatically accelerates the particle filter by avoiding expensive GPU trajectory optimization during the sampling phase.

\subsubsection{Trajectory Stitching}

After the CCSP solver finds a feasible particle, the \texttt{stitch\_motion()} method plans actual trajectories between recorded waypoints. Waypoints are grouped by attachment mode (free-space vs.~holding) to minimize collision world reconfiguration. Consecutive waypoints with the same held object form a \emph{motion group}. Within a group, the collision world (attachment configuration, disabled objects) is set up once. Between groups, attachments are reconfigured. When holding an object during motion planning, the object's collision spheres are transformed from object frame to gripper frame via the grasp inverse transform. Spheres are down-sampled or zero-padded to match cuRobo's fixed attachment sphere count. Then, All waypoints in a group are planned simultaneously using batched joint-space planning (implemented with cuRobo traj-opt). 

\subsection{Implementation Details for Real-Robot Rationality Scoring}

As detailed in \cref{sec:rir}, we use TAMP to score candidate explanation programs (task specifications) against observed demonstrations using a Boltzmann rationality extended for the bi-level planning formulation. In practice, we use the parallelized sampling to perform efficient refinements over plan skeletons and task specifications, also allowing us to quickly estimate relative specificity corrections.

\paragraph{Trajectory distance cost} In our real-robot experiments, due to the cost of running full trajectory stitching to estimate true configuration space costs for every single particle, we instead compute approximate heuristic costs that act as estimates, relying only on the particle waypoints which can be sampled in parallel. To compute the heuristic, we densely interpolate between waypoints in joint space and computing the SE(3) path length along the resulting end-effector trajectory. For each waypoint pair $(q_m^\text{start}, q_m^\text{end})$:
\begin{enumerate}
    \item Linearly interpolate $L$ configurations in joint space: $\mathbf{q}^{(\ell)} = (1 - t_\ell)\, \mathbf{q}_m^\text{start} + t_\ell\, \mathbf{q}_m^\text{end}$, where $t_\ell = \ell / (L+1)$ for $\ell = 0, 1, \ldots, L+1$ (default $L = 10$ interior points, giving $L{+}2 = 12$ total).
    \item Compute end-effector poses at all interpolated configurations via \emph{batched GPU forward kinematics} through cuRobo. All configurations across all particles and all waypoints are concatenated into a single FK batch call for maximal GPU utilization.
    \item Compute the dense SE(3) path length by summing consecutive segment distances:
\end{enumerate}
\begin{equation}
    c_\text{VALID} = \sum_{m=1}^{M} \sum_{\ell=0}^{L} \left( \|\mathbf{p}_m^{(\ell+1)} - \mathbf{p}_m^{(\ell)}\| + \alpha_\text{rot} \cdot d_\text{geo}(\mathbf{q}_m^{(\ell)}, \mathbf{q}_m^{(\ell+1)}) \right)
\end{equation}
This captures the nonlinearity of the kinematic mapping: a small joint-space displacement can produce a large Cartesian displacement (and vice versa), so the dense SE(3) path length is a more faithful estimate of the actual motion effort than the endpoint-only distance.

\paragraph{VLM sampling} For the VLM hypothesis generation, we use the \texttt{Gemini-3-Pro} model to generate 5 candidate explanation programs, following our initial TERC experiments which suggest that 5 particles was sufficient to capture the posterior and enable refinement. The VLM prompts corresponding to API access for the real2sim and downstream TAMP planning system are provided in Listings~\ref{lst:real-robot-api} and \ref{lst:real-robot-api2}.

\section{Real-World Experiments} \label{app:robot_exp}

\begin{figure}
    \centering
    \includegraphics[width=\linewidth]{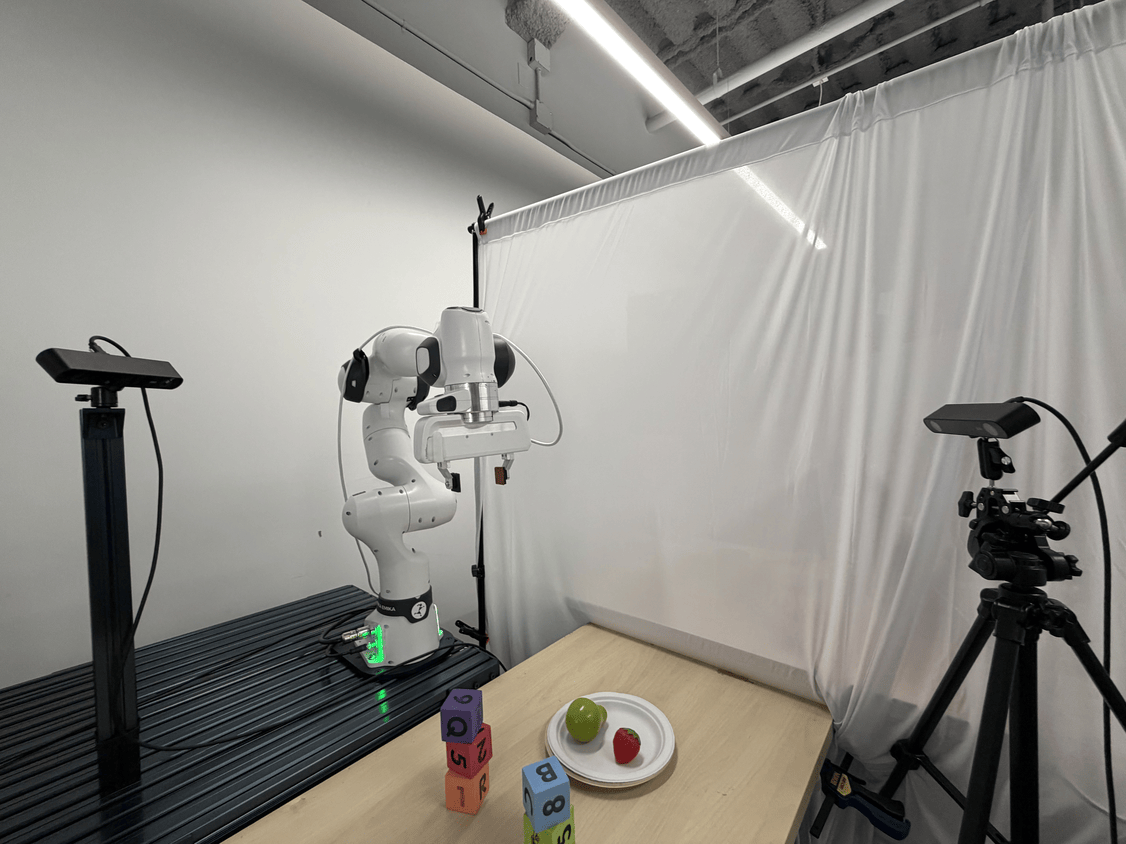}
    \caption{Overview of the real-robot setup, where we use two Zed-2i cameras viewing the tabletop scene, and a Frank Research FR3 7-DoF robotic arm to perform manipulation}
    \label{fig:real-robot-overview}
\end{figure}

In this section we outline additional details and experimental results for our deployment and scaling of the RIR algorithm to real-robot tasks. First, we discuss the experimental setup, including the task distribution, and provide examples of the demonstration initial states and inferred explanation programs from 3 demonstrations for each task. Then, we provide an assessment of the success rate associated with executing from $E$ on the real tabletop setup, shown in \cref{fig:real-robot-overview}.

\subsection{Task Distribution and Inferred Explanation Programs}

\begin{figure*}[t]
    \centering
    \includegraphics[width=\linewidth]{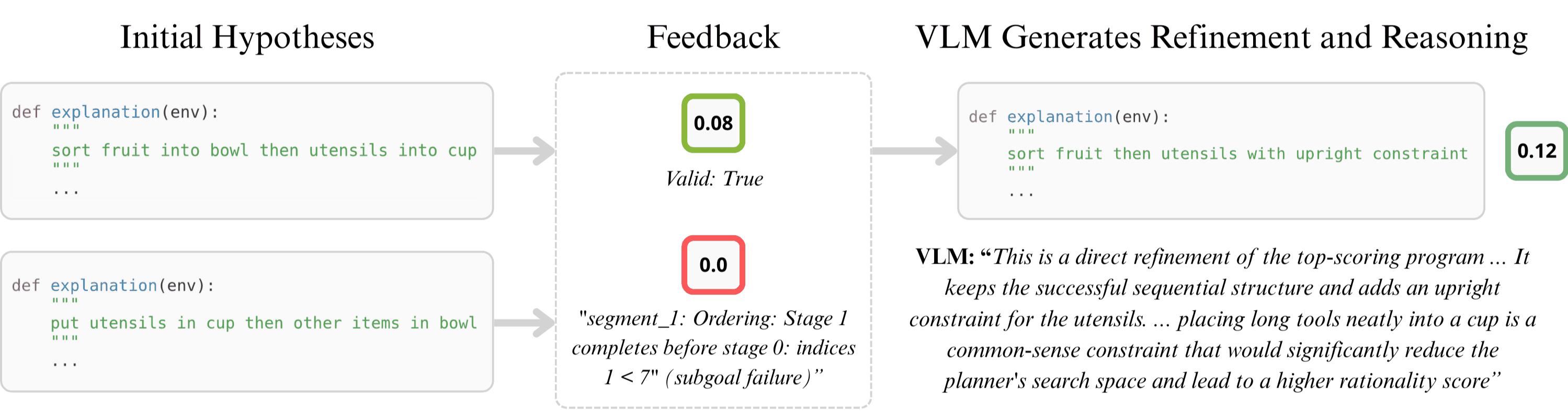}
    \caption{Case study example of the additional feedback from the planner to provide guided edits on the explanation program hypotheses. Task shown: \texttt{Tidying}.}
    \label{fig:tidying-feedback}
\end{figure*}

\begin{figure*}[t]
    \centering
    \includegraphics[width=\linewidth]{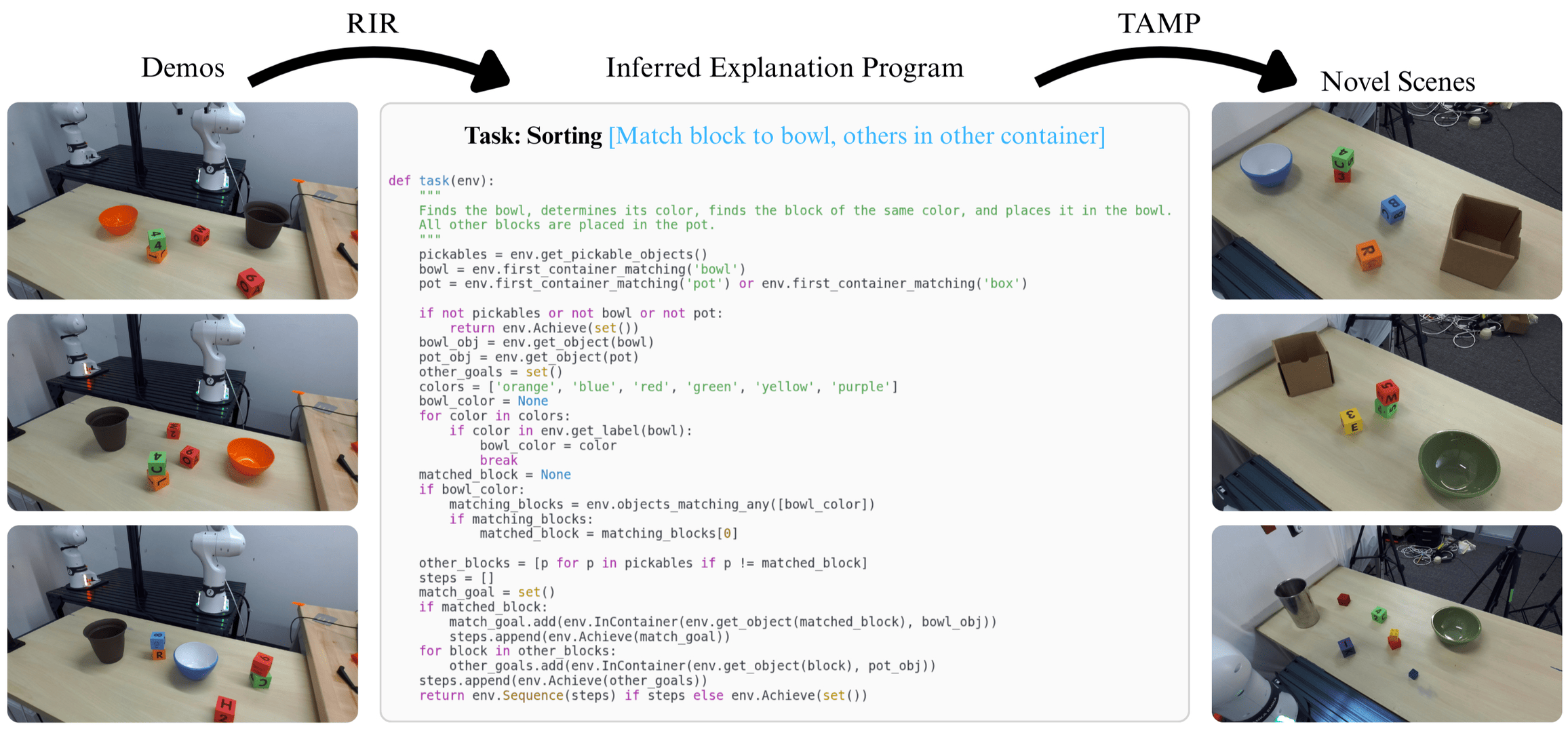}
    \caption{The RIR algorithm is used to extract the explanation shown, which is re-grounded and applied on novel scenes (3 examples shown). Here, the task is to match the block to the bowl by color first, and then drop the other blocks in the other container. The system infers an explanation program describing this task, and allows us to execute it on completely novel scenes, with novel objects and layouts.}
    \label{fig:real-robot-sorting}
\end{figure*}

\begin{figure*}[b]
    \centering
    \includegraphics[width=\linewidth]{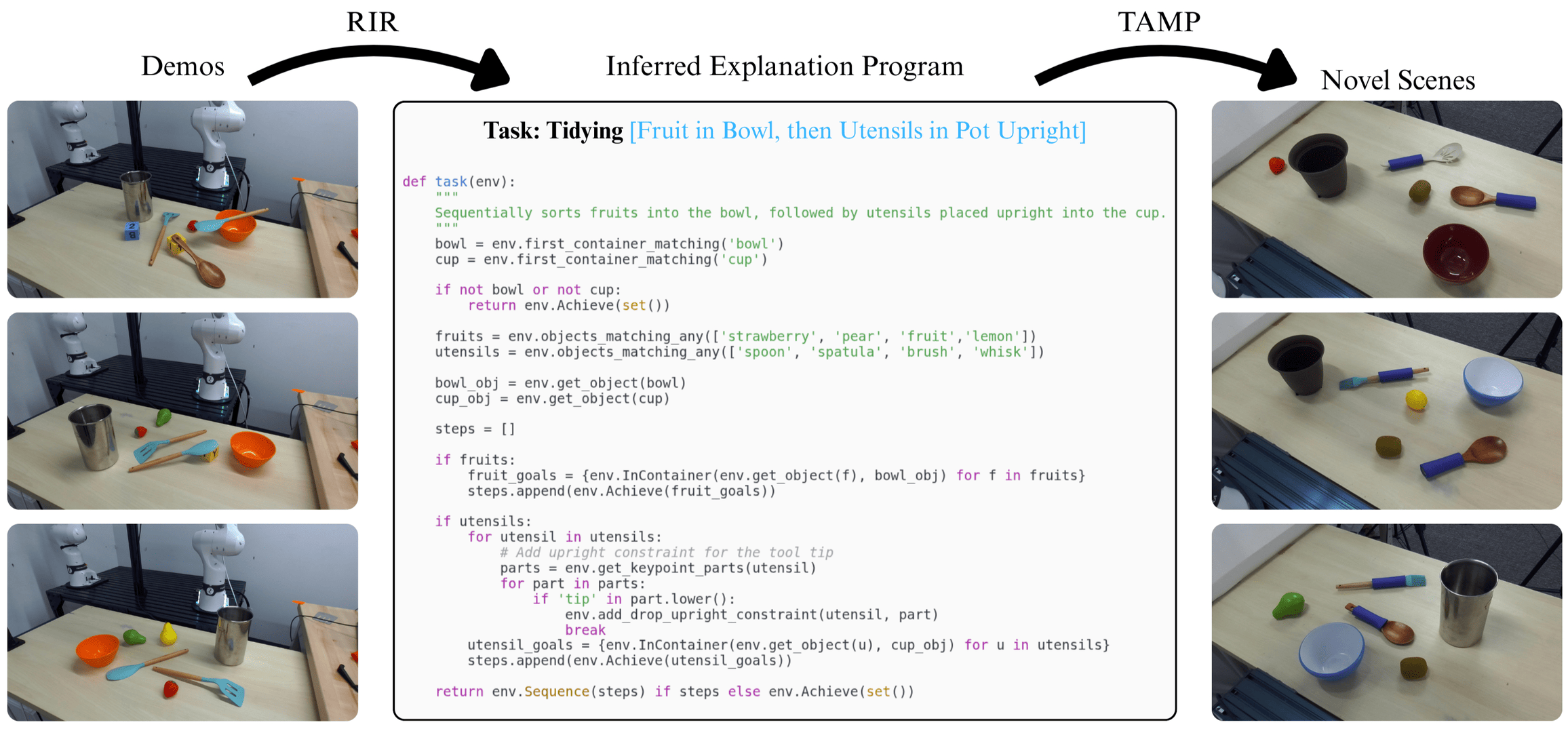}
    \caption{The RIR algorithm is used to extract the explanation shown, which is re-grounded and applied on novel scenes (3 examples shown). Here, the task is to first place the fruit in the bowl, then place the utensils upright in the pot/cup. The system infers an explanation program describing this task, and allows us to execute it on completely novel scenes, with novel objects and layouts.}
    \label{fig:real-robot-tidying}
\end{figure*}

\begin{figure*}[t]
    \centering
    \includegraphics[width=\linewidth]{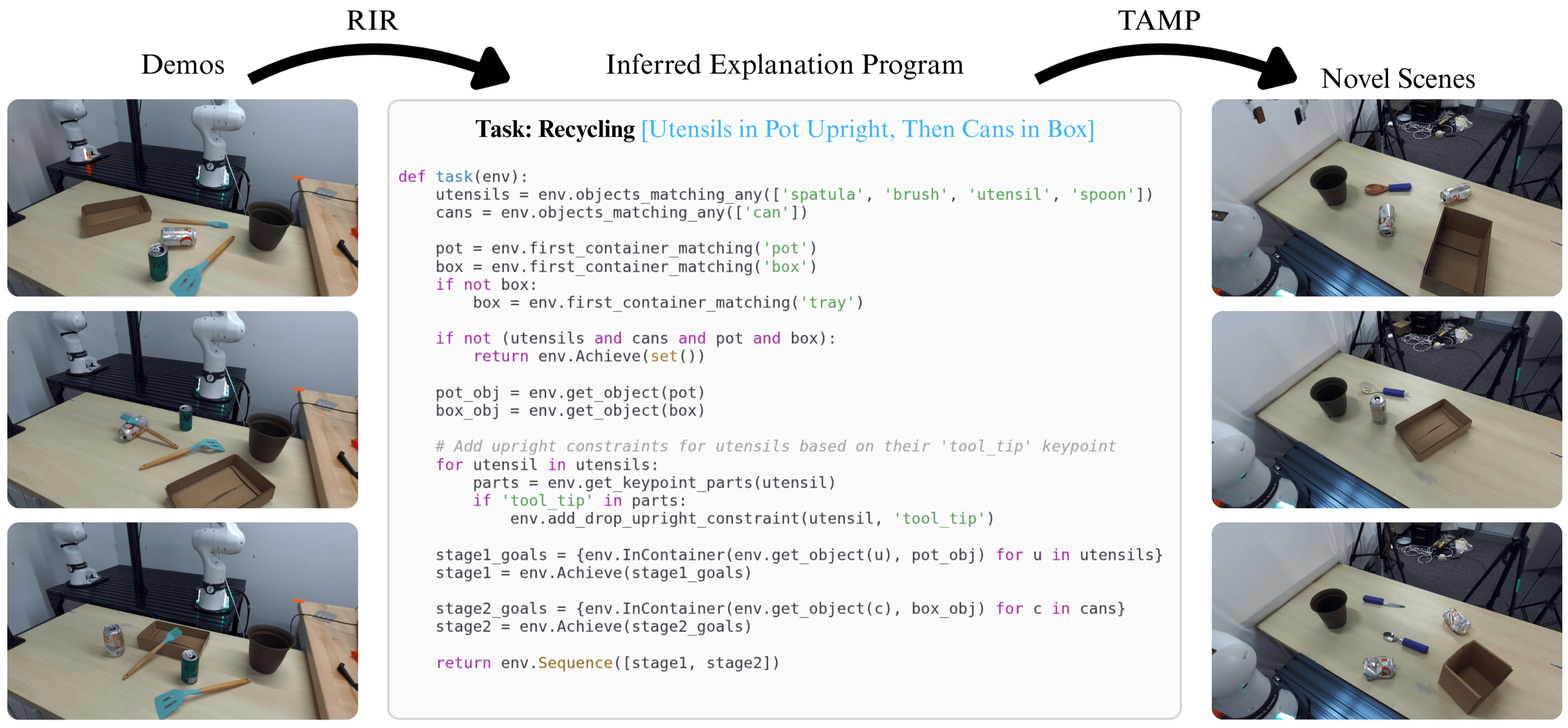}
    \caption{The RIR algorithm is used to extract the explanation shown, which is re-grounded and applied on novel scenes (3 examples shown). Here, the task is to tidy up the utensils, placing them upright in the cup, and then pack the recycling - cans - into the cardboard box. The system infers an explanation program describing this task, and allows us to execute it on completely novel scenes, with novel objects and layouts.}
    \label{fig:real-robot-recycling}
\end{figure*}

\begin{figure*}[b]
    \centering
    \includegraphics[width=\linewidth]{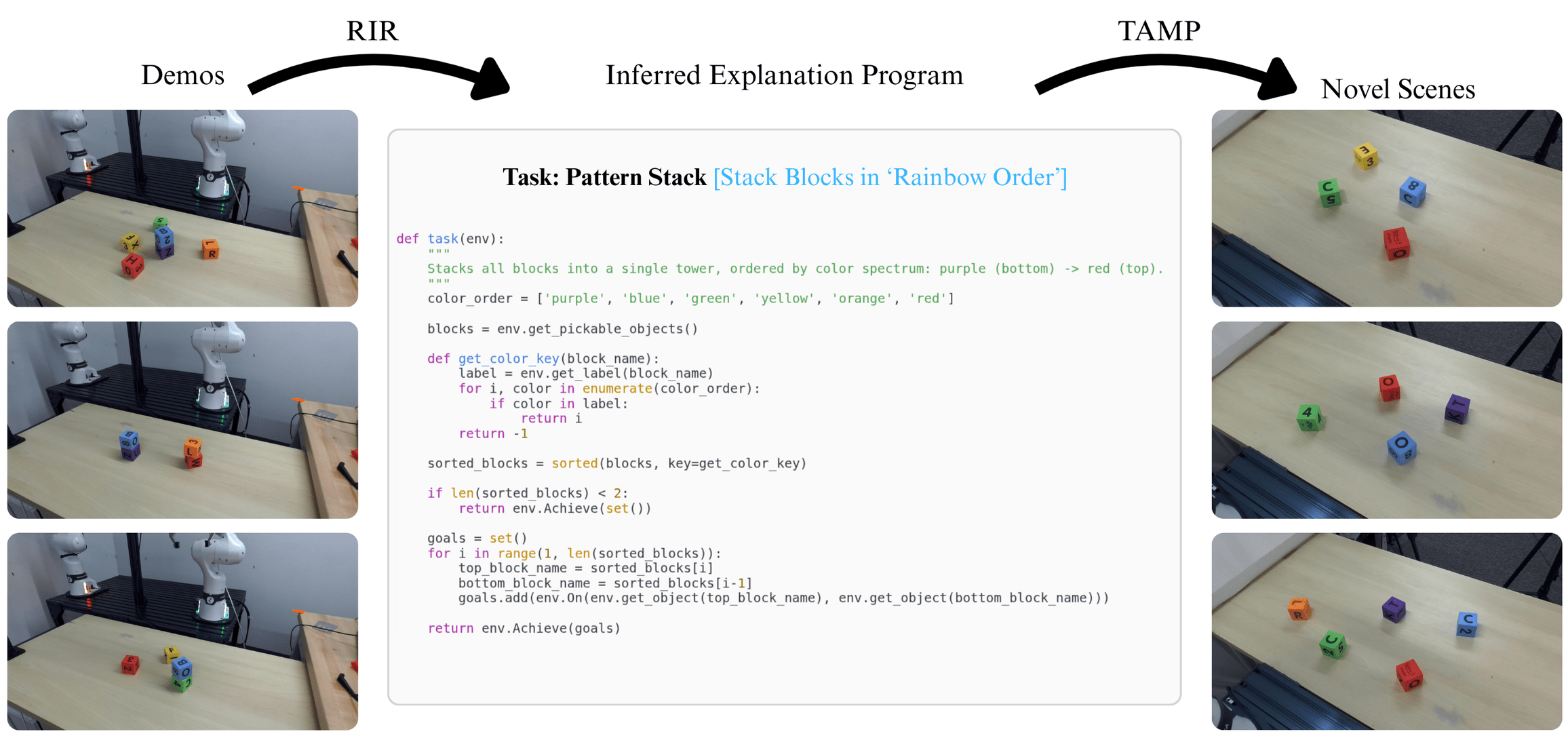}
    \caption{The RIR algorithm is used to extract the explanation shown, which is re-grounded and applied on novel scenes (3 examples shown). Here, the task is to stack blocks in "rainbow" order. That is, red, orange, yellow, green, blue, purple (top to bottom). The system infers an explanation program describing this task, and allows us to execute it on completely novel scenes, with novel objects and layouts.}
    \label{fig:real-robot-pattern}
\end{figure*}

In figures~\ref{fig:real-robot-sorting}, \ref{fig:real-robot-tidying}, \ref{fig:real-robot-recycling} and \ref{fig:real-robot-pattern} we illustrate the initial states for the demonstration distribution, as well as examples of the completely novel scenes that we evaluate on. From 1-3 demonstrations, RIR is able to infer an interpretable and executable explanation program that can be grounded in completely novel scenes, with novel objects and layouts, achieving strong few-shot generalization.

\subsection{Case-Study Example of Feedback Diagnostics}

To allow the VLM to improve its explanation hypotheses over time, we use the planner to provide rational feedback. This comes in the form of the actual likelihood, as well as additional surprise signals/diagnostic information that is useful to pin-point exactly what parts of the program are incorrect/need adjustment. In particular, if a particular explanation program results in constraints, continuous or discrete, which are violated by the demonstrations, this is referred back to the high-level VLM, to improve guided edits. For example, consider \cref{fig:tidying-feedback}, where we see an actual example of the planner feedback and subsequent improvements to the explanation programs across iterations. In this case, the VLM produces an initial hypothesis which produces the incorrect subgoals from the video (i.e. utensils first, then other times). The planner extracts this failure, and provides a natural language description, including the exact demonstration and stage where this failure occurs. \cref{fig:tidying-feedback} also contains an example of the extended reasoning behind a program candidate being added to the particle pool. We can see that the VLM processes the feedback from all of the previous particles, and updates/refines the hypotheses accordingly, increasing the rationality score across iterations.

\section{VLM Prompting} \label{app:prompting}

For completeness, we include all VLM prompts across both domains. 

\begin{figure*}[t]               % the * makes it span both columns
  \centering
  \begin{minipage}{\textwidth}   % keeps the listing the full width
  \lstset{basicstyle=\ttfamily\small,
          frame=single,
          }

  \begin{lstlisting}[label={lst:video-input-prompt}, caption={Video input prompt}]
#define demonstration video inputs
You have observed a set of short videos of the robot's behavior in the environment. Importantly, the robot is executing the same behavior in all the demos. You should use these videos, along with the other information provided, to understand and solve the task.
The videos are provided as a series of frames, which you can use to understand the robot's behavior.

#define visual description
The videos/images depict a visualization of the environment, which contains a set of objects. This is a representation of a top-down view of a 2d object rearrangement task. The robot/gripper/manipulator is represented as a small blue circle, which can pick up and move the objects around.
  \end{lstlisting}
  \end{minipage}
\end{figure*}

\begin{figure*}[t]               % the * makes it span both columns
  \centering
  \begin{minipage}{\textwidth}   % keeps the listing the full width
  \lstset{basicstyle=\ttfamily\small,
          frame=single}

  \begin{lstlisting}[caption={Explanation Function Definition Prompt}, label={lst:function-definition-prompt}]
#define explanation function problem specification
You have access to the following set of predicates and locations. You have no other predicates you can use, and the typing and arity of the predicate inputs must match the specification below.
Predicates:
- At(object, location) e.g. At(box1, Left), NOT At(box1, box2)
Locations:
- Left
- Right
- Top
- Bottom
- Corner
- Middle
Middle refers to the center of the environment within the 2D space, with a radius around it. Top, Left, Bottom, Right all refer to half-spaces in the 2D environment. Corner refers to the points on the vertices of the 2D environment, with a small radius around them. E.g. top-left refers to the quadrant, whilst top-left-corner specifies a smaller corner region.
Note: whilst locations like 'Left', 'Right', 'Top', 'Bottom', 'Corner' can be composed with multiple At statements, 'Middle' is a special location that can only be used with a single At statement. For example, 'At(obj, Middle)' is valid, but 'At(obj, Middle) AND At(obj, Top)' is not valid.
You also have access to the following domain-specific task-specification constructors. You must use these to compose the returned task specification. The definitions for these components are shown below.
- Achieve(*goals: Set[Predicate]) -> TaskSpec
- Sequence(*task_specs: TaskSpec) -> TaskSpec
Your goal is to generate a python function named `explanation` that takes in the environment specification (a collection of objects), and returns the grounded task specification.
The explanation is an abstract program that grounds abstract [goals, constraints and subgoals] (task specification) into a grounded task specfication, which can actually be executed by a TAMP planner.
The explanation function should be as general as possible such that it can be applied to achieve the required task across many environment.
For example (in natural language), our explanation function may capture the abstract goals of `stack all the green blocks`. Then mapping the environement specification `2 green blocks and 1 red block` through this yields: `stack the 2 green blocks`. Importantly, instead of natural language, the explanation function produces a task specification where goals, constraints and sub-goals are defined with a vocabulary of grounded predicates and constants. This makes it amenable for TAMP planning.
Brevity is key because we use the function length in a program synthesis prior to penalise 'complexity'. As a result, please do not write any comments in the explanation functions.
The main function should be named EXACTLY `explanation`. Do not change the names. Do not create additional classes or overwrite existing ones.
  \end{lstlisting}
  \end{minipage}
\end{figure*}

\begin{figure*}[t]               % the * makes it span both columns
  \centering
  \begin{minipage}{\textwidth}   % keeps the listing the full width
  \lstset{basicstyle=\ttfamily\small,
          frame=single}
  \begin{lstlisting}[caption={Feedback Prompt}, label={lst:feedback-prompt}]
Here is a list of some of your previously generated explanation functions. A rationalization process has been applied to these functions, and the likelihoods that they are rational and correct is provided.
Please use this list to help you formulate a new set of {N} explanation functions that perform better. Note that if two functions have similar likelihoods, they may both be correct. The probabilities are not absolute, but rather relative to each other. They also don't reflect the complexity preferences. You should prefer simpler programs, even if they have the same probability. A probability of 0 means the function is unlikely to be correct and/or is not rational.
Remember, try to make the explanation functions as general as possible to explain the behavior:
...
Please generate improved explanation functions using the ranking above.
The functions at the top of the ranking are more likely to be correct.
Please keep them short (few lines) and general - the goal is to generalize the explanation algorithm to work across many environments
  \end{lstlisting}
  \end{minipage}
\end{figure*}

\begin{figure*}[t]               % the * makes it span both columns
  \centering
  \begin{adjustbox}{max width=\textwidth, max totalheight=\textheight, keepaspectratio}
  \begin{minipage}{\textwidth}   % keeps the listing the full width
  \lstset{basicstyle=\ttfamily\small,
          frame=single}
  \begin{lstlisting}[caption={In-Context Examples Prompt (Pt.1 of 2)}, label={lst:in-context-prompt1}]
#define in-context examples
You have access to the following in-context examples. These examples are used to help you understand the task and how to generate the explanation function(s).
Description: Place the blue object(s) on the left:

def explanation(env):
    goal_set = set()
    for obj in env:
        if obj.color == 'Blue':
            goal_set.add(at(obj, Left))
    return Achieve(goal_set)

Description: Place the red object(s) at the top-right:

def explanation(env):
    goal_set = set()
    for obj in env:
        if obj.color == 'Red':
            goal_set.add(at(obj, Right))
            goal_set.add(at(obj, Top))
    return Achieve(goal_set)

Description: Place all the plates in the dishwaher:

def explanation(env):
    goal_set = set()
    for obj in env:
        if obj.type == 'plate':
            goal_set.add(at(obj, Dishwasher))
    return Achieve(goal_set)

Description: Place the big bowls in the bottom-cupboard, and the small bowls in the top-cupboard:

def explanation(env):
    goal_set = set()
    for obj in env:
        if obj.type == 'bowl':
            goal_set.add(In(obj, Cupboard))
            if obj.size == 'big':
                goal_set.add(At(obj, Bottom))
            elif obj.size == 'small':
                goal_set.add(At(obj, Top))
    return Achieve(goal_set)

Description: Place the smallest triangle in the top-left corner:

def explanation(env):
    goal_set = set()
    smallest_triangle = None
    for obj in env:
        if obj.type == 'triangle':
            if smallest_triangle is None or obj.geometry.side_length < smallest_triangle.geometry.side_length:
                smallest_triangle = obj

    if smallest_triangle is not None:
        goal_set.add(At(smallest_triangle, Top))
        goal_set.add(At(smallest_triangle, Left))
        goal_set.add(At(smallest_triangle, Corner))

    return Achieve(goal_set)
  \end{lstlisting}
  \end{minipage}
  \end{adjustbox}
\end{figure*}

\begin{figure*}[t]               % the * makes it span both columns
  \centering
  \begin{adjustbox}{max width=\textwidth, max totalheight=\textheight, keepaspectratio}
  \begin{minipage}{\textwidth}   % keeps the listing the full width
  \lstset{basicstyle=\ttfamily\small,
          frame=single,
          caption={}}
  \begin{lstlisting}[caption={In-Context Examples Prompt (Pt.2 of 2)}, label={lst:in-context-prompt2}]

Description: If there are two red objects, place them at the top, otherwise, place them at the bottom:

def explanation(env):
    goal_set = set()
    red_objects = [obj for obj in env if obj.color == 'Red']
    if len(red_objects) == 2:
        for obj in red_objects:
            goal_set.add(At(obj, Top))
    else:
        for obj in red_objects:
            goal_set.add(At(obj, Bottom))
    return Achieve(goal_set)

Place the gray box in the top-left corner, then place the red box in the bottom-right corner:

def explanation(env):
    goal_set1 = set()
    goal_set2 = set()
    for obj in env:
        if obj.color == 'Gray' and obj.type == 'box':
            goal_set1.add(At(obj, Top))
            goal_set1.add(At(obj, Left))
            goal_set1.add(At(obj, Corner))
        elif obj.color == 'Red' and obj.type == 'box':
            goal_set2.add(At(obj, Bottom))
            goal_set2.add(At(obj, Right))
            goal_set2.add(At(obj, Corner))
    return Sequence(Achieve(goal_set1), Achieve(goal_set2))

Stack all the blue cubes, then all the red cubes:

def explanation(env):
    stack_list = []
    for obj in env:
        if obj.color == 'Blue' and obj.type == 'cube':
            stack_list.append(obj)
    a1 = Achieve(Stack(stack_list))

    stack_list2 = []
    for obj in env:
        if obj.color == 'Red' and obj.type == 'cube':
            stack_list2.append(obj)
    a2 = Achieve(Stack(stack_list2))

    return Sequence(a1, a2)

Move all the objects to the middle (centre) of the environment:    

def explanation(env):
    goal_set = set()
    for obj in env:
        goal_set.add(at(obj, Middle))
    return Achieve(goal_set)

Note:
Please wrap any python functions in triple backticks (```) to ensure they are formatted correctly. For example:
```python
def explanation(env):   
    # Your code here
```

Note2: Try to make the explanation function as general as possible - usually this involves writing short functions, such that it can be applied to achieve the required abstract task across many environments.
  \end{lstlisting}
  \end{minipage}
  \end{adjustbox}
\end{figure*}

\begin{figure*}[t]               % the * makes it span both columns
  \centering
  \begin{adjustbox}{max width=\textwidth, max totalheight=\textheight, keepaspectratio}
  \begin{minipage}{\textwidth}   % keeps the listing the full width
  \lstset{basicstyle=\ttfamily\small,
          frame=single,
          }

  \begin{lstlisting}[label={lst:human-input-prompt}, caption={Survey Instructions}]
In this questionnaire, you will be shown sets of short video clips, and your goal is to formulate possible explanations for what is happening in the video.

For each abstract behavior:
First, you will be shown one video clip of a specific behavior - You will then be asked to provide a possible explanation of what happened during the video
Second, you will be shown three clips of the same behavior - You will then be asked to provide a possible explanations of the behavior again, this time given all the information across the 3 videos.
There will be 35 behavior questions - they will get more challenging.

What form should my explanation take?
try to formulate explanations that best describe the behaviors in the video, but not specific to that video/example
explanations should describe the general strategy/intention, not exactly what happened in the video(s)
We will now go through an example ...

[VIDEO EXAMPLE]


From the video, we see that the blue box is moved out of the top-right-corner, allowing the red circle to be placed there.  

(the small blue circle is the cursor/teacher)

Possible explanations/algorithms could be:

1. Move the red circle to the top-right-corner
2. Move the red objects to the top-right-corner
3. Move the circles to the top-right-corner
4. Move all the objects to the top-right
5. Move all the objects to the top
6. Move the blue square to the left, then aftewards move the red object to the top-right
7. Move every object that isn't a blue box to the top-right-corner
8. Move every object in the bottom left to the top-right corner
9. Move all the 'warm' colored objects to the top-right corner
10. If there's a blue box in the environment, move the red objects to the top-right-corner

The demonstrations are intentionally ambiguous: Try to write down the explanation you think is most likely.

The purpose of this study is:
1. To determine how you choose explanations under uncertainty
2. To determine how you resolve ambiguity given more information

All the underlying explanations for the behavior are relatively simple, and will generally consist of the following vocabulary:

General Locations:
- Top
- Bottom
- Left 
- Right
- Middle
- Corner
As well as any compositions. Note, 'corner' refers to the object being all the way in the corner, close to the edges, whereas simply specifying top-right is a larger area defined by the quadrant. 

General Shapes:
- boxes/rectangles/squares
- circles
- triangles

The behaviors will generally consist of moving various classes of objects to certain locations depending on their properties, and the relationships between them. 

  \end{lstlisting}
  \end{minipage}
  \end{adjustbox}
\end{figure*}

\begin{figure*}[t]               % the * makes it span both columns
  \centering
  \begin{minipage}{\textwidth}   % keeps the listing the full width
  \lstset{basicstyle=\ttfamily\small,
          frame=single}

  \begin{lstlisting}[caption={Real Robot API Prompt}, label={lst:real-robot-api}]
Task programs are Python functions that receive an `env` object providing access to
the reconstructed scene. The function returns a task specification that STAMP uses
for planning.
### Basic Structure
```python
def task_func(env):
    """
    Task function that specifies what the robot should do.
    Args:
        env: TaskEnvInfo object with scene access and goal constructors
    Returns:
        TaskSpec: Goal specification (Achieve, Sequence, etc.)
    """
    # 1. Query objects in the scene
    objects = env.get_pickable_objects()
    containers = env.get_containers()
    # 2. Filter based on color (colors are in labels like "blue_block_1")
    # IMPORTANT: filter_by_label includes containers, so filter to pickables!
    pickables = set(env.get_pickable_objects())
    blue_objects = [o for o in env.filter_by_label("blue") if o in pickables]
    # 3. Optionally add constraints
    for obj in blue_objects:
        parts = env.get_keypoint_parts(obj)
        if parts:
            env.add_drop_upright_constraint(obj, parts[0])
    # 4. Build and return goal
    target = env.get_object(containers[0])
    goals = {env.InContainer(env.get_object(o), target) for o in blue_objects}
    return env.Achieve(goals)
```
### Environment API Reference
#### Object Queries
```python
env.get_pickable_objects() -> List[str]     # Objects robot can pick up
env.get_containers() -> List[str]           # Container objects (bowls, boxes, etc.)
env.get_all_objects() -> List[str]          # All movable objects
env.get_statics() -> List[str]              # Static entities (tables, etc.)
env.get_object(name: str) -> DomainConstant # Get domain constant by name
env.get_object_pose(name: str) -> np.ndarray # Get [x,y,z,qw,qx,qy,qz] pose
```
#### Filtering Methods
```python
# CAUTION: filter_by_label returns ALL objects (including containers!)
env.filter_by_label(substring) -> List[str]       # ALL objects with substring in label
# Safer: These return only pickable objects
env.first_object_matching(substring) -> Optional[str]  # First pickable with substring in label
env.objects_matching_any(substrings: List[str]) -> List[str]  # All pickables matching any substring
# Container-specific
env.first_container_matching(substring) -> Optional[str]  # First container with substring in label
# Description-based
env.filter_by_description(substring) -> List[str]  # All objects with substring in VLM description
env.filter_by_metadata(field, substring) -> List[str]  # Any metadata field
```
**CRITICAL: filter_by_label INCLUDES CONTAINERS!**
- `filter_by_label("orange")` returns BOTH "orange_block_1" AND "orange_bowl"
- This can cause invalid goals like `InContainer(orange_bowl, orange_bowl)`
- **ALWAYS filter to pickable objects only!**

  \end{lstlisting}
  \end{minipage}
\end{figure*}

\begin{figure*}[t]               % the * makes it span both columns
  \centering
  \begin{minipage}{\textwidth}   % keeps the listing the full width
  \lstset{basicstyle=\ttfamily\small,
          frame=single}

  \begin{lstlisting}[caption={Real Robot API Prompt Pt.2}, label={lst:real-robot-api2}]
**CORRECT PATTERN for color-based filtering:**
```python
# Option 1: Use first_object_matching (returns single pickable)
orange_block = env.first_object_matching("orange_block")
# Option 2: Filter to pickables after filter_by_label
pickables = set(env.get_pickable_objects())
orange_items = [o for o in env.filter_by_label("orange") if o in pickables]
# Option 3: Use objects_matching_any (only searches pickables)
orange_blocks = env.objects_matching_any(["orange_block"])
**WRONG (creates invalid goals):**
```python
# DON'T DO THIS - will include orange_bowl!
orange_items = env.filter_by_label("orange")  
for item in orange_items:
    goals.add(env.InContainer(env.get_object(item), orange_bowl))  # orange_bowl in orange_bowl!
```
**Color filtering note**: Colors appear in LABELS (e.g., "red_block_1", "orange_bowl").
DO NOT use `filter_by_property("color", ...)` - it returns empty.
#### Mtadata Access
```python
env.get_metadata(name: str) -> Dict[str, Any]  # Full metadata dict
env.get_label(name: str) -> str                # Object label (e.g., "red_block_1" - CONTAINS COLOR!)
env.get_description(name: str) -> str          # VLM description (e.g., "A small red cube...")
env.get_semantic_tags(name: str) -> List[str]  # Semantic category tags
```
**TIP**: To extract color from an object, check its LABEL first (e.g., "red_block_1" contains "red"),
or parse the description (e.g., "A small **red** cube..."). The `get_color()` method often returns None.
#### Keypoint/Part Access (for constraints)
```python
env.get_keypoint_parts(obj_name: str) -> List[str]  # Semantic parts (e.g., ["tines", "handle"])
env.get_keypoint_offset(obj_name: str, part_name: str) -> np.ndarray  # Part offset
env.get_keypoint_offsets() -> Dict[str, Dict[str, np.ndarray]]  # All offsets
```
#### Constraint Methods
```python
# Drop orientation constraints - ensure part points up/down when placed
env.add_drop_upright_constraint(obj_name: str, part_name: str)
env.add_drop_downward_constraint(obj_name: str, part_name: str)
# Get all added constraints
env.get_external_constraints() -> List[ExternalConstraint]
```
#### Goal Constructors
```python
# Single goal - achieve all predicates simultaneously
env.Achieve(goals: Set[Predicate]) -> TaskSpec
# Sequential goals - achieve in order
env.Sequence(steps: List[TaskSpec]) -> TaskSpec
# Predicate factories
env.InContainer(obj, container) -> Predicate  # obj is inside container
env.On(obj, surface) -> Predicate             # obj is on surface
env.Holding(obj) -> Predicate                 # robot is holding obj
```
  \end{lstlisting}
  \end{minipage}
\end{figure*}

% \appendix

\end{document}